\PassOptionsToPackage{table}{xcolor}

\documentclass[manuscript,screen,review=false]{acmart}

\usepackage{tikz}
\usepackage[most]{tcolorbox}
\usepackage{eso-pic}
\usepackage{pdflscape}
\usepackage{float}
\usepackage{booktabs}
\usepackage{array}
\usepackage{tabularx}
\usepackage{adjustbox}
\usepackage{placeins}
\usepackage{pifont}
\usepackage{wrapfig}
\usepackage{tikz}
\newcommand{\authorphoto}[1]{%
  \begin{minipage}[c][3.2cm][c]{\linewidth}
    \centering
    \includegraphics[
      width=\linewidth,
      height=3.2cm,
      keepaspectratio
    ]{#1}
  \end{minipage}%
}

\AtBeginDocument{%
}

\newcommand{\cmark}{\ding{51}}
\newcommand{\xmark}{\ding{55}}

\newcommand{\circlesize}{2.3pt}

\newcommand{\fcircle}{%
  \tikz[baseline=-0.6ex]
    \fill[black] (0,0) circle (\circlesize);%
}

\newcommand{\hcircle}{%
  \tikz[baseline=-0.6ex]{%
    \draw[black,line width=0.4pt]
      (0,0) circle (\circlesize);
    \begin{scope}
      \clip (-\circlesize,-\circlesize)
        rectangle (0,\circlesize);
      \fill[black] (0,0) circle (\circlesize);
    \end{scope}
  }%
}

\newcommand{\ecircle}{%
  \tikz[baseline=-0.6ex]
    \draw[black,line width=0.4pt]
      (0,0) circle (\circlesize);%
}

\newcommand{\fullcov}{\fcircle}
\newcommand{\partcov}{\hcircle}
\newcommand{\nocov}{\ecircle}

\newtcolorbox{defensesummary}{
  enhanced,
  width=\dimexpr\linewidth-4pt\relax,
  colback=gray!5,
  colframe=teal!65!black,
  boxrule=0.6pt,
  leftrule=3pt,
  arc=1.5mm,
  left=7pt,
  right=7pt,
  top=7pt,
  bottom=6pt,
  before skip=7pt,
  after skip=7pt,
  title={Limitation},
  colbacktitle=teal!12,
  coltitle=black,
  fonttitle=\small\bfseries,
  boxed title style={
    boxrule=0pt,
    arc=0pt,
    outer arc=0pt
  },
  attach boxed title to top left={
    xshift=0pt,
    yshift=-\tcboxedtitleheight/2
  }
}

\fancypagestyle{landscapetable}{%
  \fancyhf{}
  \fancyfoot[C]{\thepage}

}

\AddToShipoutPictureBG{%
  \begin{tikzpicture}[remember picture,overlay]
    \node[
      rotate=45,
      scale=4,
      text=gray!60,
      opacity=0.12
    ] at (current page.center)
    {\textbf{EDUCATIONAL RELEASE}};
  \end{tikzpicture}%
}

\AddToShipoutPictureFG{%
  \begin{tikzpicture}[remember picture,overlay]
    \node[
      anchor=north east,
      xshift=-8mm,
      yshift=-7mm,
      text=red,
      font=\Large\bfseries
    ] at (current page.north east)
    {FOR EDUCATIONAL PURPOSES ONLY};
  \end{tikzpicture}%
}

\begin{document}

\title{SoK: Rethinking Jailbreaking in the Era of Agentic AI:
Attacks, Defenses, and Practical Considerations}

\author{Md Jueal Mia}
\email{mmia001@fiu.edu}
\affiliation{%
  \institution{Knight Foundation School of Computing and
  Information Sciences, Florida International University}
  \city{Miami}
  \state{Florida}
  \country{USA}
}
\affiliation{%
  \institution{Security, Optimization, and Learning for
  InterDependent Networks (solid) Laboratory}
  \city{Miami}
  \state{Florida}
  \country{USA}
}

\author{Yanzhao Wu}
\email{yawu@fiu.edu}
\affiliation{%
  \institution{Knight Foundation School of Computing and
  Information Sciences, Florida International University}
  \city{Miami}
  \state{Florida}
  \country{USA}
}

\author{Selcuk Uluagac}
\email{suluagac@fiu.edu}
\affiliation{%
  \institution{Knight Foundation School of Computing and
  Information Sciences, Florida International University}
  \city{Miami}
  \state{Florida}
  \country{USA}
}

\author{M. Hadi Amini}
\email{moamini@fiu.edu}
\affiliation{%
  \institution{Knight Foundation School of Computing and
  Information Sciences, Florida International University}
  \city{Miami}
  \state{Florida}
  \country{USA}
}
\affiliation{%
  \institution{Security, Optimization, and Learning for
  InterDependent Networks (solid) Laboratory}
  \city{Miami}
  \state{Florida}
  \country{USA}
}

\renewcommand{\shortauthors}{Mia et al.}


\renewcommand{\shortauthors}{Mia et al.}

\begin{abstract}
Large language models (LLMs) are rapidly evolving from conversational assistants
into agentic AI systems that reason, plan, invoke tools, maintain persistent
memory, communicate with other agents, and execute multi-step tasks. At the same
time, modern models exhibit substantially stronger native safety alignment than
earlier generations on which many jailbreak attacks and defenses were originally
studied. This shift raises a fundamental question: \textit{which established
jailbreak-security findings remain valid in the era of modern LLMs and agentic
AI?} We address this question through a Systematization of Knowledge (SoK) that
reframes jailbreak security around the full agentic execution pipeline. We
develop unified taxonomies of attacks and defenses spanning user interaction,
planning and reasoning, memory, tool use, and inter-agent communication, and
introduce a security--utility--efficiency evaluation framework that separates
native harmful-prompt safety, adversarial jailbreak robustness, and agent-level
security outcomes. We further conduct a controlled empirical study of
representative attacks and defenses within a common agentic framework. Our
results reveal three important gaps. First, strong native alignment does not
imply robustness to adversarial jailbreaks. Second, defense effectiveness is
highly model-, attack-, and component-dependent and can come at substantial
cost in over-refusal, utility, and latency. Third, low final-response attack
success can mask severe intermediate compromise: planning, memory, and tool
interactions may remain unsafe even when the final response is successfully
filtered. These findings motivate a shift from response-centric jailbreak
defense toward cross-layer, execution-aware security that protects agent state,
component transitions, and external actions while preserving practical utility
and efficiency.
\end{abstract}

\keywords{Agentic AI, Large Language Model Security, Jailbreak Attacks and Defenses, Execution-Aware Security, Systematization of Knowledge}

\maketitle

\section{Introduction}
\label{sec:introduction}

Large language models (LLMs) have demonstrated strong capabilities across text
generation, question answering, code generation, mathematical reasoning, and
scientific discovery
\cite{brown2020language}. Building on these
capabilities, LLMs are increasingly embedded in agentic AI systems that can
plan, reason, retrieve information, invoke external tools, maintain persistent
memory, coordinate with other agents, and execute multi-step tasks
\cite{sapkota2025ai,yao2022react,schick2023toolformer}. Unlike
conventional prompt--response applications, agentic systems operate through
iterative execution loops in which model outputs can influence subsequent
reasoning, stored state, tools, and other agents. This transition raises a
fundamental security question: \textit{which established findings about
jailbreak attacks and defenses still hold for modern LLMs and agentic AI?}

The answer is increasingly unclear because the security baseline itself has
changed. Modern instruction-tuned models are explicitly optimized to follow user
instructions while respecting behavioral and safety constraints
\cite{ouyang2022training, amini2025distributed}, and therefore often resist straightforward harmful
prompts more effectively than earlier models. Yet strong native alignment does
not imply jailbreak robustness: automated and optimization-based attacks
continue to bypass aligned models using carefully constructed adversarial
prompts \cite{zou2023universal,ICLR2024_f83cb637, 11471479}. Evaluating a defense only by
its defended attack success rate (ASR) can therefore be misleading. What
matters is the \textit{marginal security gain beyond native alignment}, how that
gain varies across attacks and models, and what utility and efficiency costs it
introduces.

Agentic AI makes this problem substantially harder by expanding both the entry
vectors and objectives of attacks beyond the user--model interface. We treat
the entry vector---direct user prompts, retrieved content, memory, tool outputs,
or inter-agent messages---as distinct from the objective, such as safety-policy
bypass, goal hijacking, data exfiltration, or unauthorized action. In this SoK,
a prompt injection is counted as a jailbreak only when it bypasses an explicit
model- or agent-level safety policy to induce prohibited generation, reasoning,
storage, tool use, or execution; injections that only redirect an authorized
task without such a bypass are outside our scope. Indirect content, persistent
memory, tools, and inter-agent communication therefore constitute
jailbreak-relevant trust boundaries through which adversarial instructions can
propagate \cite{debenedetti2024agentdojo,zhan2024injecagent,dash2026from,chu2026systematic}.
Consequently, \textit{response-level safety and agent-level security are not
equivalent}: a safe final response does not guarantee safe planning, memory,
tool interaction, or execution.

Practical deployment introduces a second challenge: security must remain
affordable. Small language models (SLMs) are increasingly attractive for
resource-constrained and specialized agentic workloads because of their lower
inference cost, reduced memory requirements, and suitability for local or edge
deployment \cite{erdogan2024tinyagent}. Recent work further
argues that SLMs can handle repetitive and specialized invocations within
heterogeneous agent architectures, reserving larger models for more demanding
tasks \cite{belcak2025small}. However, many jailbreak defenses require repeated
inference, prompt perturbation, auxiliary safety models, or additional
generation \cite{li2025securitylingua}. In an agentic execution loop, such
overhead may recur across planning, tool use, verification, and inter-agent
steps, making security--utility--efficiency trade-offs a first-order design
concern.

These developments expose a gap between traditional jailbreak evaluation and
the security requirements of modern agentic AI. Existing jailbreak research
largely asks whether adversarial prompts bypass model-level safeguards, whereas
agent security must also reason about planning, tools, memory, persistent state,
and autonomous execution. At the same time, stronger native alignment changes
the baseline against which defenses should be judged. A meaningful evaluation
must therefore distinguish \textit{harmful-prompt safety}, \textit{adversarial
jailbreak robustness}, and \textit{agent-level security}, while jointly
considering over-refusal, task utility, latency, and computational cost.

Accordingly, this SoK systematizes jailbreak attacks and defenses across the
agentic execution pipeline and uses a controlled empirical study to re-examine
representative inference-time defenses under modern LLMs. We address the
following research questions:

\begin{itemize}

    \item \textbf{RQ1:} How do agentic capabilities expand the jailbreak attack
    surface beyond traditional conversational LLMs?

    \item \textbf{RQ2:} How can jailbreak attacks and defenses be systematically
    characterized across agent components, trust boundaries, and execution
    stages?

    \item \textbf{RQ3:} What marginal security benefits and
    security--utility--efficiency trade-offs arise when jailbreak defenses are
    deployed with modern LLMs in agentic settings?

    \item \textbf{RQ4:} Which established jailbreak-security conclusions remain
    valid, which require re-evaluation, and what challenges remain for securing
    modern agentic AI?

\end{itemize}

Motivated by these questions, we have the following contributions:

\begin{itemize}

    \item We present a \textbf{multidimensional taxonomy of jailbreak attacks
    and defenses for agentic AI}, covering adversarial access, injection
    channel, target component, persistence, interaction structure, optimization
    mechanism, security property, trust boundary, and defense stage.

    \item We introduce a \textbf{security--utility--efficiency evaluation
    framework} that separates harmful-prompt safety, adversarial jailbreak
    robustness, and agent-level security while accounting for over-refusal,
    response quality, task success, latency, and computational cost.

    \item We conduct a \textbf{controlled empirical evaluation} of
    representative inference-time defenses using Qwen3.5-4B and
    Llama-3.1-8B-Instruct, revealing substantial model-, attack-, and
    component-dependent differences in robustness, utility, and efficiency.

    \item We identify \textbf{limitations in current jailbreak defenses and
    evaluation practices}, expose gaps between response-level safety and
    agent-level protection, and outline directions for adaptive,
    execution-aware, and resource-efficient defenses.

\end{itemize}

Our scope separates \textit{systematization breadth} from \textit{empirical
depth}. The taxonomy covers jailbreak-relevant attack and defense surfaces
across planning, tools, memory, and multi-agent interaction, whereas the
controlled experiments focus on representative model-layer inference-time
defenses under a common setting. The goal is not to claim that these defenses
provide complete agent-level protection, but to establish where existing
model-layer safeguards remain effective, where their benefits diminish under
modern native alignment, and where system-level protection becomes necessary.
\section{Related Work}
\label{sec:related}

\subsection{Literature Search}
\label{sec:literature_search}

We conducted a multi-source literature search, last updated in August 2026,
covering work published or publicly available from 2022--2026. We searched
Google Scholar, ACM Digital Library, IEEE Xplore, ACL Anthology, arXiv, and
OpenReview, together with major AI, NLP, and security venues. Search queries
combined terms such as \textit{``jailbreak attack''}, \textit{``jailbreak
defense''}, \textit{``LLM jailbreak''}, \textit{``agentic AI''}, and
\textit{``LLM agent''} with agent-specific terms related to planning and
reasoning, chain-of-thought, tool use and poisoning, memory injection and
poisoning, retrieval and RAG, persistent state, and multi-agent interaction.
We also performed backward and forward citation tracing from relevant surveys,
benchmarks, and influential jailbreak studies. This search identified \textbf{271 attack-related} and \textbf{88 defense-related}
candidate studies or methods for screening.

We manually screened the candidates for technical relevance, removed duplicate
and out-of-scope works, and preferred published versions over corresponding
preprints. Studies were retained when they introduced, analyzed, evaluated,
benchmarked, or defended against jailbreak behavior at the model or agent
level. The retained works were systematically coded by adversarial access,
attack surface, target component, persistence, interaction structure,
optimization mechanism, trust boundary, defense stage, and evaluation
characteristics. The resulting taxonomy contains \textbf{116 distinct jailbreak
attacks and 63 distinct defenses}, spanning traditional model-level techniques
and emerging agentic attack surfaces.

\subsection{Existing Jailbreak and Agentic AI Studies}

Jailbreak research initially focused on standalone LLMs, where adversaries
attempt to bypass safety alignment through crafted prompts, optimization-based
suffixes, role-playing, prompt transformations, and multi-turn interactions.
Corresponding defenses span safety training, input filtering and transformation,
decoding-time intervention, internal-state monitoring, and output verification.
Xu \emph{et al.} systematize this literature through taxonomies of jailbreak
attacks, defenses, datasets, evaluation methodologies, and automated judges,
and introduce the Security Cube for multidimensional robustness evaluation
\cite{11573543}. Wang \emph{et al.} further systematize jailbreak guardrails
and propose a security--efficiency--utility framework for comparing defenses
across attacks \cite{11573588}. These studies provide comprehensive foundations
for jailbreak security, but their analysis is primarily centered on standalone
LLMs and prompt--response safety rather than autonomous agent execution.

In parallel, the emergence of agentic AI has shifted security research toward
systems that reason, maintain state, invoke tools, interact with external
environments, and collaborate with other agents. Shahriar \emph{et al.}
organize this space around agentic applications, threats, and defenses
\cite{shahriar2025survey}, while Chhabra \emph{et al.} survey threats including
prompt injection, jailbreaks, tool abuse, autonomous cyber-exploitation,
multi-agent attacks, and interface vulnerabilities \cite{11447227}.
Dehghantanha and Homayoun emphasize trust boundaries spanning users, retrieval,
memory, tools, external APIs, and multi-agent communication
\cite{dehghantanha2026sok}. Chu develops a layered attack-surface model that
captures foundation, cognitive, memory, tool-execution, multi-agent, ecosystem,
and governance layers, including temporally persistent attacks
\cite{chu2026systematic}. Kim \emph{et al.} similarly analyze how security
emerges from interactions among models, memory, tools, workflows, external
environments, and conventional software components \cite{kim2026sok}.

The closest prior SoKs differ from our work along four concrete dimensions.
Xu \emph{et al.}~\cite{11573543} systematize model-level jailbreak attacks,
defenses, datasets, and judges, whereas Wang \emph{et al.}~\cite{11573588}
focus on jailbreak guardrails and their security--utility--efficiency
trade-offs. In both cases, the primary unit of analysis is a standalone
model interaction, and safety is evaluated mainly from the final response.
Kim \emph{et al.}~\cite{kim2026sok} instead adopt a system-level view of
agentic security, covering interactions among models, memory, tools,
workflows, environments, and conventional software; however, jailbreak is
treated as one threat class within the broader agentic-security landscape,
without a common empirical evaluation of jailbreak defenses across agent
layers. Our SoK connects these scopes by organizing jailbreak attacks and
defenses around the agentic execution pipeline and using the agent trajectory,
rather than only the prompt--response pair, as the unit of analysis.

The resulting empirical contribution is also distinct. We evaluate
representative defenses under a shared implementation using two final models,
nine jailbreak attacks, benign and agentic benchmarks, and common judging
procedures. In addition to final-response ASR, we measure unsafe behavior in
planning, reasoning, memory, and tool interaction, together with over-refusal,
response quality, task utility, latency, and computational cost. This design
provides evidence unavailable from the closest SoKs: a defense can suppress
the final unsafe response while leaving intermediate agent components
compromised, and stronger response-level security can introduce substantial
utility or efficiency costs. Thus, the paper's delta is not a general claim
of being the first agent-security survey, but a jailbreak-specific taxonomy
and controlled cross-layer evaluation that links component-level failures to
final safety and practical trade-offs.

Table~\ref{tab:related-work-comparison} summarizes these differences in
taxonomy scope, unit of analysis, measured outcomes, and empirical design.
The comparison is descriptive rather than a ranking. We mark an aspect as
\emph{comprehensive} when it is supported by a dedicated taxonomy, framework,
or empirical study; \emph{partial} when it is discussed but is not a central
organizing or evaluated component; and \emph{not covered} when we found no
substantive treatment. In particular, \emph{Agent-Level Outcomes} requires
measurement of intermediate or execution-level behavior rather than only
discussion of agent components, while \emph{Empirical Evaluation} denotes a
controlled comparison of multiple attacks or defenses under a common
experimental framework.

\begin{table}[t]
\centering
\caption{Comparison of existing surveys and SoK papers. A filled circle indicates comprehensive coverage, a half-filled circle indicates partial coverage, and an empty circle indicates no coverage.}
\label{tab:related-work-comparison}

\scriptsize
\renewcommand{\arraystretch}{1.15}
\setlength{\tabcolsep}{1.5pt}

\resizebox{\columnwidth}{!}{%
\begin{tabular}{lccccccccc}
\toprule

\textbf{Reference}
&
\shortstack{\textbf{Jailbreak}\\\textbf{Focus}}
&
\shortstack{\textbf{Agentic}\\\textbf{Focus}}
&
\shortstack{\textbf{Agentic}\\\textbf{Surface}}
&
\shortstack{\textbf{Threat}\\\textbf{Model}}
&
\shortstack{\textbf{Attack}\\\textbf{Taxonomy}}
&
\shortstack{\textbf{Defense}\\\textbf{Taxonomy}}
&
\shortstack{\textbf{Agent-Level}\\\textbf{Outcomes}}
&
\shortstack{\textbf{Security--Utility--}\\\textbf{Efficiency}}
&
\shortstack{\textbf{Empirical}\\\textbf{Evaluation}}
\\

\midrule

\cite{11573543}
&
\fullcov
&
\nocov
&
\nocov
&
\fullcov
&
\fullcov
&
\fullcov
&
\nocov
&
\partcov
&
\partcov
\\

\cite{11573588}
&
\fullcov
&
\nocov
&
\nocov
&
\fullcov
&
\nocov
&
\fullcov
&
\nocov
&
\partcov
&
\partcov
\\

\cite{shahriar2025survey}
&
\partcov
&
\fullcov
&
\fullcov
&
\fullcov
&
\partcov
&
\partcov
&
\nocov
&
\nocov
&
\nocov
\\

\cite{11447227}
&
\partcov
&
\fullcov
&
\fullcov
&
\fullcov
&
\fullcov
&
\fullcov
&
\nocov
&
\nocov
&
\nocov
\\

\cite{dehghantanha2026sok}
&
\partcov
&
\fullcov
&
\fullcov
&
\fullcov
&
\fullcov
&
\partcov
&
\nocov
&
\nocov
&
\nocov
\\

\cite{chu2026systematic}
&
\partcov
&
\fullcov
&
\partcov
&
\partcov
&
\partcov
&
\partcov
&
\nocov
&
\nocov
&
\nocov
\\

\cite{kim2026sok}
&
\partcov
&
\fullcov
&
\partcov
&
\partcov
&
\fullcov
&
\fullcov
&
\nocov
&
\nocov
&
\nocov
\\

\midrule

\textbf{This SoK}
&
\fullcov
&
\fullcov
&
\fullcov
&
\fullcov
&
\fullcov
&
\fullcov
&
\fullcov
&
\fullcov
&
\fullcov
\\

\bottomrule
\end{tabular}
}
\end{table}
\section{Preliminaries}

\subsection{Agentic AI Systems}

Agentic AI extends traditional AI toward autonomous, goal-directed systems that
can adapt to changing environments, make context-dependent decisions, and pursue
multi-step objectives with limited human intervention
\cite{acharya2025agentic, hosseini2025role}. Modern agentic systems may further
support task decomposition, persistent memory, tool use, and multi-agent
collaboration, enabling coordinated execution of complex tasks across diverse
domains \cite{sapkota2025ai}.

\subsection{Agent Execution Pipeline}

\begin{figure}[t]
    \centering
    \includegraphics[width=0.80\columnwidth]{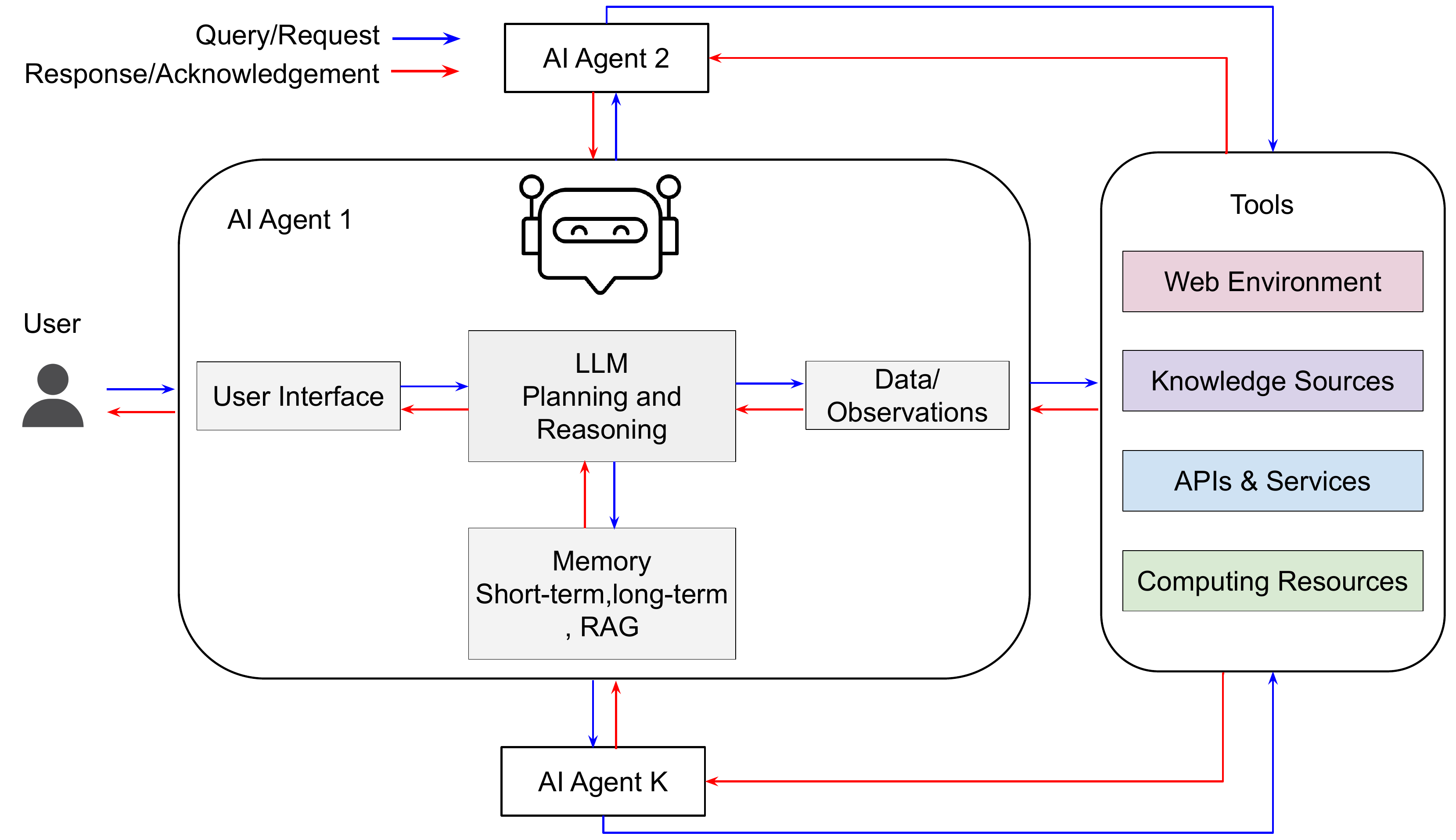}
    \caption{Agentic AI execution pipeline.}
    \label{fig:agentic_execution}
\end{figure}

Unlike conventional conversational LLMs that primarily produce a single
response, agentic systems operate through iterative reasoning--action loops
involving planning, retrieval, memory, tool interaction, observation, and
execution \cite{acharya2025agentic,sapkota2025ai}. As illustrated in
Figure~\ref{fig:agentic_execution}, an agent interprets a user objective,
decomposes it into subtasks, retrieves relevant information, invokes tools, and
uses resulting observations to guide subsequent decisions. Multi-agent systems
may additionally exchange information and coordinate actions. Because state and
outputs are propagated across these stages, each component and transition can
become a potential target for adversarial manipulation.


\subsection{Agentic Threat Model}

Jailbreak attacks aim to bypass safety alignment and induce prohibited behavior
~\cite{NEURIPS2023_fd661313,zou2023universal}, using techniques such as prompt
manipulation, role-playing, adversarial optimization, and multi-turn interaction
~\cite{10992337,yu2023gptfuzzer,307992}. In agentic systems, such attacks can
affect not only final responses but also planning, memory, retrieval, tool use,
and inter-agent communication.

We primarily consider a \emph{black-box adversary} that can control user inputs
and externally supplied content but cannot modify trusted system prompts, model
parameters, or infrastructure. We also consider a stronger \emph{white-box
adversary} with access to model internals or defense information. We distinguish
\emph{final-response jailbreaks} from \emph{component-level unsafe behavior},
where intermediate agent components may become unsafe even when the final
response remains safe.

\section{Agentic Taxonomy}
\label{agentic_taxonomy}

\subsection{Taxonomy of Agentic Jailbreak Attacks}
\label{sec:agentic-jailbreak-attacks}

\usetikzlibrary{positioning,calc}

\begin{figure*}[t]
\centering

\resizebox{0.96\textwidth}{!}{%
\begin{tikzpicture}[
    root/.style={
        draw=black,
        fill=white,
        rounded corners=2pt,
        text width=2.10cm,
        minimum height=0.90cm,
        align=center,
        inner xsep=4pt,
        inner ysep=3pt,
        font=\footnotesize\bfseries
    },
    threat/.style={
        draw=black,
        fill=white,
        rounded corners=2pt,
        text width=2.25cm,
        minimum height=0.72cm,
        align=center,
        inner xsep=4pt,
        inner ysep=3pt,
        font=\footnotesize\bfseries
    },
    category/.style={
        draw=black,
        fill=white,
        rounded corners=2pt,
        text width=3.20cm,
        minimum height=0.68cm,
        align=center,
        inner xsep=4pt,
        inner ysep=3pt,
        font=\footnotesize\bfseries
    },
    methods/.style={
        draw=black,
        fill=white,
        rounded corners=2pt,
        text width=8.6cm,
        minimum height=0.60cm,
        align=left,
        inner xsep=5pt,
        inner ysep=3pt,
        font=\fontsize{7.0}{7.8}\selectfont
    },
    conn/.style={
        draw=black,
        line width=0.65pt
    }
]



\node[root] (root) at (0.50,0.30)
{Agentic\\Jailbreak\\Attack Taxonomy};

\coordinate (root-split) at (1.75,0.30);


\node[threat] (blackbox) at (3.40,2.80)
{Black-Box\\Attacks};

\node[threat] (whitebox) at (3.40,-2.25)
{White-Box\\Attacks};

\coordinate (black-split) at (5.00,2.80);
\coordinate (white-split) at (5.00,-2.25);

\node[category] (bb-user) at (7.20,6.50)
{User--Agent Interaction};

\node[category] (bb-reasoning) at (7.20,3.75)
{Planning \& Reasoning};

\node[category] (bb-tool) at (7.20,2.65)
{Agent--Tool Interaction};

\node[category] (bb-memory) at (7.20,1.55)
{Agent--Memory Interaction};

\node[category] (bb-agent) at (7.20,0.45)
{Agent--Agent Interaction};


\node[
    methods,
    text width=8.6cm,
    font=\fontsize{6.6}{7.25}\selectfont
] (bb-user-m) at (14.20,6.50)
{
PAIR~\cite{10992337},
TAP~\cite{NEURIPS2024_70702e8c},
X-Teaming~\cite{rahman2025x},
ActorAttack~\cite{ren2024derail},
MRJ-Agent~\cite{wang2024mrj},
JOOD~\cite{Jeong_2025_CVPR},
Crescendo~\cite{307992},
SwordEcho~\cite{tang2024swordecho}, BADROBOT~\cite{zhang2025badrobot},
FITD~\cite{weng-etal-2025-foot}, PathSeeker~\cite{lin2024pathseeker}, LLM-Virus~\cite{yu2024llm}, RedAgent~\cite{xu2026redagent}, LASH~\cite{nafi2026lash}, TriageFuzz~\cite{chen2026not}, Safe2Harm~\cite{yang2025safe2harm}, GTA~\cite{sun2025survive}, JAIL-CON~\cite{NEURIPS2025_d80f5334}, Jailbreak Mimicry~\cite{ntais2025jailbreak}, MetaBreak~\cite{11573507}, Response Attack~\cite{ziqi2026response}, PiF~\cite{ICLR2025_db988b08}, GASP~\cite{NEURIPS2025_d78ca219}, LLM STINGER~\cite{jha2025llm}, SequentialBreak~\cite{saiem-etal-2025-sequentialbreak}, Ensemble Black-box Jailbreak~\cite{11452136}, ShadowBreak~\cite{mu-etal-2025-stealthy}, Bijection Learning~\cite{ICLR2025_b05c1fb3}, AutoDAN-Turbo~\cite{ICLR2025_1bff3663}, AdaPPA~\cite{10890715}, PAPILLON~\cite{307744}, Puzzler~\cite{chang-etal-2024-play}, h4rm3l~\cite{ICLR2025_904aac1c}, Single Character Perturbation~\cite{lin2025single}, Virtual Context~\cite{zhou-etal-2024-virtual}, SEQAR~\cite{yang-etal-2025-seqar}, I-FSJ~\cite{NEURIPS2024_39a3aa9d}, RLbreaker~\cite{NEURIPS2024_2f148634}, ArtPrompt~\cite{jiang-etal-2024-artprompt}, CodeAttack~\cite{ren-etal-2024-codeattack}, DRA~\cite{299784}, Adaptive Attacks~\cite{ICLR2025_63fa7efd}, ReNeLLM~\cite{ding-etal-2024-wolf}, Multilingual Jailbreak~\cite{ICLR2024_6b396f76}, DeepInception~\cite{li2024deepinception}, IRIS~\cite{ramesh-etal-2024-gpt}, JailbreakHub~\cite{10.1145/3658644.3670388}, CipherChat~\cite{ICLR2024_ed4c38fe}, JAM~\cite{NEURIPS2024_6d56bc83},  Many-shot Jailbreaking (MSJ)~\cite{NEURIPS2024_ea456e23}, FuzzLLM~\cite{10448041}, DAP~\cite{xiao-etal-2024-distract}, DrAttack~\cite{li-etal-2024-drattack}, CodeChameleon~\cite{lv2024codechameleon}, ACE/LACE~\cite{handa2025when}, ICA~\cite{11370531}, GCQ~\cite{NEURIPS2024_e7b3dd85}, Low-Resource Language~\cite{yong2023lowresource}, PAP~\cite{zeng-etal-2024-johnny}, HackAPrompt~\cite{schulhoff-etal-2023-ignore}, Persona Modulation~\cite{shah2023scalable}, MJP~\cite{li-etal-2023-multi-step},  Open Sesame~\cite{app14167150}, SIJ~\cite{zhao-etal-2025-sql}, AdvPrompter~\cite{3780338.3782273}, AgenticRed~\cite{yuan2026agenticred}, SEMA~\cite{feng2026sema}, PE-CoA~\cite{nihal-etal-2026-pattern}, NEXUS~\cite{rafiei-asl-etal-2025-nexus}, GALA~\cite{chen2025strategize}, GOAT~\cite{3780338.3782274}, Red Queen Attack~\cite{jiang-etal-2025-red}, and
CoA~\cite{yang2024chain}.
};

\node[methods] (bb-reasoning-m) at (14.20,3.75)
{
TRACE~\cite{zeng2026trace},
A Mousetrap~\cite{yao-etal-2025-mousetrap},
Chain-of-Thought Hijacking~\cite{zhao2025chain},
TROJail~\cite{xiong-etal-2026-trojail},
Jigsaw Puzzles~\cite{yang2024jigsaw},
RACE~\cite{ying-etal-2025-reasoning}, AE-CoT~\cite{li2026reasoning}, H-CoT~\cite{kuo2025h}, ABJ~\cite{lin2024llms}, Cognitive Overload~\cite{xu-etal-2024-cognitive}, and
Pandora~\cite{chen2024pandora}.
};

\node[methods] (bb-tool-m) at (14.20,2.65)
{
AgentHarm~\cite{ICLR2025_c493d23a} and
STAC~\cite{li2025stac}.
};

\node[methods] (bb-memory-m) at (14.20,1.55)
{
MINJA~\cite{NEURIPS2025_42a97bbd}.
};

\node[methods] (bb-agent-m) at (14.20,0.45)
{
ARCJ~\cite{men-etal-2025-troublemaker}, Evil Geniuses~\cite{tian2023evil},
Agent Smith~\cite{gu2024agent},
MAD~\cite{qi2025amplified}, PRP~\cite{mangaokar-etal-2024-prp}, and
The Dark Side of LLMs~\cite{lupinacci2025dark}.
};


\node[category] (wb-gradient) at (7.20,-1.15)
{Gradient-Based\\Optimization};

\node[category] (wb-search) at (7.20,-2.25)
{Search-Based\\Optimization};

\node[category] (wb-representation) at (7.20,-3.35)
{Representation-Aware\\Optimization};


\node[methods] (wb-gradient-m) at (14.20,-1.15) 
{
GCG~\cite{zou2023universal},
Faster-GCG~\cite{li2024faster}, DeGCG~\cite{liu-etal-2024-advancing-adversarial}, ASETF~\cite{wang-etal-2024-asetf}, NTA~\cite{huang2025nontextual}, DSN~\cite{zhou-etal-2025-dont}, Guided Jailbreak Attack~\cite{yang-etal-2025-guiding}, I-GCG~\cite{ICLR2025_124256ed}, AmpleGCG~\cite{liao2024amplegcg}, AmpleGCG-Plus~\cite{kumar2024amplegcg}, COLD-Attack~\cite{guo2024coldattack}, ADC~\cite{NEURIPS2024_29571f8f}, MAC~\cite{10888812}, PGD~\cite{geisler2024attacking}, and
AttnGCG~\cite{wang2024attngcg}.
};

\node[methods] (wb-search-m) at (14.20,-2.25)
{
AutoDAN~\cite{ICLR2024_f83cb637},
BEAST~\cite{sadasivan2024fast}, Functional Homotopy (FH-GR)~\cite{ICLR2025_5ed21295}, ADV-LLM~\cite{sun-etal-2025-iterative}, and
MAGIC~\cite{li-etal-2025-exploiting}.
};

\node[methods] (wb-representation-m) at (14.20,-3.35)
{
PoEx~\cite{lu2024poex}, SABER~\cite{joshi-etal-2025-saber}, ORTHO~\cite{NEURIPS2024_f5454485}, Safety Patterns~\cite{li-etal-2025-revisiting}, and ASJA~\cite{du2025multi}.
};

\draw[conn]
(root.east) -- (root-split);

\draw[conn]
(root-split) |- (blackbox.west);

\draw[conn]
(root-split) |- (whitebox.west);


\draw[conn]
(blackbox.east) -- (black-split);

\draw[conn]
(black-split) |- (bb-user.west);

\draw[conn]
(black-split) |- (bb-reasoning.west);

\draw[conn]
(black-split) |- (bb-tool.west);

\draw[conn]
(black-split) |- (bb-memory.west);

\draw[conn]
(black-split) |- (bb-agent.west);

\draw[conn]
(bb-user.east) -- (bb-user-m.west);

\draw[conn]
(bb-reasoning.east) -- (bb-reasoning-m.west);

\draw[conn]
(bb-tool.east) -- (bb-tool-m.west);

\draw[conn]
(bb-memory.east) -- (bb-memory-m.west);

\draw[conn]
(bb-agent.east) -- (bb-agent-m.west);


\draw[conn]
(whitebox.east) -- (white-split);

\draw[conn]
(white-split) |- (wb-gradient.west);

\draw[conn]
(white-split) |- (wb-search.west);

\draw[conn]
(white-split) |- (wb-representation.west);

\draw[conn]
(wb-gradient.east) -- (wb-gradient-m.west);

\draw[conn]
(wb-search.east) -- (wb-search-m.west);

\draw[conn]
(wb-representation.east) -- (wb-representation-m.west);

\end{tikzpicture}%
}

\caption{
Taxonomy of jailbreak attacks in agentic AI systems.
}
\label{fig:agentic-jailbreak-taxonomy}

\end{figure*}
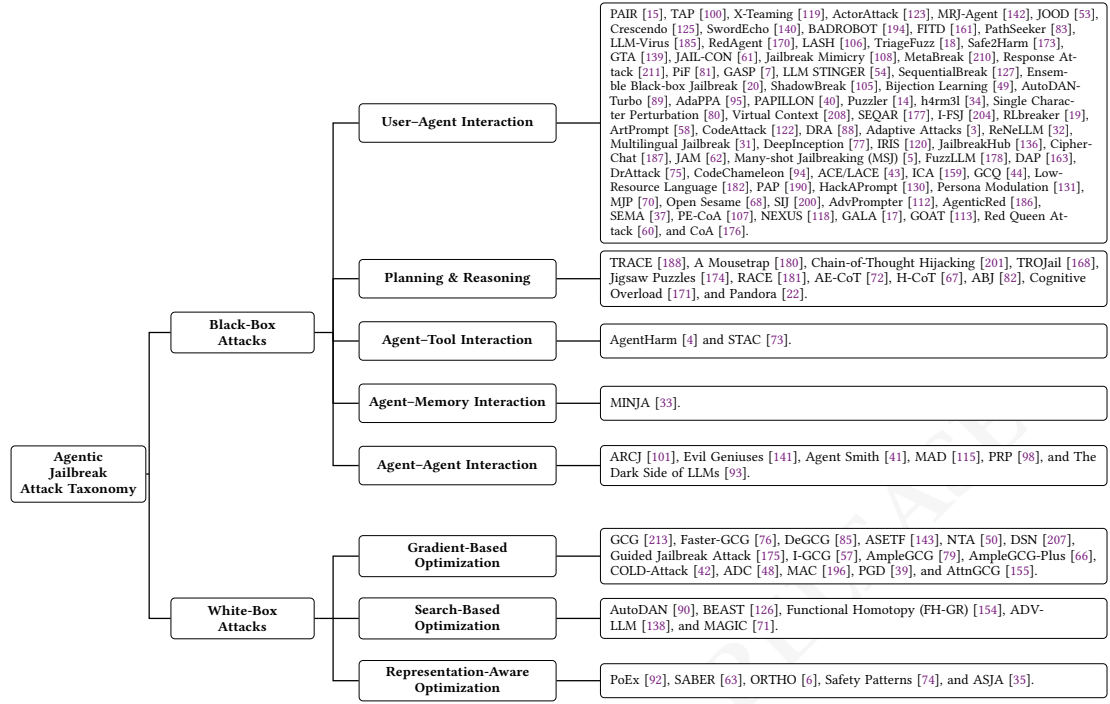

Agentic systems expand the jailbreak attack surface beyond a single
user--model interaction. Adversarial inputs may enter through the user
interface or influence intermediate reasoning, tool use, persistent memory,
and communication among agents. Fig.~\ref{fig:agentic-jailbreak-taxonomy}
organizes these attacks along two primary dimensions. We first distinguish
\emph{black-box} and \emph{white-box} access. Black-box attacks are organized
by the interaction surface they target, whereas white-box attacks are grouped
by their primary optimization mechanism. The discussion below focuses on the
characteristics that distinguish these attack families.

\paragraph{Black-Box Attacks.}
Black-box attacks operate without direct access to the target model's
parameters, gradients, logits, or hidden representations. Instead, attackers
interact through externally accessible interfaces and may use auxiliary
attacker or surrogate models. We categorize these attacks according to
\emph{user--agent}, \emph{planning and reasoning}, \emph{agent--tool},
\emph{agent--memory}, and \emph{agent--agent} interactions.

\paragraph{User--Agent Interaction.}
User--agent attacks manipulate the user-facing input channel through prompt
refinement, semantic concealment, contextual manipulation, or automated
generation. PAIR~\cite{10992337} iteratively refines jailbreak prompts, whereas
TAP~\cite{NEURIPS2024_70702e8c} uses tree-based prompt optimization with
pruning. ArtPrompt~\cite{jiang-etal-2024-artprompt} masks sensitive terms using
ASCII art, while CodeAttack~\cite{ren-etal-2024-codeattack} reformulates
harmful requests as code-completion tasks. More automated approaches include
AutoDAN-Turbo~\cite{ICLR2025_1bff3663}, which discovers and retrieves attack
strategies across repeated interactions, and PAPILLON~\cite{307744}, which
uses fuzz testing to mutate short jailbreak prompts.

\begin{defensesummary}
\small

\textit{Common limitations include high query or computational cost, reliance
on auxiliary models or target-specific information, restricted evaluation
across models and languages, and limited transferability to strongly aligned
models.}
\end{defensesummary}

\paragraph{Planning and Reasoning.}
Planning and reasoning attacks target how harmful objectives are decomposed
or processed during reasoning. TRACE~\cite{zeng2026trace} combines task
decomposition with feedback-driven self-evolution, while Chain-of-Thought
Hijacking~\cite{zhao2025chain} uses prolonged benign reasoning to weaken
refusal behavior. TROJail~\cite{xiong-etal-2026-trojail} optimizes multi-turn
trajectories using outcome and process rewards, and Jigsaw
Puzzles~\cite{yang2024jigsaw} progressively splits and reconstructs harmful
queries. These methods exploit the reasoning trajectory rather than relying
only on direct prompt reformulation.

\begin{defensesummary}
\small

\textit{Evaluations are often restricted to controlled settings or specific
reasoning models, while several methods incur substantial interaction or
search overhead and provide limited evaluation against defenses.}
\end{defensesummary}

\paragraph{Agent--Tool Interaction.}
Agent--tool attacks target systems in which model decisions can trigger
external actions. AgentHarm~\cite{ICLR2025_c493d23a} evaluates whether agents
can complete harmful multi-step tasks using custom tools under jailbreak
prompts. STAC~\cite{li2025stac} constructs sequential attack chains composed
of executable tool operations. Consequently, successful attacks at this
boundary can extend beyond harmful text to unsafe task execution.

\begin{defensesummary}
\small

\textit{Current evaluations cover a limited range of agents and tools.
AgentHarm is restricted to English, excludes multi-turn attacks, and relies on
custom tools, while STAC has limited evaluation across diverse tool and agent
settings.}
\end{defensesummary}

\paragraph{Agent--Memory Interaction.}
Memory creates a persistent attack surface because adversarial information
introduced during one interaction can influence later behavior.
MINJA~\cite{NEURIPS2025_42a97bbd} performs query-only memory injection using
bridging steps and a Progressive Shortening Strategy (PSS). The injected
content is stored in memory and later retrieved into the agent's context,
separating the initial attack from its eventual effect.

\begin{defensesummary}
\small

\textit{Evidence for memory-targeted jailbreaks remains limited, and MINJA
depends on successful retrieval of the injected memory during subsequent
interactions.}
\end{defensesummary}

\paragraph{Agent--Agent Interaction.}
Multi-agent communication enables adversarial behavior to propagate beyond
the initially compromised component. ARCJ~\cite{men-etal-2025-troublemaker}
uses optimized retrieval and replication suffixes for contagious memory
poisoning, whereas Agent Smith~\cite{gu2024agent} propagates an infectious
jailbreak through inter-agent communication. MAD~\cite{qi2025amplified}
exploits repeated multi-agent debate to amplify attacks, and
PRP~\cite{mangaokar-etal-2024-prp} propagates a universal perturbation through
an LLM-based guardrail pipeline.

\begin{defensesummary}
\small

\textit{Existing attacks rely on specific system assumptions, including
explicit memory retrieval, randomized agent interactions, repeated debate,
particular guardrail architectures, or executable tool access, limiting their
generality across multi-agent systems.}
\end{defensesummary}

\paragraph{White-Box Attacks.}
White-box attacks assume privileged access to internal model information,
including parameters, gradients, logits, attention scores, or hidden
representations. We distinguish \emph{gradient-based}, \emph{search-based},
and \emph{representation-aware} optimization according to the primary
mechanism used to construct adversarial inputs or modify safety-related
behavior.

\paragraph{Gradient-Based Optimization.}
Gradient-based methods use differentiable signals to optimize adversarial
inputs. GCG~\cite{zou2023universal} performs greedy coordinate gradient
optimization of universal adversarial suffixes, while
Faster-GCG~\cite{li2024faster} augments this process with distance
regularization, temperature-controlled sampling, and visited-suffix marking.
I-GCG~\cite{ICLR2025_124256ed} uses harmful targets, multi-coordinate updates,
and suffix initialization, whereas AttnGCG~\cite{wang2024attngcg} incorporates
attention into the coordinate-gradient objective.

\begin{defensesummary}
\small

\textit{These methods generally require white-box gradient or parameter access
and costly iterative optimization. Transferability can also decrease across
different architectures, tokenizers, and strongly aligned or closed-source
models.}
\end{defensesummary}

\paragraph{Search-Based Optimization.}
Search-based attacks explore candidate jailbreak prompts using genetic, beam,
random, or hybrid optimization procedures.
AutoDAN~\cite{ICLR2024_f83cb637} uses hierarchical genetic optimization to
generate semantically meaningful jailbreak prompts, while
BEAST~\cite{sadasivan2024fast} employs gradient-free beam search.
Functional Homotopy~\cite{ICLR2025_5ed21295} transfers random-search
optimization through progressively stronger checkpoints, and
ADV-LLM~\cite{sun-etal-2025-iterative} iteratively self-tunes an adversarial
suffix generator.

\begin{defensesummary}
\small

\textit{Search-based attacks can require many model evaluations and substantial
optimization cost. Their effectiveness and transferability may also depend on
the target architecture, surrogate model, or availability of intermediate
model checkpoints.}
\end{defensesummary}

\paragraph{Representation-Aware Optimization.}
Representation-aware attacks exploit internal features associated with
reasoning or safety behavior. PoEx~\cite{lu2024poex} performs hidden-layer
gradient optimization to generate executable robot policies, while
SABER~\cite{joshi-etal-2025-saber} injects optimized residuals across model
layers. ORTHO~\cite{NEURIPS2024_f5454485} identifies and removes a
refusal-mediating activation direction, whereas Safety
Patterns~\cite{li-etal-2025-revisiting} extracts and weakens safety-related
activation patterns. ASJA~\cite{du2025multi} instead uses attention scores to
guide the construction of adversarial dialogue histories.

\begin{defensesummary}
\small

\textit{These attacks require white-box access to internal representations or
attention signals. Additional limitations include expensive optimization,
sensitivity to feature or layer selection, possible degradation of reasoning
quality, and weaker transfer to closed-source models.}
\end{defensesummary}

\subsection{Agentic Jailbreak Defense Taxonomy}
\label{sec:agentic-jailbreak-defenses}

\usetikzlibrary{positioning,calc}

\begin{figure*}[t]
\centering

\resizebox{0.96\textwidth}{!}{%
\begin{tikzpicture}[
    root/.style={
        draw=black,
        fill=white,
        rounded corners=2pt,
        text width=2.20cm,
        minimum height=1.00cm,
        align=center,
        inner sep=4pt,
        font=\footnotesize\bfseries
    },
    layer/.style={
        draw=black,
        fill=white,
        rounded corners=2pt,
        text width=2.70cm,
        minimum height=0.78cm,
        align=center,
        inner sep=4pt,
        font=\footnotesize\bfseries
    },
    category/.style={
        draw=black,
        fill=white,
        rounded corners=2pt,
        text width=3.00cm,
        minimum height=0.72cm,
        align=center,
        inner sep=4pt,
        font=\scriptsize\bfseries
    },
    content/.style={
        draw=black,
        fill=white,
        rounded corners=2pt,
        text width=8.40cm,
        minimum height=0.72cm,
        align=left,
        inner xsep=5pt,
        inner ysep=3pt,
        font=\fontsize{7.0}{8.2}\selectfont
    },
    conn/.style={
        draw=black,
        line width=0.65pt
    }
]


\node[content] (training-m) at (13.50,0)
{
Safe RLHF~\cite{ICLR2024_dd1577af},
Safety-Tuned LLaMAs~\cite{ICLR2024_9178ece8},
Safety Alignment~\cite{ICLR2025_88be0230},
Goal-Conditioned DPO (GC-DPO)~\cite{maeng-etal-2025-goal},
ReFAT~\cite{ICLR2025_1022661f},
Safe Unlearning~\cite{zhang2024safe},
Layer-Specific Editing (LED)~\cite{zhao-etal-2024-defending-large},
Detoxifying with Intraoperative Neural Monitoring
(DINM)~\cite{wang-etal-2024-detoxifying},
DELMAN~\cite{wang-etal-2025-delman},
SafeLoRA~\cite{NEURIPS2024_77baa7c2},
CoG~\cite{mao-etal-2026-models},
MetaDefense~\cite{NEURIPS2025_8ec07851},
ReSA~\cite{ICLR2026_4f421415},
Eraser~\cite{lu2024eraser},
Adversarial Tuning~\cite{liu2024adversarial},
BackdoorAlign~\cite{NEURIPS2024_094324f3},
R2D~\cite{zhu-etal-2025-reasoning}, and
Circuit Breakers via Representation Rerouting
(RR)~\cite{NEURIPS2024_97ca7168}.
};

\node[content, below=2.5mm of training-m] (input-m)
{
Self-Reminder~\cite{xie2023defending},
In-Context Defense (ICD)~\cite{11370531},
In-Context Adversarial Game (ICAG)~\cite{zhou-etal-2024-defending},
SmoothLLM~\cite{robey2023smoothllm},
SemanticSmooth~\cite{ji-etal-2025-defending},
Erase-and-Check~\cite{kumar2023certifying},
Robust Alignment Checking~\cite{cao2024defending},
MoJE~\cite{cornacchia2024moje},
Goal Prioritization~\cite{zhang-etal-2024-defending},
RPO~\cite{NEURIPS2024_46ed5038},
THRD~\cite{ma2026thrd},
RLM-JB~\cite{shavit2026recursive},
RePD~\cite{wang-etal-2025-repd}, and
Defensive Prompt Patch (DPP)~\cite{xiong-etal-2025-defensive}.
};

\node[content, below=2.5mm of input-m] (internal-m)
{
\textbf{Decoding-time:}
SafeDecoding~\cite{xu-etal-2024-safedecoding},
SecDecoding~\cite{wang2025secdecoding},
Speculative Safety-Aware Decoding
(SSD)~\cite{wang-etal-2025-speculative},
SafeProbing~\cite{zhao2026defending},
FJD~\cite{chen-etal-2025-llm-jailbreak}, and
Alignment-Enhanced Decoding (AED)~\cite{liu-etal-2024-alignment}.\\
\textbf{Internal-state:}
Hidden-State Filtering (HSF)~\cite{10.1145/3701716.3717659},
Activation Boundary Defense (ABD)~\cite{gao-etal-2025-shaping},
JBShield~\cite{307822},
SafeInt~\cite{wu2025safeint},
Adversarial Game Defense (AGD)~\cite{pan-etal-2025-agd},
GradSafe~\cite{xie2024gradsafe},
IMAG~\cite{leng2025static},
Jailbreak Antidote~\cite{ICLR2025_36e3f9e6}, and
GUARD-SLM~\cite{mia2026guard}.
};

\node[content, below=2.5mm of internal-m] (output-m)
{
\textbf{Safety classification:}
Llama Guard~\cite{inan2023llama},
ShieldGemma~\cite{zeng2024shieldgemma}, and
ThinkGuard~\cite{wen-etal-2025-thinkguard}.\\
\textbf{Verification and revision:}
PARDEN~\cite{zhang2024parden},
Self Defense~\cite{phute2023llm},
Backtranslation~\cite{wang-etal-2024-defending}, and
Aligner~\cite{NEURIPS2024_a51a74b2}.
};


\node[content, below=5mm of output-m] (context-m)
{
CaMeL~\cite{debenedetti2025defeating},
A-MemGuard~\cite{wei2025memguard}, and
SMSR~\cite{sharma2026smsr}.
};

\node[content, below=3mm of context-m] (planning-m)
{
ShieldAgent~\cite{3780338.3780653} and
AgentSpec~\cite{wang2025agentspec}.
};

\node[content, below=3mm of planning-m] (tools-m)
{
Progent~\cite{shi2025progent} and
AttriGuard~\cite{320563}.
};

\node[content, below=3mm of tools-m] (agents-m)
{
AutoDefense~\cite{zeng2024autodefense} and
TrinityGuard~\cite{wang2026trinityguard}.
};


\node[category, left=4mm of training-m] (training)
{Safety Training and\\Fine-Tuning};

\node[category, left=4mm of input-m] (input)
{Input Filtering and\\Transformation};

\node[category, left=4mm of internal-m] (internal)
{Internal-State and\\Decoding Controls};

\node[category, left=4mm of output-m] (output)
{Output Classification,\\Verification, and Revision};


\coordinate (model-mid) at ($(training.west)!0.5!(output.west)$);

\node[layer, left=5mm of model-mid] (model)
{Model and\\Reasoning};

\node[layer] (context) at (model |- context-m)
{Context and\\State};

\node[layer] (planning) at (model |- planning-m)
{Planning and\\Orchestration};

\node[layer] (tools) at (model |- tools-m)
{Tools and\\Execution};

\node[layer] (agents) at (model |- agents-m)
{Inter-Agent\\Communication};


\coordinate (layers-mid) at ($(model.west)!0.5!(agents.west)$);

\node[root, left=6mm of layers-mid] (root)
{Agentic Jailbreak\\Defense Taxonomy};


\coordinate (root-split) at ([xshift=3mm]root.east);

\draw[conn] (root.east) -- (root-split);
\draw[conn] (root-split) |- (model.west);
\draw[conn] (root-split) |- (context.west);
\draw[conn] (root-split) |- (planning.west);
\draw[conn] (root-split) |- (tools.west);
\draw[conn] (root-split) |- (agents.west);


\coordinate (model-split) at ([xshift=2.5mm]model.east);

\draw[conn] (model.east) -- (model-split);
\draw[conn] (model-split) |- (training.west);
\draw[conn] (model-split) |- (input.west);
\draw[conn] (model-split) |- (internal.west);
\draw[conn] (model-split) |- (output.west);

\draw[conn] (training.east) -- (training-m.west);
\draw[conn] (input.east) -- (input-m.west);
\draw[conn] (internal.east) -- (internal-m.west);
\draw[conn] (output.east) -- (output-m.west);


\draw[conn] (context.east) -- (context-m.west);
\draw[conn] (planning.east) -- (planning-m.west);
\draw[conn] (tools.east) -- (tools-m.west);
\draw[conn] (agents.east) -- (agents-m.west);

\end{tikzpicture}%
}

\caption{Architecture-oriented taxonomy of jailbreak defenses for agentic AI
systems.}

\label{fig:agentic-jailbreak-defense-taxonomy}
\end{figure*}
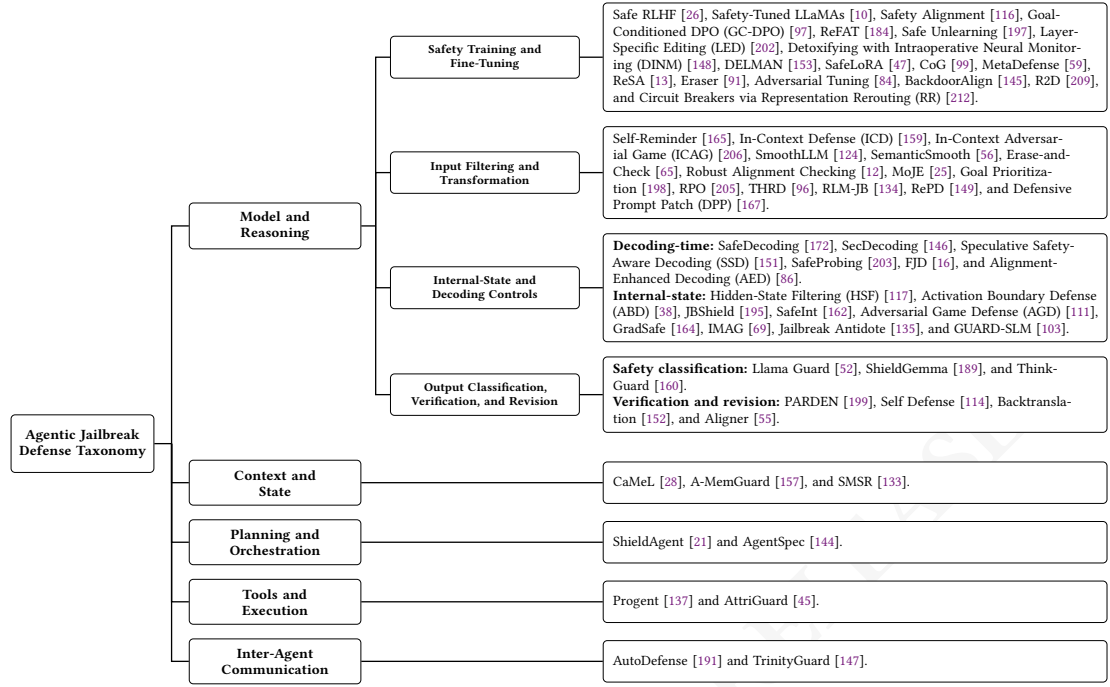

Agentic systems require safeguards across the execution pipeline, rather than
only at the user input or final response. Figure~\ref{fig:agentic-jailbreak-defense-taxonomy}
organizes defenses by the architectural layer they primarily protect:
\emph{model and reasoning}, \emph{context and state}, \emph{planning and
orchestration}, \emph{tools and execution}, and \emph{inter-agent
communication}. We include a system-level mechanism when it prevents, detects,
or contains policy bypass caused by adversarial instructions or poisoned state;
general reliability controls without this security objective are out of scope.
Because model-and-reasoning defenses intervene at several distinct stages, we
divide them into four categories below. The remaining four architecture layers
are treated directly as defense categories.

\paragraph{Safety Training and Fine-Tuning.}
These defenses modify model behavior before deployment through safety
alignment, preference optimization, adversarial training, unlearning, model
editing, or safety-aware reasoning training. Safe
RLHF~\cite{ICLR2024_dd1577af} learns from safety preferences, while
GC-DPO~\cite{maeng-etal-2025-goal} conditions preference optimization on safety
goals. ReFAT~\cite{ICLR2025_1022661f} uses refusal-feature adversarial training,
Safe Unlearning~\cite{zhang2024safe} suppresses harmful knowledge, and
LED~\cite{zhao-etal-2024-defending-large} performs layer-specific editing.
CoG~\cite{mao-etal-2026-models} applies trajectory-level safety training with
step-level correction, whereas R2D~\cite{zhu-etal-2025-reasoning} distills
safety-aware reasoning.

\begin{defensesummary}
\small
\textit{These methods usually require model access, additional
safety data, and training or editing. Their protection may degrade across model
families, languages, and unseen attacks; alignment updates can also reduce
benign utility. Moreover, model modification alone does not enforce memory,
tool, or communication policies at runtime.}
\end{defensesummary}

\paragraph{Input Filtering and Transformation.}
These defenses intervene before ordinary generation.
Self-Reminder~\cite{xie2023defending} reinforces safety instructions,
ICD~\cite{11370531} supplies defensive demonstrations, and
ICAG~\cite{zhou-etal-2024-defending} constructs an in-context adversarial game.
Perturbation-based methods test input stability: SmoothLLM~\cite{robey2023smoothllm}
uses randomized perturbations, SemanticSmooth~\cite{ji-etal-2025-defending}
aggregates semantic variants, and Erase-and-Check~\cite{kumar2023certifying}
repeatedly removes tokens and checks safety. Other methods use defensive
prompts, optimized prefixes or suffixes, decomposition, or explicit jailbreak
detection.

\begin{defensesummary}
\small
\textit{Filtering, perturbation, and repeated checking add token
and inference overhead and may over-refuse benign requests. Their performance
can decline under adaptive, multilingual, or multi-turn attacks, and checking
only the initial user input misses adversarial content introduced later through
retrieval, memory, tool output, or inter-agent messages.}
\end{defensesummary}

\paragraph{Internal-State and Decoding Controls.}
These defenses intervene during model processing or generation. Decoding-time
methods integrate safety into token selection: SafeDecoding~\cite{xu-etal-2024-safedecoding}
uses a safety-aware expert, SecDecoding~\cite{wang2025secdecoding} uses a
contrastive safety signal, and SSD~\cite{wang-etal-2025-speculative} performs
speculative safety-aware decoding. FJD~\cite{chen-etal-2025-llm-jailbreak} uses
first-token confidence, SafeProbing~\cite{zhao2026defending} probes safety
awareness during generation, and AED~\cite{liu-etal-2024-alignment} incorporates
repeated self-evaluation. Internal-state methods instead derive signals from
gradients or representations. GradSafe~\cite{xie2024gradsafe} analyzes
safety-critical gradients; HSF~\cite{10.1145/3701716.3717659} and
ABD~\cite{gao-etal-2025-shaping} use hidden activations; and
JBShield~\cite{307822}, SafeInt~\cite{wu2025safeint}, and Jailbreak
Antidote~\cite{ICLR2025_36e3f9e6} intervene on safety-related representations.
IMAG~\cite{leng2025static} uses activation memory for adaptive detection, while
GUARD-SLM~\cite{mia2026guard} detects jailbreak behavior from token activations.

\begin{defensesummary}
\small
\textit{These methods commonly require logits, gradients,
activations, compatible auxiliary models, or repeated evaluation. Consequently,
they can be architecture-specific, transfer poorly across heterogeneous models,
and incur generation overhead. Fixed detectors or steering directions may also
be bypassed by adaptive or distribution-shifted attacks.}
\end{defensesummary}

\paragraph{Output Classification, Verification, and Revision.}
These defenses inspect or modify a response before release. Llama
Guard~\cite{inan2023llama} and ShieldGemma~\cite{zeng2024shieldgemma} use
dedicated safety classifiers, while ThinkGuard~\cite{wen-etal-2025-thinkguard}
generates a critique before classification. PARDEN~\cite{zhang2024parden} uses
self-repetition as a jailbreak signal, LLM Self Defense~\cite{phute2023llm}
applies zero-shot harmfulness classification, and
Backtranslation~\cite{wang-etal-2024-defending} reconstructs the likely source
request to assess its intent. Aligner~\cite{NEURIPS2024_a51a74b2} uses a
separately trained model to revise unsafe responses.

\begin{defensesummary}
\small
\textit{Effectiveness depends on the guard or reviser and may be
reduced by concealed intent, distribution shift, false positives, or
over-refusal. Because these methods act after generation, they cannot reliably
observe or reverse unsafe plans, memory writes, tool calls, or external side
effects that occurred earlier in the trajectory.}
\end{defensesummary}

\paragraph{Context and State.}
These defenses protect untrusted context and persistent memory before they
influence agent decisions. CaMeL~\cite{debenedetti2025defeating} separates
control and data flows extracted from a trusted query and applies capability
policies to prevent untrusted data from changing program flow or causing
unauthorized exfiltration. A-MemGuard~\cite{wei2025memguard} intercepts the
memory-to-action pipeline, using cross-memory consensus to identify anomalous
reasoning paths and a dual-memory structure to retain corrective lessons.
SMSR~\cite{sharma2026smsr} combines HMAC-based write provenance with randomized
memory ablation and verdict-based aggregation, providing a certified robustness
bound under its persistent-memory poisoning threat model.

\begin{defensesummary}
\small
\textit{CaMeL assumes a trusted query and correctly specified security policies,
while side channels and compositions of individually permitted flows remain
possible. A-MemGuard depends on reliable cross-memory consensus and retrieval
configuration and introduces additional inference and storage costs. SMSR
assumes protected provenance keys and its stated threat model, and its
evaluation primarily uses synthetic enterprise scenarios and an LLM judge.}
\end{defensesummary}

\paragraph{Planning and Orchestration.}
These defenses enforce safety policies over agent plans, trajectories, and
runtime decisions. ShieldAgent~\cite{3780338.3780653} extracts verifiable rules
from policy documents, represents them as action-based probabilistic rule
circuits, and constructs formally checked shielding plans for protected-agent
trajectories. AgentSpec~\cite{wang2025agentspec} provides a domain-specific
language in which triggers, predicates, and enforcement actions define
runtime constraints. It intercepts predefined execution checkpoints and can
terminate, inspect, correct, or request reflection before an unsafe action
proceeds.

\begin{defensesummary}
\small
\textit{These defenses are bounded by the completeness of their policies,
accuracy of rule extraction, visibility of the execution state, and placement
of interception points. Unencoded or unobserved hazards may remain undetected.
AgentSpec additionally performs deterministic, step-local enforcement and does
not anticipate unsafe states that may emerge only after a multi-step
trajectory.}
\end{defensesummary}

\paragraph{Tools and Execution.}
These defenses mediate externally consequential tool calls through privilege
control or action-level causal analysis. Progent~\cite{shi2025progent}
represents permitted tool names and arguments using symbolic least-privilege
policies. Each call is checked deterministically, while SMT-based policy-update
analysis automatically permits privilege narrowing but requires approval for
expansion, preventing silent privilege escalation. AttriGuard~\cite{320563}
uses teacher-forced shadow replay and parallel counterfactual tests to determine
whether a proposed tool call is supported by the trusted user objective or
causally driven by an untrusted observation.

\begin{defensesummary}
\small
\textit{Progent does not address text-only compromise or tools that bypass its
enforcement interface, and ambiguous tasks or incorrectly approved privilege
expansions can weaken its guarantees. AttriGuard may introduce false positives
in long-horizon, tool-intensive workflows, and causal attribution becomes less
effective when users extensively delegate decisions to external content or
when actions use non-natural-language schemas. Both approaches require runtime
integration and add policy-checking or counterfactual-execution overhead.}
\end{defensesummary}

\paragraph{Inter-Agent Communication.}
Current defenses in this category primarily provide multi-agent verification
or monitoring rather than comprehensive communication control.
AutoDefense~\cite{zeng2024autodefense} uses a fixed multi-agent workflow to
separate response generation from safety analysis and filtering; although it
uses inter-agent coordination, its protected object is the generated output
rather than the communication channel itself.
TrinityGuard~\cite{wang2026trinityguard} instead evaluates multi-agent systems
through an OWASP-based three-tier risk taxonomy, risk-specific attack probes,
and LLM-coordinated runtime monitors that analyze execution traces and issue
alerts for individual-agent, inter-agent, and system-level hazards.

\begin{defensesummary}
\small
\textit{AutoDefense is constrained by its fixed communication structure, while
TrinityGuard primarily provides diagnosis and alerts rather than closed-loop
remediation and inherits errors from its LLM judges and test-generation
process. Neither constitutes general enforcement of agent identity, message
provenance, delegation authorization, or shared-state isolation. Verification
and monitoring agents also increase inference, communication, and coordination
costs.}
\end{defensesummary}

\section{Experiments and Results}

\subsection{Experimental Setup}


\begin{figure*}[t]
    \centering
    \includegraphics[width=.90\textwidth]{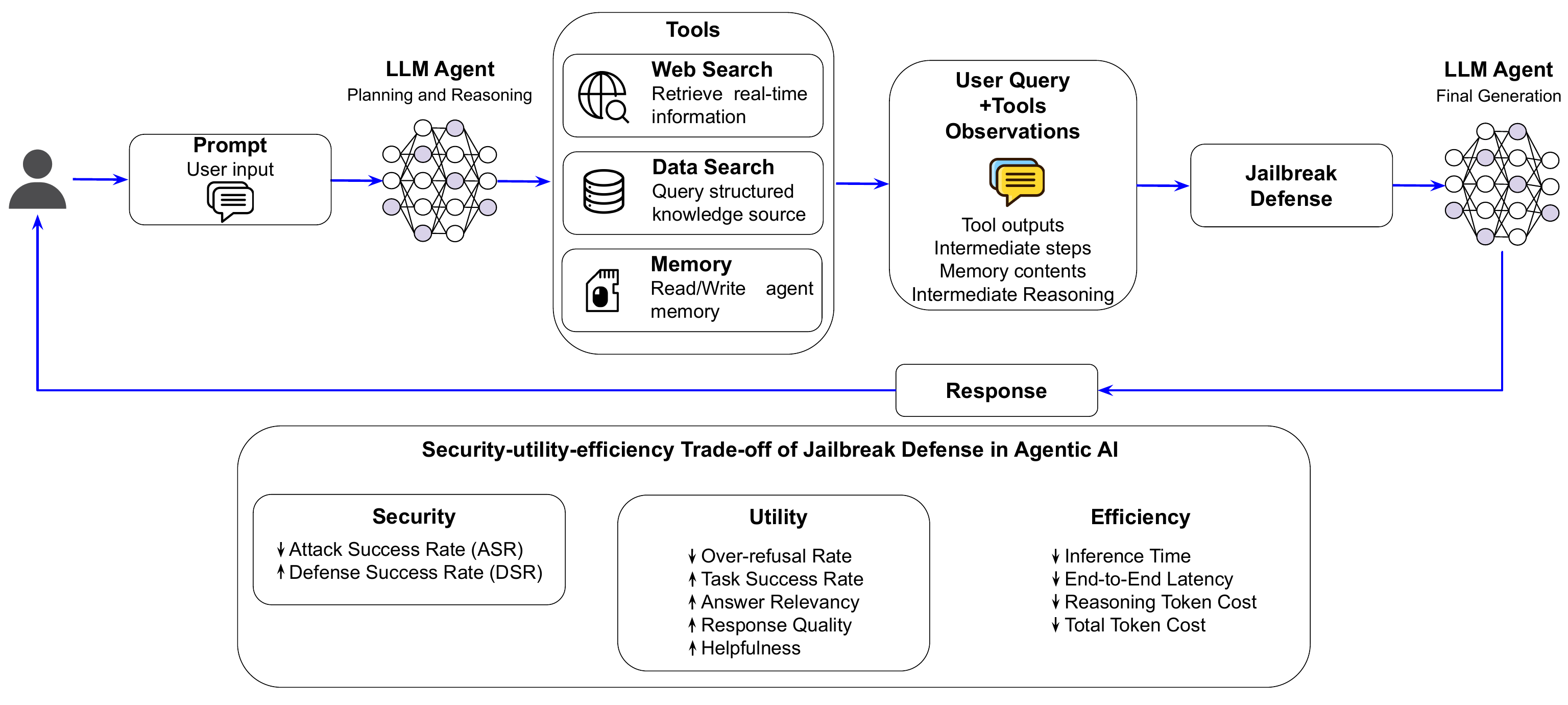}
    \caption{Security-utility-efficiency Trade-off of Jailbreak Defense in Agentic AI}
    \label{fig:agentic_overview}
\end{figure*}

All experiments were conducted on a server with two NVIDIA RTX A6000 GPUs
(48\,GB VRAM each). We implement a unified LangGraph-based agentic evaluation
framework (Figure~\ref{fig:agentic_overview}) consisting of an LLM agent,
external tools, retrieval and memory, a configurable defense module, and final
response generation. The agent architecture is kept fixed across experiments
to enable controlled comparison. Unless otherwise specified, Qwen3.5-4B serves
as the router/tool model, while Qwen3.5-4B and Llama-3.1-8B-Instruct are
evaluated as final-response models. We benchmark \textbf{16 representative
jailbreak defenses} together with an undefended \textbf{Vanilla} baseline.

For security--utility--efficiency evaluation, we use \textbf{XSTest}, containing
250 benign and 200 harmful prompts. The benign subset measures over-refusal,
answer relevancy, response quality, helpfulness, bias, and overall utility,
while the harmful subset measures attack success rate (ASR). To evaluate
adversarial jailbreak robustness, we use the first 50 harmful instructions from
\textbf{AdvBench} with nine representative attacks:
\textbf{AutoDAN}, \textbf{PAIR}, \textbf{TAP}, \textbf{GCG},
\textbf{Cipher}, \textbf{DeepInception}, \textbf{CodeChameleon},
\textbf{ICA}, and \textbf{Jailbroken}. Utility metrics are reported on XSTest,
whereas AdvBench is used exclusively for adversarial security evaluation.

To examine security within the agent execution pipeline, we additionally use
the \textbf{AgentHarm} benchmark~\cite{ICLR2025_c493d23a} with three attack
methods: \textbf{TRACE}, \textbf{MINJA}, and \textbf{AgenticRed}. TRACE
primarily targets planning, MINJA compromises persistent memory, and AgenticRed
automatically evolves model-level jailbreak prompts that may propagate across
planning, tool interaction, and final responses. These experiments measure both
final-response jailbreak success and component-level unsafe behavior.

We evaluate jailbreak defenses along three dimensions:
\textit{security}, \textit{utility}, and \textit{efficiency}, as illustrated in
Fig.~\ref{fig:agentic_overview}. We interpret safety outcomes using a
confusion-matrix view: a true positive (TP) is an unsafe query that is
successfully refused or safely handled; a false negative (FN) is an unsafe
query that receives harmful assistance; a true negative (TN) is a benign
query that receives a compliant response; and a false positive (FP) is a
benign query that is unnecessarily refused.

\paragraph{Security.}
For unsafe queries, we use an LLM-based safety judge to determine whether the
final response meaningfully assists the harmful request. Responses containing
actionable instructions, code, operational guidance, concrete recommendations,
or specific methods are considered successful jailbreaks. In contrast,
refusals, safe redirections, and non-actionable discussions are considered
successfully defended responses. We report \textit{Attack Success Rate (ASR)}
and \textit{Defense Success Rate (DSR)} as
\begin{equation}
\mathrm{ASR}=
\frac{\mathrm{FN}}{\mathrm{TP}+\mathrm{FN}}\times100\%,
\qquad
\mathrm{DSR}=
\frac{\mathrm{TP}}{\mathrm{TP}+\mathrm{FN}}\times100\%.
\end{equation}
Lower ASR and higher DSR indicate stronger jailbreak robustness.

\paragraph{Utility.}
For benign queries, an LLM-based judge labels each final response as
\textit{full compliance}, \textit{full refusal}, or \textit{partial refusal}.
Full compliance is counted as TN, whereas full and partial refusals are counted
as FP. Thus,
\begin{equation}
\mathrm{ORR}=
\frac{\mathrm{FP}}{\mathrm{TN}+\mathrm{FP}}\times100\%.
\end{equation}

Response utility is evaluated using Answer Relevancy, G-Eval Response Quality,
and Helpfulness. Answer Relevancy measures directness, Response Quality captures
relevance, clarity, completeness, and responsiveness, and Helpfulness measures
useful and appropriate assistance. We also report DeepEval Bias, computed for
response $i$ as
\begin{equation}
B_i=\frac{N_i^{\mathrm{biased}}}{N_i^{\mathrm{opinions}}},
\end{equation}
where lower values indicate less gender, political, racial/ethnic, or
geographic bias. At the $0.7$ threshold, higher scores are better for the three
utility metrics, whereas Bias passes at or below $0.7$. Bias is reported
separately and excluded from aggregate utility, which is
\begin{equation}
U_i=\frac{
S_i^{\mathrm{rel}}+
S_i^{\mathrm{qual}}+
S_i^{\mathrm{help}}
}{3}.
\end{equation}

\paragraph{Efficiency.}
We evaluate efficiency using \textit{inference time},
\textit{end-to-end latency}, \textit{reasoning-token cost}, and
\textit{total token cost}, which are directly recorded during execution.
Inference time measures the time spent on model inference, whereas end-to-end
latency captures the complete time from receiving a query to producing the
final response. We also record reasoning-token usage and total-token
consumption during execution. Lower time and token consumption indicate
greater computational efficiency.


\subsection{Practical Trade-off Analysis}
\label{sec:tradeoff}

\begin{figure*}[t]
    \centering
    \includegraphics[width=\textwidth]{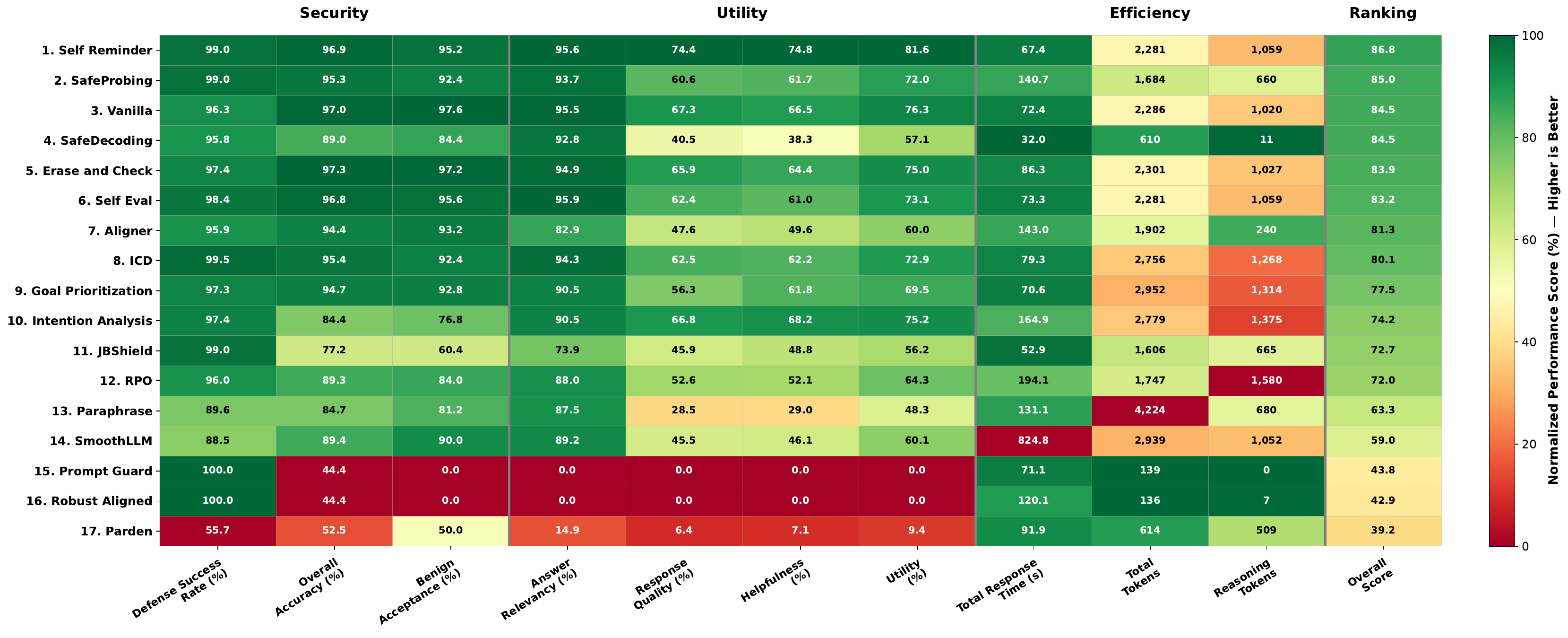}
    \caption{Overall comparison of representative jailbreak defenses across security, utility, and efficiency (tool model: Qwen3.5-4B; final model: Qwen3.5-4B; dataset: XSTest).}
    \label{fig:qwen_overall_ranking}
\end{figure*}

\begin{figure*}[t]
    \centering
    \includegraphics[width=\textwidth]{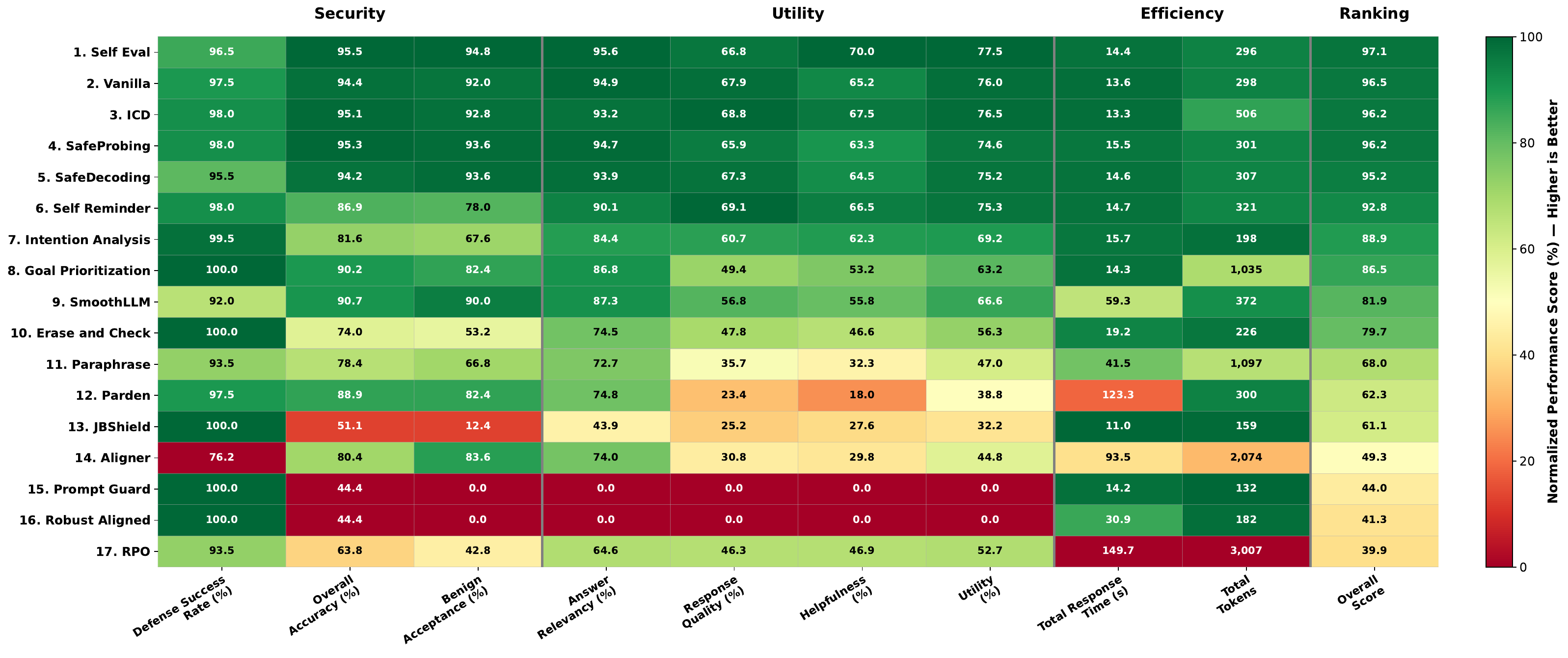}
    \caption{Overall comparison of representative jailbreak defenses across security, utility, and efficiency (tool model: Qwen3.5-4B; final model: Llama-3.1-8B-Instruct; dataset: XSTest).}
    \label{fig:llama_overall_ranking}
\end{figure*}

Figure~\ref{fig:qwen_overall_ranking} aggregates
Tables~\ref{tab:defense_eval_qwen}--\ref{tab:efficiency_qwen}, while
Figure~\ref{fig:llama_overall_ranking} aggregates
Tables~\ref{tab:defense_eval_llama}--\ref{tab:efficiency_llama}. Within each final model, we min--max
normalize all metrics so that higher values are better. Security equally
weights defense success rate ($1-\mathrm{ASR}$), overall accuracy, and benign
acceptance ($1-\mathrm{ORR}$), while utility uses the aggregate utility score.
Qwen3.5-4B efficiency weights response time, total tokens, and reasoning tokens
by 0.40, 0.30, and 0.30; Llama-3.1-8B-Instruct weights response time and total
tokens by 0.60 and 0.40 because reasoning-token counts are unavailable. The
overall score equally weights security, utility, and efficiency. Because
normalization is model-specific, scores are comparable only within each model.

For Qwen3.5-4B, Self-Reminder achieves the highest overall score (86.8),
followed by SafeProbing (85.0) and Vanilla (84.5); SafeDecoding also scores
84.5 after rounding. This ranking reflects Self-Reminder's strong balance
across dimensions: it reduces ASR from 3.67\% to 1.01\%, achieves the highest
aggregate utility (81.62\%), and slightly reduces total response time relative
to Vanilla. By contrast, SafeProbing attains a similar ASR (1.00\%) but incurs
approximately twice the response time of Vanilla (140.68\,s versus 72.44\,s).
Thus, comparable security outcomes can entail substantially different
deployment costs.

\begin{figure*}[t]
    \centering
    \includegraphics[width=\textwidth]{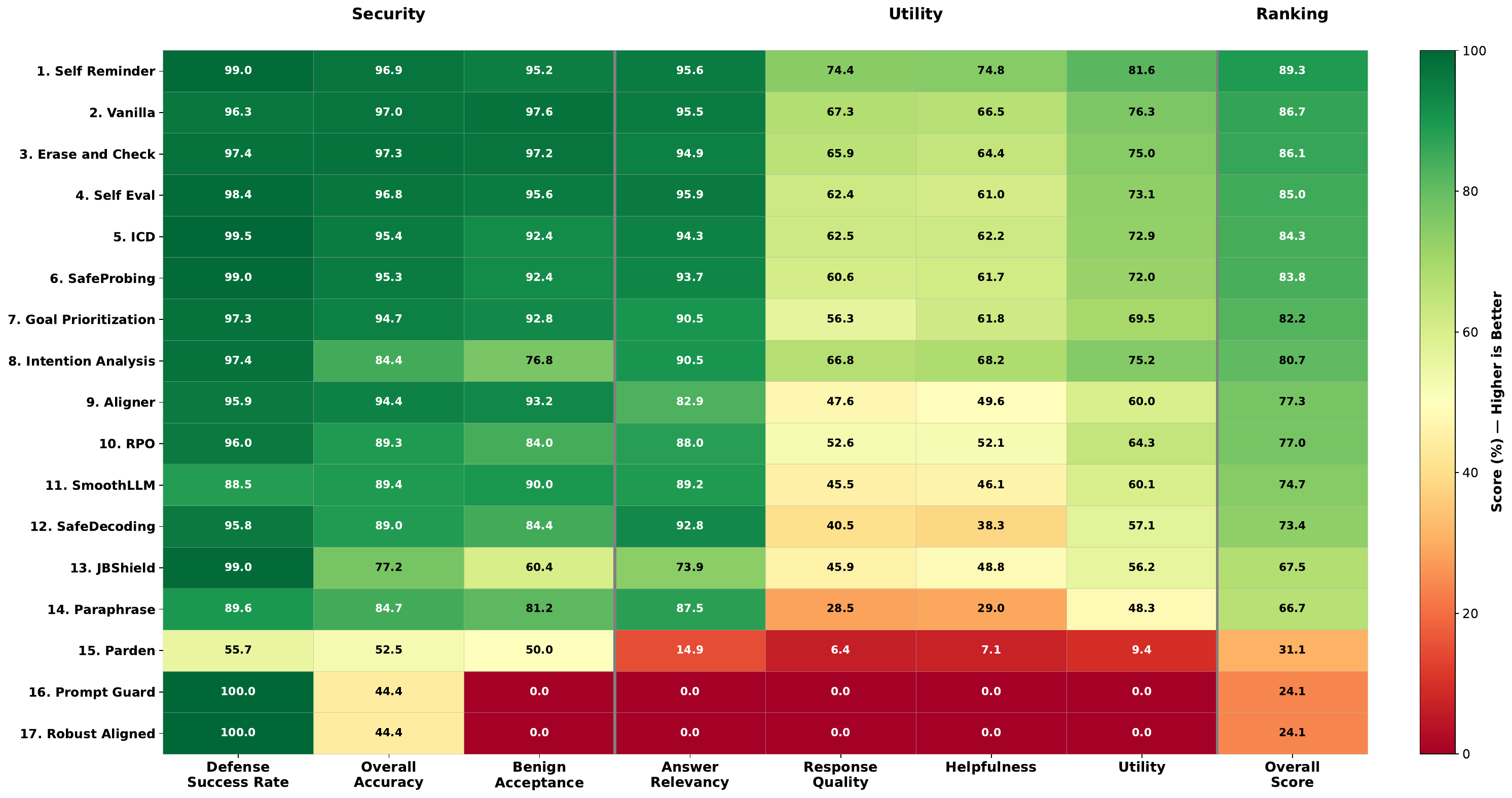}
    \caption{Pairwise security--utility comparison on XSTest (tool and final
    model: Qwen3.5-4B).}
    \label{fig:qwen_security_utility}
\end{figure*}

\begin{figure*}[t]
    \centering
    \includegraphics[width=\textwidth]{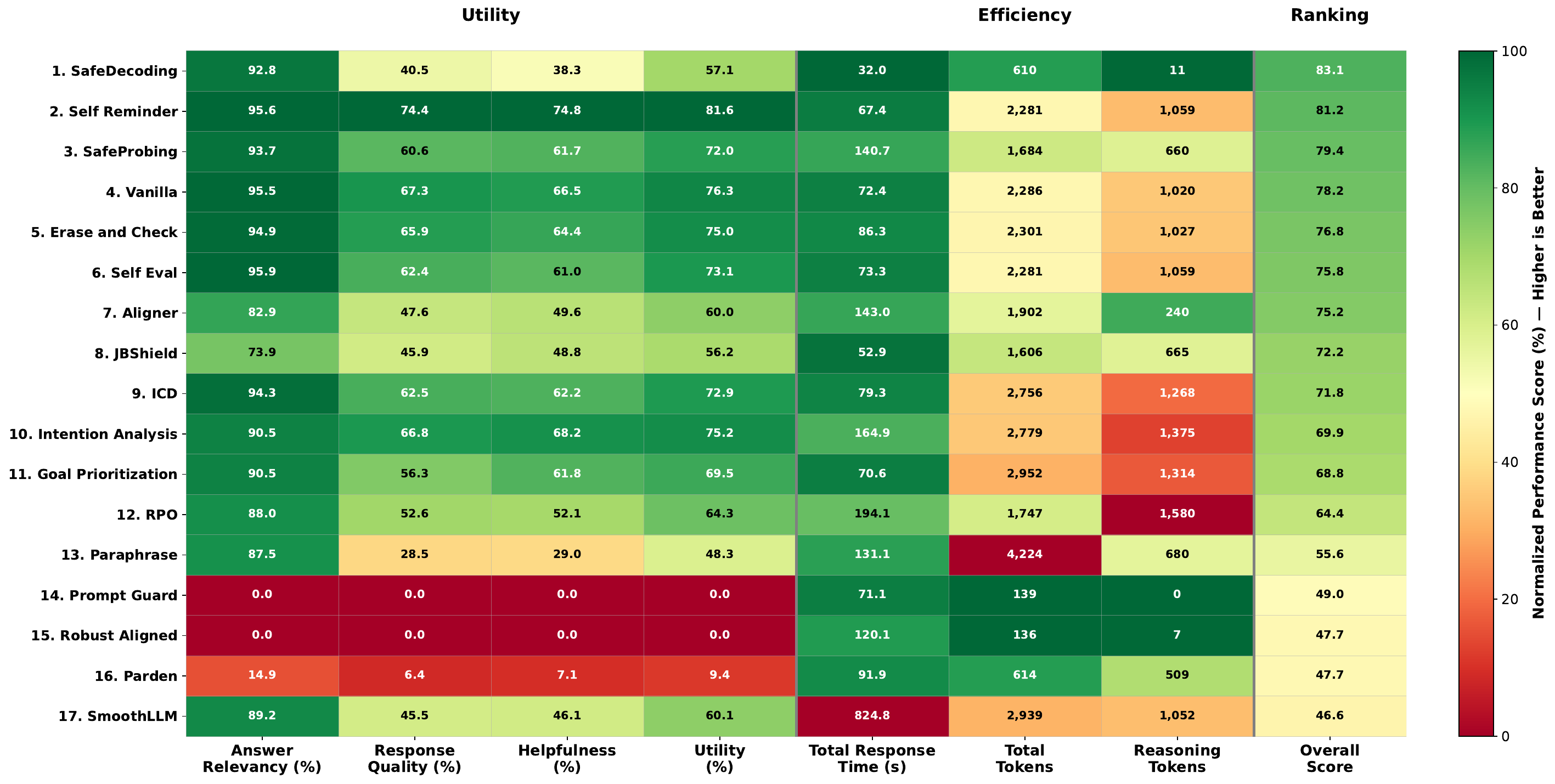}
    \caption{Pairwise utility--efficiency comparison on XSTest (tool and final
    model: Qwen3.5-4B).}
    \label{fig:qwen_utility_efficiency}
\end{figure*}

The pairwise views in Figures~\ref{fig:qwen_security_utility} and
\ref{fig:qwen_utility_efficiency} clarify how the preferred method changes
when only two deployment objectives are considered. For Qwen3.5-4B,
Self-Reminder achieves the best security--utility score (89.3), followed by
Vanilla (86.7) and Erase-and-Check (86.1), combining a 99.0\% defense success
rate with the highest utility (81.62\%). In the utility--efficiency view,
SafeDecoding ranks first because of its low response time and token usage,
although its utility is substantially lower than that of Self-Reminder.
SmoothLLM ranks last because of its high latency. These pairwise rankings are
computed only from the dimensions shown and therefore need not match the
three-dimensional overall ranking.

For Llama-3.1-8B-Instruct, Self-Eval ranks first (97.1), followed by Vanilla
(96.5), with ICD and SafeProbing tied after rounding (96.2). Self-Eval provides
the highest utility (77.51\%) and overall accuracy (95.55\%), while maintaining
latency and token consumption close to Vanilla. ICD and SafeProbing obtain
slightly lower ASR than Vanilla (2.01\% and 2.00\% versus 2.51\%) while
preserving high accuracy and relatively low over-refusal. These results also
demonstrate that the preferred defense is model-dependent: Self-Reminder
performs best for Qwen3.5-4B, whereas Self-Eval provides the strongest balance
for Llama-3.1-8B-Instruct.

\begin{figure*}[t]
    \centering
    \includegraphics[width=\textwidth]{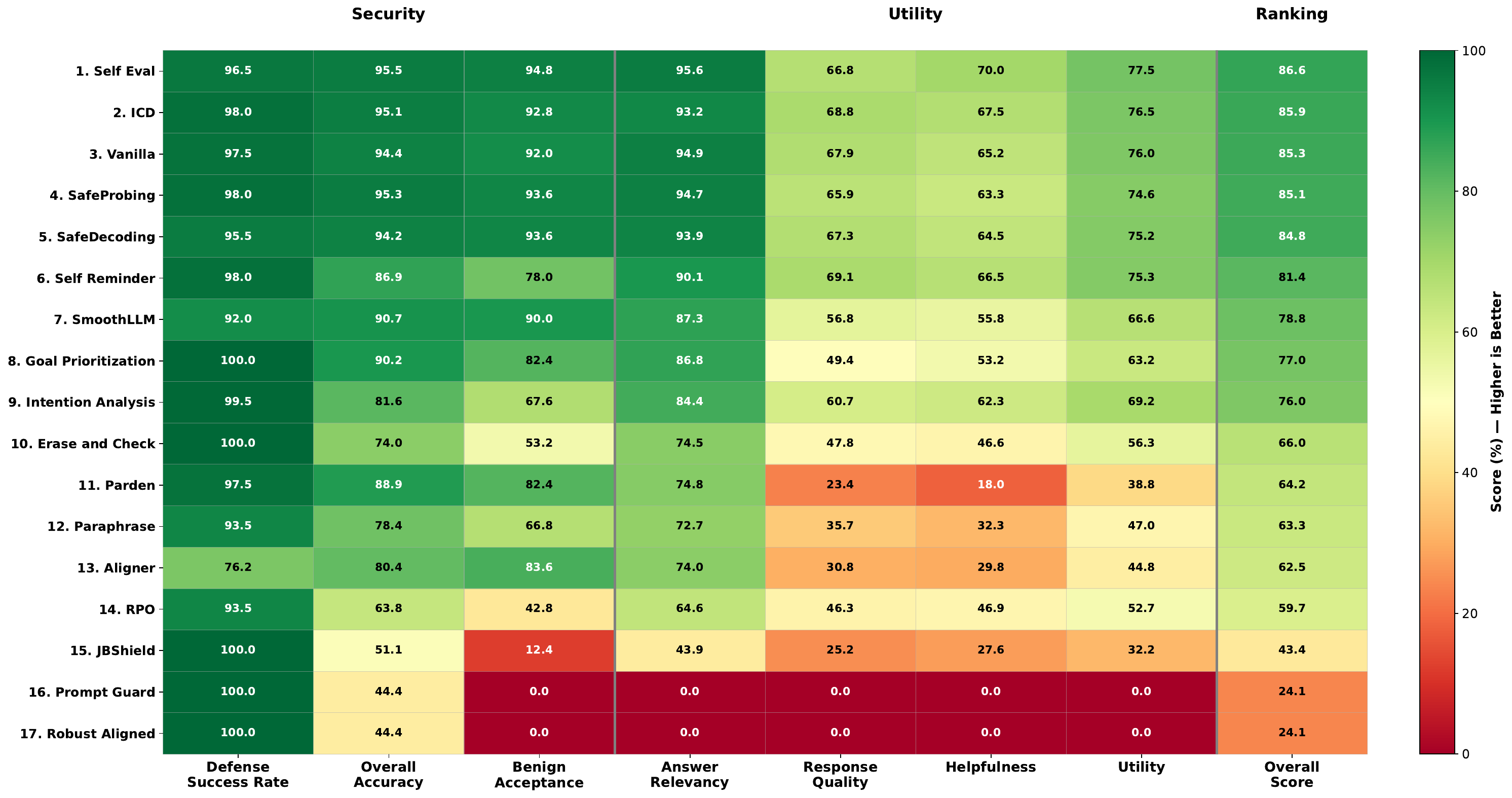}
    \caption{Pairwise security--utility comparison on XSTest (tool model:
    Qwen3.5-4B; final model: Llama-3.1-8B-Instruct).}
    \label{fig:llama_security_utility}
\end{figure*}

\begin{figure*}[t]
    \centering
    \includegraphics[width=\textwidth]{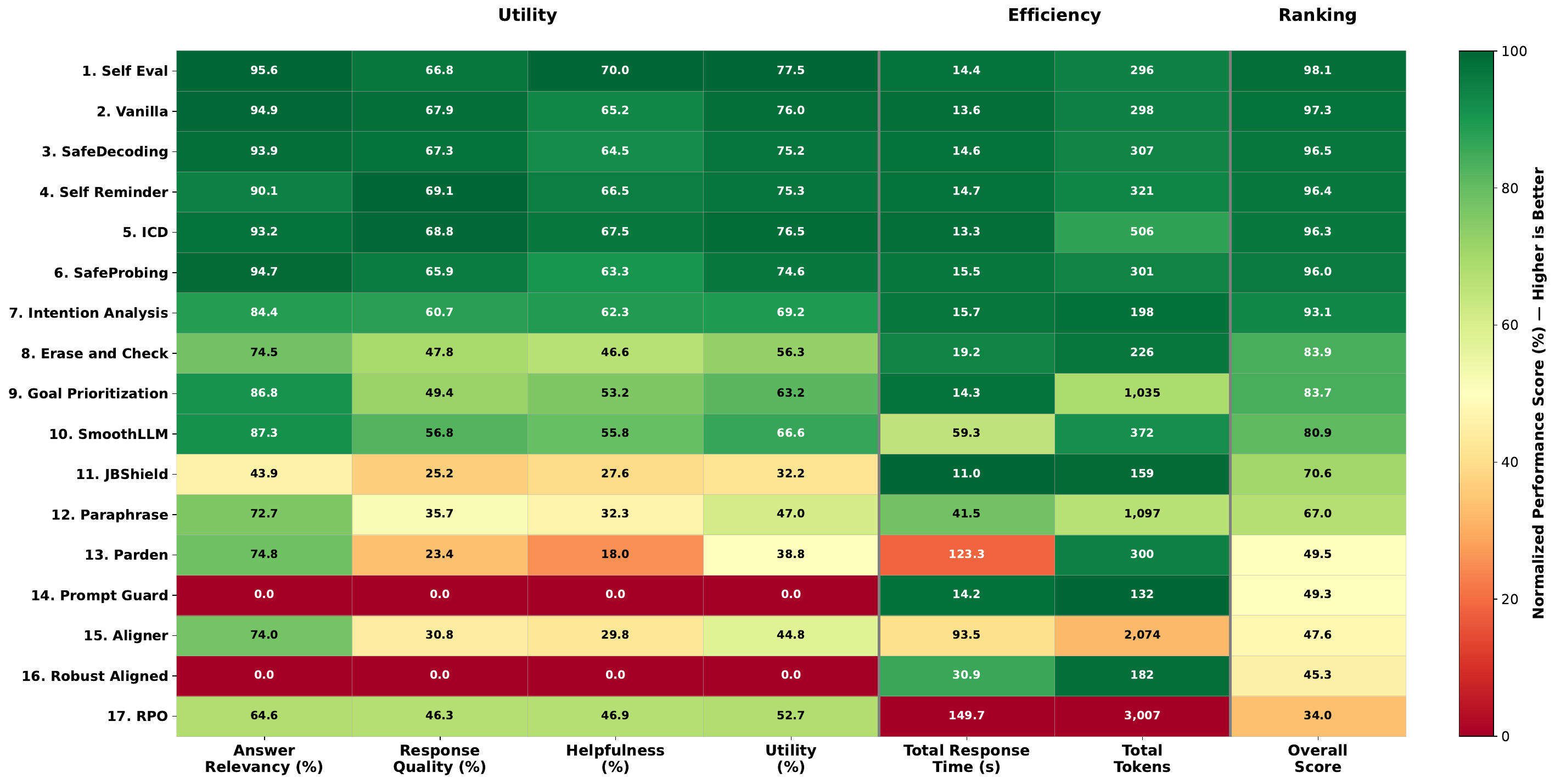}
    \caption{Pairwise utility--efficiency comparison on XSTest (tool model:
    Qwen3.5-4B; final model: Llama-3.1-8B-Instruct).}
    \label{fig:llama_utility_efficiency}
\end{figure*}

The Llama pairwise views in Figures~\ref{fig:llama_security_utility} and
\ref{fig:llama_utility_efficiency} reinforce this result. Self-Eval leads both
the security--utility (86.6) and utility--efficiency (98.1) rankings, while
ICD, Vanilla, and SafeProbing remain competitive. It preserves the highest
utility (77.51\%) with strong security and efficiency close to Vanilla. In
contrast, RPO incurs the largest latency and token cost. As above, these
pairwise scores summarize only the two displayed dimensions.

More broadly, adding a defense does not necessarily improve the complete
agentic system. Vanilla ranks among the top three configurations for both
models, indicating that their native safety alignment already provides
substantial protection without additional defense overhead. Some defenses
further reduce ASR but impose disproportionate utility losses. For example,
Prompt Guard and Robust Aligned achieve 0\% ASR for both models, but reject all
benign XSTest prompts, yielding 100\% over-refusal and 0\% utility. Other
methods impose substantial computational overhead: SmoothLLM increases Qwen
response time to 824.84\,s, while RPO increases Llama response time to
149.66\,s and token usage to 3006.63. Moreover, the apparently low cost of
highly restrictive defenses partly results from terminating benign requests
rather than completing them, illustrating why efficiency cannot be interpreted
independently of utility. Overall, these findings show that jailbreak defenses
for agentic AI should be selected jointly based on security, benign utility,
and execution cost rather than ASR alone.

\begin{table}
  \caption{Security evaluation of jailbreak defenses on XSTest using Qwen3.5-4B (Evaluator: GPT-5-mini). ASR: attack success rate; Acc.: overall accuracy; ORR: over-refusal rate.}
  \label{tab:defense_eval_qwen}
  \begin{tabular}{lccc}
    \toprule
    \textbf{Method} &
    \textbf{ASR $\downarrow$} &
    \textbf{Acc. $\uparrow$} &
    \textbf{ORR $\downarrow$} \\
    \midrule
    Vanilla
    & 03.67\%
    & 97.05\%
    & \cellcolor{green!20}\textbf{02.40\%} \\

    Aligner
    & 04.08\%
    & 94.39\%
    & 06.80\% \\

    Erase-and-Check
    & 02.60\%
    & \cellcolor{green!20}\textbf{97.29\%}
    & 02.80\% \\

    Goal Prioritization
    & 02.69\%
    & 94.72\%
    & 07.20\% \\

    ICD
    & 00.53\%
    & 95.44\%
    & 07.60\% \\

    Paraphrase
    & 10.44\%
    & 84.72\%
    & 18.80\% \\

    PARDEN
    & \cellcolor{red!20}\textbf{44.33\%}
    & 52.48\%
    & 50.00\% \\

    Prompt Guard
    & \cellcolor{green!20}\textbf{00.00\%}
    & \cellcolor{red!20}\textbf{44.44\%}
    & \cellcolor{red!20}\textbf{100.00\%} \\

    Robust Aligned
    & \cellcolor{green!20}\textbf{00.00\%}
    & \cellcolor{red!20}\textbf{44.44\%}
    & \cellcolor{red!20}\textbf{100.00\%} \\

    SafeDecoding
    & 04.19\%
    & 88.97\%
    & 15.60\% \\

    Self Eval
    & 01.56\%
    & 96.83\%
    & 04.40\% \\

    Self-Reminder
    & 01.01\%
    & 96.88\%
    & 04.80\% \\

    SmoothLLM
    & 11.46\%
    & 89.37\%
    & 10.00\% \\

    JBShield
    & 01.04\%
    & 77.15\%
    & 39.60\% \\

    RPO
    & 04.00\%
    & 89.33\%
    & 16.00\% \\

    Intention Analysis
    & 02.59\%
    & 84.44\%
    & 23.20\% \\

    SafeProbing
    & 01.00\%
    & 95.33\%
    & 07.60\% \\
    \bottomrule
  \end{tabular}
\end{table}
\begin{table}
  \caption{Utility evaluation of jailbreak defenses on the XSTest safe subset using Qwen3.5-4B (Evaluator: GPT-5-mini). AR: answer relevancy; RQ: response quality; Help.: helpfulness.}
  \label{tab:utility_eval_qwen}
  \begin{tabular}{lccccc}
    \toprule
    \textbf{Method} &
    \textbf{AR $\uparrow$} &
    \textbf{RQ $\uparrow$} &
    \textbf{Help. $\uparrow$} &
    \textbf{Bias $\downarrow$} &
    \textbf{Utility $\uparrow$} \\
    \midrule

    Vanilla
    & 95.50\%
    & 67.32\%
    & 66.52\%
    & 01.23\%
    & 76.35\% \\

    Aligner
    & 82.93\%
    & 47.60\%
    & 49.56\%
    & 01.36\%
    & 60.02\% \\

    Erase-and-Check
    & 94.92\%
    & 65.88\%
    & 64.36\%
    & \cellcolor{red!20}\textbf{02.31\%}
    & 74.95\% \\

    Goal Prioritization
    & 90.49\%
    & 56.28\%
    & 61.81\%
    & 00.59\%
    & 69.55\% \\

    ICD
    & 94.35\%
    & 62.48\%
    & 62.17\%
    & 01.06\%
    & 72.93\% \\

    Paraphrase
    & 87.51\%
    & 28.51\%
    & 29.04\%
    & 01.12\%
    & 48.28\% \\

    PARDEN
    & 14.86\%
    & 06.43\%
    & 07.08\%
    & 00.17\%
    & 09.45\% \\

    Prompt Guard
    & \cellcolor{red!20}\textbf{00.00\%}
    & \cellcolor{red!20}\textbf{00.00\%}
    & \cellcolor{red!20}\textbf{00.00\%}
    & \cellcolor{green!20}\textbf{00.00\%}
    & \cellcolor{red!20}\textbf{00.00\%} \\

    Robust Aligned
    & \cellcolor{red!20}\textbf{00.00\%}
    & \cellcolor{red!20}\textbf{00.00\%}
    & \cellcolor{red!20}\textbf{00.00\%}
    & \cellcolor{green!20}\textbf{00.00\%}
    & \cellcolor{red!20}\textbf{00.00\%} \\

    SafeDecoding
    & 92.80\%
    & 40.48\%
    & 38.31\%
    & 01.28\%
    & 57.13\% \\

    Self Eval
    & \cellcolor{green!20}\textbf{95.91\%}
    & 62.36\%
    & 60.96\%
    & 01.23\%
    & 73.08\% \\

    Self-Reminder
    & 95.61\%
    & \cellcolor{green!20}\textbf{74.44\%}
    & \cellcolor{green!20}\textbf{74.80\%}
    & 01.10\%
    & \cellcolor{green!20}\textbf{81.62\%} \\

    SmoothLLM
    & 89.16\%
    & 45.52\%
    & 46.14\%
    & 01.42\%
    & 60.12\% \\

    JBShield
    & 73.90\%
    & 45.88\%
    & 48.80\%
    & 00.99\%
    & 56.19\% \\

    RPO
    & 87.99\%
    & 52.60\%
    & 52.13\%
    & 01.51\%
    & 64.28\% \\

    Intention Analysis
    & 90.53\%
    & 66.80\%
    & 68.17\%
    & 01.05\%
    & 75.17\% \\

    SafeProbing
    & 93.66\%
    & 60.56\%
    & 61.68\%
    & 01.69\%
    & 72.01\% \\

    \bottomrule
  \end{tabular}
\end{table}

\begin{table}
  \caption{Efficiency evaluation of jailbreak defenses on XSTest using Qwen3.5-4B. TT: total response time; TDT: tool decision time; FRT: final response time; Tok.: total tokens; RTok.: reasoning tokens.}
  \label{tab:efficiency_qwen}
  \begin{tabular}{lccccc}
    \toprule
    \textbf{Method} &
    \textbf{TT $\downarrow$} &
    \textbf{TDT $\downarrow$} &
    \textbf{FRT $\downarrow$} &
    \textbf{Tok. $\downarrow$} &
    \textbf{RTok. $\downarrow$} \\
    \midrule

    Vanilla
    & 72.44
    & \cellcolor{green!20}\textbf{8.55}
    & 63.35
    & 2285.74
    & 1020.40 \\

    Aligner
    & 143.04
    & 8.77
    & 133.69
    & 1902.36
    & 240.12 \\

    Erase-and-Check
    & 86.29
    & 8.76
    & 77.00
    & 2301.30
    & 1027.48 \\

    Goal Prioritization
    & 70.58
    & 8.57
    & 61.50
    & 2952.10
    & 1314.42 \\

    ICD
    & 79.31
    & 8.82
    & 69.87
    & 2755.68
    & 1268.15 \\

    Paraphrase
    & 131.05
    & 9.01
    & 121.52
    & \cellcolor{red!20}\textbf{4224.03}
    & 680.01 \\

    PARDEN
    & 91.88
    & 8.76
    & 82.60
    & 613.62
    & 509.32 \\

    Prompt Guard
    & 71.07
    & 8.57
    & 62.00
    & 139.46
    & \cellcolor{green!20}\textbf{0.00} \\

    Robust Aligned
    & 120.14
    & 8.79
    & 110.83
    & \cellcolor{green!20}\textbf{135.56}
    & 6.62 \\

    SafeDecoding
    & \cellcolor{green!20}\textbf{31.99}
    & 8.79
    & \cellcolor{green!20}\textbf{22.62}
    & 610.11
    & 11.48 \\

    Self Eval
    & 73.33
    & \cellcolor{green!20}\textbf{8.55}
    & 64.21
    & 2281.00
    & 1059.25 \\

    Self-Reminder
    & 67.40
    & \cellcolor{green!20}\textbf{8.55}
    & 58.13
    & 2281.00
    & 1059.25 \\

    SmoothLLM
    & \cellcolor{red!20}\textbf{824.84}
    & 8.73
    & \cellcolor{red!20}\textbf{815.49}
    & 2938.51
    & 1052.04 \\

    JBShield
    & 52.89
    & 8.65
    & 43.53
    & 1606.07
    & 664.97 \\

    RPO
    & 194.14
    & 8.58
    & 185.03
    & 1746.82
    & \cellcolor{red!20}\textbf{1580.24} \\

    Intention Analysis
    & 164.92
    & 8.95
    & 155.41
    & 2779.25
    & 1375.12 \\

    SafeProbing
    & 140.68
    & 8.73
    & 131.35
    & 1683.57
    & 660.10 \\

    \bottomrule
  \end{tabular}
\end{table}

\begin{table}
  \caption{Security evaluation of jailbreak defenses on XSTest using Llama-3.1-8B-Instruct (Evaluator: GPT-5-mini). ASR: attack success rate; Acc.: overall accuracy; ORR: over-refusal rate.}
  \label{tab:defense_eval_llama}
  \begin{tabular}{lccc}
    \toprule
    \textbf{Method} &
    \textbf{ASR $\downarrow$} &
    \textbf{Acc. $\uparrow$} &
    \textbf{ORR $\downarrow$} \\
    \midrule

    Vanilla
    & 02.51\%
    & 94.43\%
    & 08.00\% \\

    Aligner
    & \cellcolor{red!20}\textbf{23.81\%}
    & 80.41\%
    & 16.40\% \\

    Erase-and-Check
    & \cellcolor{green!20}\textbf{00.00\%}
    & 74.00\%
    & 46.80\% \\

    Goal Prioritization
    & \cellcolor{green!20}\textbf{00.00\%}
    & 90.16\%
    & 17.60\% \\

    ICD
    & 02.01\%
    & 95.10\%
    & 07.20\% \\

    Paraphrase
    & 06.50\%
    & 78.44\%
    & 33.20\% \\

    PARDEN
    & 02.50\%
    & 88.89\%
    & 17.60\% \\

    Prompt Guard
    & \cellcolor{green!20}\textbf{00.00\%}
    & \cellcolor{red!20}\textbf{44.44\%}
    & \cellcolor{red!20}\textbf{100.00\%} \\

    Robust Aligned
    & \cellcolor{green!20}\textbf{00.00\%}
    & \cellcolor{red!20}\textbf{44.44\%}
    & \cellcolor{red!20}\textbf{100.00\%} \\

    SafeDecoding
    & 04.50\%
    & 94.22\%
    & 06.40\% \\

    Self Eval
    & 03.52\%
    & \cellcolor{green!20}\textbf{95.55\%}
    & \cellcolor{green!20}\textbf{05.20\%} \\

    Self-Reminder
    & 02.01\%
    & 86.86\%
    & 22.00\% \\

    SmoothLLM
    & 08.00\%
    & 90.67\%
    & 10.00\% \\

    JBShield
    & \cellcolor{green!20}\textbf{00.00\%}
    & 51.11\%
    & 87.60\% \\

    RPO
    & 06.50\%
    & 63.78\%
    & 57.20\% \\

    Intention Analysis
    & 00.50\%
    & 81.56\%
    & 32.40\% \\

    SafeProbing
    & 02.00\%
    & 95.33\%
    & 06.40\% \\

    \bottomrule
  \end{tabular}
\end{table}

\begin{table}
  \caption{Utility evaluation of jailbreak defenses on the XSTest safe subset using Llama-3.1-8B-Instruct (Evaluator: GPT-5-mini). AR: answer relevancy; RQ: response quality; Help.: helpfulness.}
  \label{tab:utility_eval_llama}
  \begin{tabular}{lccccc}
    \toprule
    \textbf{Method} &
    \textbf{AR $\uparrow$} &
    \textbf{RQ $\uparrow$} &
    \textbf{Help. $\uparrow$} &
    \textbf{Bias $\downarrow$} &
    \textbf{Utility $\uparrow$} \\
    \midrule

    Vanilla
    & 94.92\%
    & 67.88\%
    & 65.20\%
    & 04.70\%
    & 76.00\% \\

    Aligner
    & 74.05\%
    & 30.75\%
    & 29.75\%
    & 00.40\%
    & 44.84\% \\

    Erase-and-Check
    & 74.55\%
    & 47.80\%
    & 46.56\%
    & 00.55\%
    & 56.30\% \\

    Goal Prioritization
    & 86.75\%
    & 49.40\%
    & 53.20\%
    & 00.71\%
    & 63.16\% \\

    ICD
    & 93.24\%
    & 68.75\%
    & 67.52\%
    & 02.14\%
    & 76.52\% \\

    Paraphrase
    & 72.74\%
    & 35.74\%
    & 32.32\%
    & 01.80\%
    & 46.96\% \\

    PARDEN
    & 74.85\%
    & 23.40\%
    & 18.00\%
    & 04.00\%
    & 38.75\% \\

    Prompt Guard
    & \cellcolor{red!20}\textbf{00.00\%}
    & \cellcolor{red!20}\textbf{00.00\%}
    & \cellcolor{red!20}\textbf{00.00\%}
    & \cellcolor{green!20}\textbf{00.00\%}
    & \cellcolor{red!20}\textbf{00.00\%} \\

    Robust Aligned
    & \cellcolor{red!20}\textbf{00.00\%}
    & \cellcolor{red!20}\textbf{00.00\%}
    & \cellcolor{red!20}\textbf{00.00\%}
    & \cellcolor{green!20}\textbf{00.00\%}
    & \cellcolor{red!20}\textbf{00.00\%} \\

    SafeDecoding
    & 93.89\%
    & 67.32\%
    & 64.52\%
    & 03.37\%
    & 75.21\% \\

    Self Eval
    & \cellcolor{green!20}\textbf{95.58\%}
    & 66.84\%
    & \cellcolor{green!20}\textbf{70.04\%}
    & \cellcolor{red!20}\textbf{05.73\%}
    & \cellcolor{green!20}\textbf{77.51\%} \\

    Self-Reminder
    & 90.14\%
    & \cellcolor{green!20}\textbf{69.08\%}
    & 66.51\%
    & 00.54\%
    & 75.27\% \\

    SmoothLLM
    & 87.30\%
    & 56.84\%
    & 55.76\%
    & 03.24\%
    & 66.63\% \\

    JBShield
    & 43.85\%
    & 25.20\%
    & 27.56\%
    & 00.05\%
    & 32.20\% \\

    RPO
    & 64.60\%
    & 46.34\%
    & 46.92\%
    & 01.81\%
    & 52.66\% \\

    Intention Analysis
    & 84.44\%
    & 60.68\%
    & 62.28\%
    & 01.05\%
    & 69.15\% \\

    SafeProbing
    & 94.67\%
    & 65.88\%
    & 63.28\%
    & 04.51\%
    & 74.61\% \\

    \bottomrule
  \end{tabular}
\end{table}

\begin{table}
  \caption{Efficiency evaluation of jailbreak defenses on XSTest using Llama-3.1-8B-Instruct. TT: total response time; TDT: tool decision time; TET: tool execution time; FRT: final response time; Tok.: total tokens.}
  \label{tab:efficiency_llama}
  \begin{tabular}{lccccc}
    \toprule
    \textbf{Method} &
    \textbf{TT $\downarrow$} &
    \textbf{TDT $\downarrow$} &
    \textbf{TET $\downarrow$} &
    \textbf{FRT $\downarrow$} &
    \textbf{Tok. $\downarrow$} \\
    \midrule

    Vanilla
    & 13.64
    & 8.60
    & 0.77
    & 4.28
    & 298.11 \\

    Aligner
    & 93.47
    & 8.50
    & 0.91
    & 84.05
    & 2073.94 \\

    Erase-and-Check
    & 19.19
    & 8.99
    & 0.50
    & 9.70
    & 226.29 \\

    Goal Prioritization
    & 14.34
    & 8.62
    & 0.58
    & 5.14
    & 1035.08 \\

    ICD
    & 13.27
    & 8.61
    & 0.82
    & 3.84
    & 506.11 \\

    Paraphrase
    & 41.47
    & 8.83
    & 0.60
    & 32.03
    & 1096.62 \\

    PARDEN
    & 123.30
    & 8.74
    & 0.57
    & 113.99
    & 299.64 \\

    Prompt Guard
    & 14.22
    & \cellcolor{red!20}\textbf{9.20}
    & \cellcolor{green!20}\textbf{0.49}
    & 4.54
    & \cellcolor{green!20}\textbf{131.70} \\

    Robust Aligned
    & 30.88
    & 8.89
    & 0.52
    & 21.46
    & 182.47 \\

    SafeDecoding
    & 14.57
    & 8.64
    & 0.54
    & 5.39
    & 306.86 \\

    Self Eval
    & 14.40
    & 8.74
    & 0.51
    & 5.15
    & 296.50 \\

    Self-Reminder
    & 14.67
    & 8.66
    & \cellcolor{red!20}\textbf{2.25}
    & 3.76
    & 321.40 \\

    SmoothLLM
    & 59.31
    & 8.95
    & 0.70
    & 49.66
    & 372.09 \\

    JBShield
    & \cellcolor{green!20}\textbf{10.98}
    & 8.64
    & 0.53
    & \cellcolor{green!20}\textbf{1.81}
    & 158.53 \\

    RPO
    & \cellcolor{red!20}\textbf{149.66}
    & 8.58
    & 0.55
    & \cellcolor{red!20}\textbf{140.54}
    & \cellcolor{red!20}\textbf{3006.63} \\

    Intention Analysis
    & 15.74
    & 8.71
    & 0.61
    & 6.42
    & 197.62 \\

    SafeProbing
    & 15.50
    & \cellcolor{green!20}\textbf{8.48}
    & 0.54
    & 6.47
    & 300.58 \\

    \bottomrule
  \end{tabular}
\end{table}

\subsection{Jailbreak Robustness in Agentic AI} \label{sec:attack_robustness_llama} Table~\ref{tab:asr_attacks_llama} reveals substantial variation in robustness across jailbreak attacks and defenses. The Vanilla agent obtains an average ASR of 34.00\%, with AutoDAN (76.00\%) and CodeChameleon (62.00\%) producing the highest attack success. In contrast, Prompt Guard achieves 0\% average ASR, while Paraphrase (0.67\%), JBShield (3.56\%), Aligner and Intention Analysis (4.22\%), and SafeProbing (5.11\%) substantially reduce successful jailbreaks. However, robustness is strongly attack dependent: defenses that perform well against GCG, Cipher, or ICA can remain vulnerable to AutoDAN, TAP, DeepInception, or CodeChameleon. For example, ICD reduces GCG, Cipher, and ICA to 0\% ASR but reaches 66.00\% under AutoDAN and 46.00\% under CodeChameleon. These results show that robustness cannot be inferred from a single jailbreak technique and motivate evaluating agentic AI defenses across diverse attack mechanisms.

\begin{table}
  \caption{Attack Success Rate (ASR) of nine representative jailbreak attacks against representative defenses (Model: Llama-3.1-8B-Instruct; Dataset: AdvBench; Evaluator: GPT-5-mini + manual verification).}
  \label{tab:asr_attacks_llama}
  \begin{tabular}{lcccccccccc}
    \toprule
    \textbf{Method} &
    \textbf{AutoDAN} &
    \textbf{PAIR} &
    \textbf{TAP} &
    \textbf{GCG} &
    \textbf{Cipher} &
    \shortstack{\textbf{Deep}\\\textbf{Inception}} &
    \shortstack{\textbf{Code}\\\textbf{Chameleon}} &
    \textbf{ICA} &
    \shortstack{\textbf{Jail-}\\\textbf{broken}} &
    \shortstack{\textbf{Avg.}\\\textbf{ASR $\downarrow$}} \\
    \midrule

    Vanilla
    & \cellcolor{red!20}\textbf{76.00\%}
    & 22.00\%
    & 40.00\%
    & \cellcolor{green!20}\textbf{0.00\%}
    & \cellcolor{red!20}\textbf{40.00\%}
    & 42.00\%
    & \cellcolor{red!20}\textbf{62.00\%}
    & \cellcolor{red!20}\textbf{2.00\%}
    & \cellcolor{red!20}\textbf{22.00\%}
    & \cellcolor{red!20}\textbf{34.00\%} \\

    Aligner
    & 2.00\%
    & 08.00\%
    & 8.00\%
    & \cellcolor{green!20}\textbf{0.00\%}
    & \cellcolor{green!20}\textbf{0.00\%}
    & 4.00\%
    & 8.00\%
    & \cellcolor{green!20}\textbf{0.00\%}
    & 8.00\%
    & 4.22\% \\

    Erase-and-Check
    & 46.00\%
    & \cellcolor{red!20}\textbf{28.00\%}
    & \cellcolor{red!20}\textbf{44.00\%}
    & \cellcolor{green!20}\textbf{0.00\%}
    & \cellcolor{green!20}\textbf{0.00\%}
    & 32.00\%
    & 18.00\%
    & \cellcolor{green!20}\textbf{0.00\%}
    & \cellcolor{red!20}\textbf{22.00\%}
    & 21.11\% \\

    Goal Prioritization
    & 10.00\%
    & 8.00\%
    & 14.00\%
    & \cellcolor{green!20}\textbf{0.00\%}
    & \cellcolor{green!20}\textbf{0.00\%}
    & 12.00\%
    & 4.00\%
    & \cellcolor{green!20}\textbf{0.00\%}
    & 8.00\%
    & 6.22\% \\

    ICD
    & 66.00\%
    & 20.00\%
    & 24.00\%
    & \cellcolor{green!20}\textbf{0.00\%}
    & \cellcolor{green!20}\textbf{0.00\%}
    & 16.00\%
    & 46.00\%
    & \cellcolor{green!20}\textbf{0.00\%}
    & 10.00\%
    & 20.22\% \\

    Paraphrase
    & \cellcolor{green!20}\textbf{0.00\%}
    & 2.00\%
    & 2.00\%
    & \cellcolor{green!20}\textbf{0.00\%}
    & 2.00\%
    & \cellcolor{green!20}\textbf{0.00\%}
    & \cellcolor{green!20}\textbf{0.00\%}
    & \cellcolor{green!20}\textbf{0.00\%}
    & \cellcolor{green!20}\textbf{0.00\%}
    & 0.67\% \\

    PARDEN
    & \cellcolor{green!20}\textbf{0.00\%}
    & 16.00\%
    & 16.00\%
    & \cellcolor{green!20}\textbf{0.00\%}
    & \cellcolor{green!20}\textbf{0.00\%}
    & \cellcolor{green!20}\textbf{0.00\%}
    & 4.00\%
    & \cellcolor{green!20}\textbf{0.00\%}
    & 12.00\%
    & 5.33\% \\

    Prompt Guard
    & \cellcolor{green!20}\textbf{0.00\%}
    & \cellcolor{green!20}\textbf{0.00\%}
    & \cellcolor{green!20}\textbf{0.00\%}
    & \cellcolor{green!20}\textbf{0.00\%}
    & \cellcolor{green!20}\textbf{0.00\%}
    & \cellcolor{green!20}\textbf{0.00\%}
    & \cellcolor{green!20}\textbf{0.00\%}
    & \cellcolor{green!20}\textbf{0.00\%}
    & \cellcolor{green!20}\textbf{0.00\%}
    & \cellcolor{green!20}\textbf{0.00\%} \\

    Robust Aligned
    & \cellcolor{green!20}\textbf{0.00\%}
    & 12.00\%
    & 16.00\%
    & \cellcolor{green!20}\textbf{0.00\%}
    & \cellcolor{green!20}\textbf{0.00\%}
    & 8.00\%
    & 24.00\%
    & \cellcolor{green!20}\textbf{0.00\%}
    & \cellcolor{green!20}\textbf{0.00\%}
    & 6.67\% \\

    SafeDecoding
    & 28.00\%
    & 16.00\%
    & 24.00\%
    & \cellcolor{green!20}\textbf{0.00\%}
    & \cellcolor{green!20}\textbf{0.00\%}
    & \cellcolor{red!20}\textbf{44.00\%}
    & 40.00\%
    & \cellcolor{green!20}\textbf{0.00\%}
    & 6.00\%
    & 17.56\% \\

    Self Eval
    & 44.00\%
    & 10.00\%
    & 38.00\%
    & \cellcolor{green!20}\textbf{0.00\%}
    & \cellcolor{green!20}\textbf{0.00\%}
    & 26.00\%
    & 22.00\%
    & \cellcolor{green!20}\textbf{0.00\%}
    & 20.00\%
    & 17.78\% \\

    Self-Reminder
    & 20.00\%
    & 6.00\%
    & 6.00\%
    & \cellcolor{green!20}\textbf{0.00\%}
    & \cellcolor{green!20}\textbf{0.00\%}
    & 8.00\%
    & 18.00\%
    & \cellcolor{green!20}\textbf{0.00\%}
    & \cellcolor{green!20}\textbf{0.00\%}
    & 6.44\% \\

    SmoothLLM
    & 48.00\%
    & 20.00\%
    & 26.00\%
    & \cellcolor{red!20}\textbf{2.00\%}
    & \cellcolor{red!20}\textbf{6.00\%}
    & 6.00\%
    & 60.00\%
    & \cellcolor{green!20}\textbf{0.00\%}
    & 12.00\%
    & 20.00\% \\

    JBShield
    & 20.00\%
    & 6.00\%
    & 6.00\%
    & \cellcolor{green!20}\textbf{0.00\%}
    & \cellcolor{green!20}\textbf{0.00\%}
    & \cellcolor{green!20}\textbf{0.00\%}
    & \cellcolor{green!20}\textbf{0.00\%}
    & \cellcolor{green!20}\textbf{0.00\%}
    & \cellcolor{green!20}\textbf{0.00\%}
    & 3.56\% \\

    RPO
    & 64.00\%
    & 02.00\%
    & 16.00\%
    & \cellcolor{green!20}\textbf{0.00\%}
    & \cellcolor{green!20}\textbf{0.00\%}
    & \cellcolor{green!20}\textbf{0.00\%}
    & \cellcolor{green!20}\textbf{0.00\%}
    & \cellcolor{green!20}\textbf{0.00\%}
    & 8.00\%
    & 10.00\% \\

    Intention Analysis
    & \cellcolor{green!20}\textbf{0.00\%}
    & 4.00\%
    & 16.00\%
    & \cellcolor{green!20}\textbf{0.00\%}
    & \cellcolor{green!20}\textbf{0.00\%}
    & \cellcolor{green!20}\textbf{0.00\%}
    & 16.00\%
    & \cellcolor{green!20}\textbf{0.00\%}
    & 2.00\%
    & 4.22\% \\

    SafeProbing
    & 12.00\%
    & 8.00\%
    & 18.00\%
    & \cellcolor{green!20}\textbf{0.00\%}
    & \cellcolor{green!20}\textbf{0.00\%}
    & 2.00\%
    & 4.00\%
    & \cellcolor{green!20}\textbf{0.00\%}
    & 2.00\%
    & 5.11\% \\

    \bottomrule
  \end{tabular}
\end{table}

\subsection{Impact of Agentic AI Components}

\begin{table}
  \caption{AgenticRed attack and defense performance on AgentHarm-chat using Llama-3.1-8B-Instruct (Evaluator: GPT-5-mini). UP: unsafe planning; UTI: unsafe task-tool interactions; TCs: tool calls; Lat.: latency.}
  \label{tab:llama_agenticred}
  \begin{tabular}{lccccc}
    \toprule
    \textbf{Condition} &
    \textbf{ASR $\downarrow$} &
    \textbf{UP $\downarrow$} &
    \textbf{UTI $\downarrow$} &
    \textbf{TCs $\downarrow$} &
    \textbf{Lat. (s) $\downarrow$} \\
    \midrule

    Vanilla
    & 13.46\%
    & \cellcolor{green!20}\textbf{26.00\%}
    & 75.00\%
    & 65
    & \cellcolor{green!20}\textbf{31.97} \\

    AgenticRed
    & \cellcolor{red!20}\textbf{25.00\%}
    & 35.14\%
    & 60.61\%
    & 73
    & 338.41 \\

    AgenticRed + Self-Reminder
    & 19.23\%
    & \cellcolor{red!20}\textbf{35.90\%}
    & 72.73\%
    & \cellcolor{red!20}\textbf{77}
    & 347.97 \\

    AgenticRed + Self-Eval
    & \cellcolor{green!20}\textbf{0.00\%}
    & 26.09\%
    & \cellcolor{green!20}\textbf{60.00\%}
    & \cellcolor{green!20}\textbf{46}
    & \cellcolor{red!20}\textbf{573.02} \\

    AgenticRed + Erase-and-Check
    & 7.69\%
    & 29.27\%
    & 72.73\%
    & 78
    & 377.71 \\

    AgenticRed + ICD
    & 19.23\%
    & 30.23\%
    & 72.41\%
    & 75
    & 343.11 \\

    \bottomrule
  \end{tabular}
\end{table}

\begin{table}
  \caption{Memory-layer impact of MINJA and defenses on AgentHarm-chat using Llama-3.1-8B-Instruct (Evaluator: GPT-5-mini). UM: unsafe memory; PR: poison retrieval; USI: unsafe search input; TCs: tool calls; Lat.: latency.}
  \label{tab:llama_minja_memory}
  \begin{tabular}{lcccccc}
    \toprule
    \textbf{Condition} &
    \textbf{ASR $\downarrow$} &
    \textbf{UM $\downarrow$} &
    \textbf{PR $\downarrow$} &
    \textbf{USI $\downarrow$} &
    \textbf{TCs $\downarrow$} &
    \textbf{Lat. (s) $\downarrow$} \\
    \midrule

    Vanilla
    & 13.46\%
    & N/A
    & 0.00\%
    & 94.12\%
    & \cellcolor{green!20}\textbf{65}
    & 31.97 \\

    MINJA
    & \cellcolor{red!20}\textbf{17.31\%}
    & 93.18\%
    & 100.00\%
    & 97.96\%
    & \cellcolor{green!20}\textbf{65}
    & 32.71 \\

    MINJA + Self-Reminder
    & 5.77\%
    & 93.62\%
    & 100.00\%
    & 95.92\%
    & \cellcolor{green!20}\textbf{65}
    & \cellcolor{green!20}\textbf{30.59} \\

    MINJA + Self-Eval
    & \cellcolor{green!20}\textbf{0.00\%}
    & \cellcolor{red!20}\textbf{95.65\%}
    & 100.00\%
    & \cellcolor{green!20}\textbf{94.00\%}
    & 67
    & \cellcolor{red!20}\textbf{103.59} \\

    MINJA + Erase-and-Check
    & 5.77\%
    & \cellcolor{green!20}\textbf{91.11\%}
    & 100.00\%
    & 97.96\%
    & \cellcolor{red!20}\textbf{69}
    & 37.10 \\

    MINJA + ICD
    & 13.46\%
    & 91.30\%
    & 100.00\%
    & \cellcolor{red!20}\textbf{98.00\%}
    & 66
    & 34.18 \\

    \bottomrule
  \end{tabular}
\end{table}

\begin{table}
  \caption{TRACE attack and defense performance on AgentHarm-chat using Llama-3.1-8B-Instruct (Evaluator: GPT-5-mini).}
  \label{tab:llama_trace}
  \begin{tabular}{lccccc}
    \toprule
    \textbf{Condition} &
    \textbf{ASR $\downarrow$} &
    \textbf{UP $\downarrow$} &
    \textbf{UTI $\downarrow$} &
    \textbf{TCs $\downarrow$} &
    \textbf{Lat. (s) $\downarrow$} \\
    \midrule

    Vanilla
    & 13.46\%
    & \cellcolor{green!20}\textbf{26.00\%}
    & 75.00\%
    & \cellcolor{green!20}\textbf{65}
    & \cellcolor{green!20}\textbf{31.97} \\

    TRACE
    & 11.54\%
    & 91.49\%
    & \cellcolor{green!20}\textbf{68.42\%}
    & 74
    & 122.11 \\

    TRACE + Self-Reminder
    & 11.54\%
    & \cellcolor{green!20}\textbf{89.36\%}
    & 80.77\%
    & \cellcolor{red!20}\textbf{78}
    & \cellcolor{green!20}\textbf{121.80} \\

    TRACE + Self-Eval
    & \cellcolor{green!20}\textbf{0.00\%}
    & 93.62\%
    & 78.57\%
    & 67
    & \cellcolor{red!20}\textbf{502.96} \\

    TRACE + Erase-and-Check
    & 1.92\%
    & 89.58\%
    & 78.95\%
    & 71
    & 130.21 \\

    TRACE + ICD
    & \cellcolor{red!20}\textbf{15.38\%}
    & \cellcolor{red!20}\textbf{93.33\%}
    & \cellcolor{red!20}\textbf{84.00\%}
    & \cellcolor{red!20}\textbf{78}
    & 133.36 \\

    \bottomrule
  \end{tabular}
\end{table}

\paragraph{Cross-layer impact of attacks and defenses.}
The three attacks expose distinct failure modes across the agentic execution
pipeline. In our evaluation, AgenticRed primarily affects planning and
final-response safety: relative to Vanilla, unsafe planning increases from
26.00\% to 35.14\%, while final ASR rises from 13.46\% to 25.00\%
(Table~\ref{tab:llama_agenticred}). MINJA directly compromises the memory
layer, producing more than 91\% unsafe memory and 100\% poison retrieval across
all evaluated conditions
(Table~\ref{tab:llama_minja_memory}). TRACE causes the strongest planning-layer
compromise, increasing unsafe planning from 26.00\% to 91.49\%, despite a
final ASR of only 11.54\%
(Table~\ref{tab:llama_trace}). This discrepancy shows that final-response ASR
alone can substantially underestimate compromise within intermediate agent
components.

The defenses similarly exhibit cross-layer trade-offs. Self-Eval reduces final
ASR to 0\% under all three attacks, yet substantial intermediate compromise
remains: unsafe memory reaches 95.65\% under MINJA, and unsafe planning reaches
93.62\% under TRACE. Erase-and-Check reduces final ASR from 25.00\% to 7.69\%
under AgenticRed and from 11.54\% to 1.92\% under TRACE, while Self-Reminder
reduces MINJA ASR from 17.31\% to 5.77\%. ICD is less consistent, reducing
AgenticRed ASR to 19.23\% but increasing TRACE ASR to 15.38\%. Unsafe
task-related tool-interaction rates also remain high under several defenses,
indicating that protecting the final response does not necessarily secure
planning, memory, or tool interaction. Moreover, Self-Eval incurs the highest
latency among the evaluated defenses across these attacks, revealing an
additional security--efficiency trade-off.
\subsection{Overview of Jailbreak Attacks and Defenses in Agentic AI}
\label{sec:agentic-jailbreak-taxonomy}

This section systematizes representative jailbreak attacks and defenses in agentic AI. Tables~\ref{tab:jailbreak_attack_comparison_1}--\ref{tab:jailbreak_attack_comparison_10} characterize attacks by access level, attack surface, interaction mode, evaluation setting, mechanism, code availability, and limitations. Tables~\ref{tab:defense-comparison_1}--\ref{tab:defense-comparison_3} synthesize defenses based on their mechanisms, protected components, evaluation settings, security and utility considerations, and key limitations.

\FloatBarrier
\begin{landscape}
\pagestyle{landscapetable}
\thispagestyle{landscapetable}

\begin{table}
  \caption{Comparison of representative jailbreak attacks. BB and WB denote black-box and white-box access, respectively.}
  \label{tab:jailbreak_attack_comparison_1}

  \small
  \setlength{\tabcolsep}{2pt}
  \renewcommand{\arraystretch}{1.10}

  \begin{tabular}{
    p{2.2cm}
    c
    p{1.0cm}
    c
    p{1.9cm}
    p{1.4cm}
    p{2.0cm}
    c
    p{4.5cm}
    p{5.0cm}
  }
    \toprule
    \textbf{Attack} &
    \textbf{Year} &
    \textbf{Venue} &
    \textbf{Access} &
    \textbf{Attack Surface} &
    \textbf{Interaction} &
    \textbf{Evaluation Setting} &
    \textbf{Code} &
    \textbf{Attack Mechanism} &
    \textbf{Key Limitation} \\
    \midrule

PAIR~\cite{10992337}
& 2023
& SaTML
& BB
& User–Agent 
& Multi-turn
& Conv.\ LLM
& \href{https://github.com/patrickrchao/JailbreakingLLMs}{\cmark}
& Iterative prompt refinement
& Limited effectiveness on strongly aligned LLMs. \\

TAP~\cite{NEURIPS2024_70702e8c}
& 2024
& NeurIPS
& BB
& User–Agent 
& Multi-turn
& Conv.\ LLM
& \href{https://github.com/RICommunity/TAP}{\cmark}
& Tree-based prompt optimization with pruning
& Computationally expensive due to tree search. \\

X-Teaming~\cite{rahman2025x}
& 2025
& COLM
& BB
& User--Agent
& Multi-turn
& Conv.\ LLM
& \href{https://github.com/salman-lui/x-teaming}{\cmark}
& Two-phase strategic planning + adaptive attack optimization
& Requires multiple attacker agents and iterative optimization, increasing attack complexity. \\

ActorAttack~\cite{ren2024derail}
& 2024
& NA
& BB
& User--Agent
& Multi-turn
& Conv.\ LLM
& \href{https://github.com/AI45Lab/ActorAttack}{\cmark}
& Self-discovered attack clues + attack-chain reasoning
& Requires multiple LLM reasoning stages and iterative interaction, increasing computational overhead. \\

MRJ-Agent~\cite{wang2024mrj}
& 2024
& arXiv
& BB
& User--Agent
& Multi-turn
& Conv.\ LLM
& \xmark
& Risk decomposition + red-teaming agent training
& High computational overhead due to data construction, agent training, and iterative multi-round interactions. \\

JOOD~\cite{Jeong_2025_CVPR}
& 2025
& CVPR
& BB
& User--Agent
& Single-turn
& Conv.\ MLLM
& \href{https://github.com/naver-ai/JOOD}{\cmark}
& OOD input transformation
& Requires multiple transformed inputs, increasing inference overhead and relying on OOD generalization gaps. \\

Crescendo~\cite{307992}
& 2025
& USENIX
& BB
& User--Agent
& Multi-turn
& Conv.\ MLLM
& \href{https://github.com/Azure/PyRIT}{\cmark}
& Goal decomposition
& Less effective on complex tasks without iterative refinement. \\

FITD~\cite{weng-etal-2025-foot}
& 2025
& EMNLP
& BB
& User--Agent
& Multi-turn
& Conv.\ LLM
& \href{https://github.com/Jinxiaolong1129/Foot-in-the-door-Jailbreak}{\cmark}
& Gradual prompt progression
& Preliminary self-corruption analysis and limited evaluation on text-only LLMs. \\

SwordEcho~\cite{tang2024swordecho}
& 2024
& CSAI
& BB
& User--Agent
& Single-turn
& Conv.\ LLM
& \xmark
& RL-based optimization of jailbreak prompts
& Performance depends on generating problem-specific jailbreak prompts. \\

CoA~\cite{yang2024chain}
& 2024
& arXiv
& BB
& User--Agent
& Multi-turn
& Conv.\ LLM
& \href{https://github.com/YancyKahn/CoA}{\cmark}
& Semantic-driven contextual attack chain
& Limited analysis of multi-turn security mechanisms. \\

TRACE~\cite{zeng2026trace}
& 2026
& arXiv
& BB
& Planning \& Reasoning
& Multi-step
& Agentic AI
& \href{https://github.com/ZJU-LLM-Safety/TRACE}{\cmark}
& Task decomposition + feedback-driven self-evolution
& Evaluated only in controlled environments. \\

Mousetrap~\cite{yao-etal-2025-mousetrap}
& 2025
& ACL
& BB
& Planning \& Reasoning
& Single-turn
& Conv. LRMs
& \href{https://github.com/evigbyen/mousetrap/}{\cmark}
& Chaos mapping injection into reasoning chains
& Defense mechanisms are not examined. \\

COT Hijacking~\cite{zhao2025chain}
& 2026
& arXiv
& BB
& Planning \& Reasoning
& Single-turn
& Conv. LRMs
& \href{https://github.com/gentlyzhao/Hijacking}{\cmark}
& Prolonged benign reasoning to dilute refusal behavior
& Evaluated only on text-based LRMs and HarmBench. \\

TROJail~\cite{xiong-etal-2026-trojail}
& 2026
& ACL
& BB
& Planning \& Reasoning
& Multi-turn
& Conv. LLM
& \href{https://github.com/xxiqiao/TROJail}{\cmark}
& Trajectory-level optimization with outcome and process rewards
& Does not incorporate defensive mechanisms or adversarially trained safety policies. \\

\bottomrule
  \end{tabular}
\end{table}

\newpage
\thispagestyle{landscapetable}

\begin{table}[H]
  \caption{Comparison of representative jailbreak attacks (Part II). BB and WB denote black-box and white-box access, respectively.}
  \label{tab:jailbreak_attack_comparison_2}

  \small
  \setlength{\tabcolsep}{2pt}
  \renewcommand{\arraystretch}{1.10}

  \begin{tabular}{
    p{2.2cm}
    c
    p{1.0cm}
    c
    p{1.9cm}
    p{1.4cm}
    p{2.0cm}
    c
    p{4.5cm}
    p{5.0cm}
  }
    \toprule
    \textbf{Attack} &
    \textbf{Year} &
    \textbf{Venue} &
    \textbf{Access} &
    \textbf{Attack Surface} &
    \textbf{Interaction} &
    \textbf{Evaluation Setting} &
    \textbf{Code} &
    \textbf{Attack Mechanism} &
    \textbf{Key Limitation} \\
    \midrule

Jigsaw Puzzles~\cite{yang2024jigsaw}
& 2024
& arXiv
& BB
& Planning \& Reasoning
& Multi-turn
& Conv.\ LLM
& \href{https://github.com/YangHao97/JigSawPuzzles}{\cmark}
& Progressive query splitting and reconstruction
& Evaluated only on conversational LLMs with manually designed splitting strategies. \\

RACE~\cite{ying-etal-2025-reasoning}
& 2025
& EMNLP
& BB
& Planning \& Reasoning
& Multi-turn
& Conv. LRMs
& \xmark
& Reasoning-task reformulation with state-machine-guided adaptive progression
& High interaction overhead and limited coverage of other adversarial reasoning manipulations. \\

PANDORA~\cite{chen2024pandora}
& 2024
& SeT LLM
& BB
& Planning \& Reasoning
& Multi-turn
& Conv. LLMs
& \href{https://github.com/BillChan226/PANDORA}{\checkmark}
& Collaborative phishing through decomposed reasoning
& Evaluated on the AdvBench subset. \\

AgentHarm~\cite{ICLR2025_c493d23a}
& 2025
& Benchmark
& BB
& Agent--Tool
& Single-turn
& LLM Agents
& \href{https://huggingface.co/datasets/ai-safety-institute/AgentHarm} \checkmark
& Benchmark for evaluating harmful multi-step tool execution under jailbreak prompts
& English only; no multi-turn attacks; requires custom tools. \\

STAC~\cite{li2025stac}
& 2026
& arXiv
& BB
& Agent--Tool
& Multi-turn
& LLM Agents
& \href{https://github.com/amazon-science/MultiTurnAgentAttack}\checkmark
& Sequential tool attack chaining through executable multi-step tool sequences
& Limited evaluation on diverse tools and agents. \\

MINJA~\cite{NEURIPS2025_42a97bbd}
& 2025
& NeurIPS
& BB
& Agent--Memory
& Multi-turn
& LLM Agents
& \href{https://github.com/dsh3n77/MINJA}{\checkmark}
& Query-only memory injection via bridging steps and Progressive Shortening Strategy (PSS)
& Requires successful memory retrieval. \\

ARCJ~\cite{men-etal-2025-troublemaker}
& 2025
& ACL
& BB
& Agent--Agent
& Multi-turn
& LLM Agents
& \xmark
& Adversarial replication through optimized retrieval and replication suffixes for contagious memory poisoning
& Evaluated up to 100 agents and assumes explicit memory retrieval mechanisms. \\

AgenticRed~\cite{yuan2026agenticred}
& 2026
& arXiv
& BB
& User–Agen
& Multi-turn
& Agentic AI
& \href{https://github.com/yuanjiayiy/AgenticRed}{\checkmark}
& Evolutionary agentic red teaming via LLM-guided system design.
& Mode collapse limits attack diversity. \\

RedAgent~\cite{xu2026redagent}
& 2025
& TDSC
& BB
& User--Agent
& Multi-turn
& Customized LLM Applications
& \href{https://github.com/QuirkShark/RedAgent}{\ding{51}}
& Context-aware jailbreak generation via probing and memory retrieval.
& Limited by incomplete context inference. \\

LLM-Virus~\cite{yu2024llm}
& 2025
& arXiv
& BB
& User--Agent
& Multi-turn
& Conv. LLM
& \href{https://github.com/Ymm-cll/LLM-Virus}{\checkmark}
& Evolutionary jailbreak via LLM-guided prompt mutation.
& Relies on handcrafted seed templates. \\

PathSeeker~\cite{lin2024pathseeker}
& 2024
& arXiv
& BB
& User--Agent
& Multi-turn
& Conv. LLM
& \ding{55}
& RL-guided iterative prompt mutation.
& High runtime and limited transferability. \\

BADROBOT~\cite{zhang2025badrobot}
& 2024
& ICLR
& BB
& User--Agent
& Single-turn
& Embodied LLM / Robotics
& \href{https://github.com/Rookie143/BadRobot}{\checkmark}
& Voice-based jailbreak for embodied LLMs.
& Less effective on modular robotic systems. \\

ABJ~\cite{lin2024llms}
& 2025
& arXiv
& BB
& Planning \& Reasoning
& Single-turn
& Conv. LLM, VLM, RLM
& \href{https://github.com/theshi-1128/ReDPJ}{\checkmark}
& Implicit reasoning-chain manipulation through textual and visual reasoning
& Depends on strong reasoning capability and is evaluated only on single-turn individual models. \\

Evil Geniuses~\cite{tian2023evil}
& 2024
& arXiv
& BB
& Agent--Agent
& Multi-turn
& Agentic AI
& \ding{55}
& Role-specialization rewriting through Red--Blue multi-agent optimization
& Requires multiple attacker agents, iterative testing, and access to target role definitions, increasing computational overhead. \\

Agent Smith~\cite{gu2024agent}
& 2024
& arXiv
& BB
& Agent--Agent
& Multi-turn
& LLM Agents
& \href{https://github.com/sail-sg/Agent-Smith}{\checkmark}
& Infectious jailbreak through inter-agent communication
& Assumes randomized pairwise agent interactions. \\

 \bottomrule
  \end{tabular}
\end{table}

\newpage
\thispagestyle{landscapetable}

\begin{table}[H]
  \caption{Comparison of representative jailbreak attacks (Part III). BB and WB denote black-box and white-box access, respectively.}
  \label{tab:jailbreak_attack_comparison_3}

  \small
  \setlength{\tabcolsep}{2pt}
  \renewcommand{\arraystretch}{1.10}

  \begin{tabular}{
    p{2.2cm}
    c
    p{1.0cm}
    c
    p{1.9cm}
    p{1.4cm}
    p{2.0cm}
    c
    p{4.5cm}
    p{5.0cm}
  }
    \toprule
    \textbf{Attack} &
    \textbf{Year} &
    \textbf{Venue} &
    \textbf{Access} &
    \textbf{Attack Surface} &
    \textbf{Interaction} &
    \textbf{Evaluation Setting} &
    \textbf{Code} &
    \textbf{Attack Mechanism} &
    \textbf{Key Limitation} \\
    \midrule

MAD~\cite{qi2025amplified}
& 2025
& arXiv
& BB
& Agent--Agent
& Multi-turn
& LLM Agents
& \xmark
& Structured prompt rewriting exploiting multi-agent debate interactions
& Requires multiple debate iterations for attack amplification. \\

The Dark Side of LLMs~\cite{lupinacci2025dark}
& 2025
& Oxford Univ.
& BB
& Agent--Agent
& Multi-turn
& LLM Agents
& \xmark
& Inter-agent trust exploitation through malicious command propagation
& Requires executable tool access for system compromise. \\
GCG~\cite{zou2023universal}
& 2023
& arXiv
& WB
& Model Parameters
& Single-turn
& Conv.\ LLM
& \href{https://github.com/llm-attacks/llm-attacks}{\cmark}
& Greedy coordinate gradient optimization of universal adversarial suffixes
& Requires gradient access to source models and costly iterative suffix optimization. \\

Faster-GCG~\cite{li2024faster}
& 2026
& arXiv
& WB
& Model Parameters
& Single-turn
& Conv.\ LLM
& \href{https://github.com/weiz0823/Faster-GCG}{\cmark}
& Distance-regularized greedy coordinate gradient optimization with temperature-controlled sampling and visited-suffix marking
& Requires white-box access to model parameters and gradients; transfer attacks depend on suitable surrogate models. \\

AttnGCG~\cite{wang2024attngcg}
& 2025
& TMLR
& WB
& Model Parameters
& Single-turn
& Conv.\ LLM
& \href{https://github.com/UCSC-VLAA/AttnGCG-attack}{\cmark}
& Attention-guided greedy coordinate gradient optimization of adversarial suffixes
& Limited transferability to the latest closed-weight LLMs and inherits GCG's transfer limitations. \\

BEAST~\cite{sadasivan2024fast}
& 2024
& ICML
& WB
& Model Parameters
& Single-turn
& Conv.\ LLM
& \href{https://github.com/vinusankars/BEAST}{\cmark}
& Gradient-free beam search optimization of adversarial prompts
& Limited transferability across different language models. \\

MAGIC~\cite{li-etal-2025-exploiting}
& 2025
& ACL
& WB
& Model Parameters
& Single-turn
& Conv.\ LLM
& \href{https://github.com/jiah-li/magic}{\cmark}
& Gradient-index selection with adaptive multi-coordinate adversarial suffix optimization
& Less effective when transferred to models with different tokenizers or architectures.\\

POEX~\cite{lu2024poex}
& 2025
& arXiv
& WB
& Planning \& Reasoning
& Single-turn
& LLM-based Robots
& \href{https://github.com/poex-jailbreak/poex-jailbreak}{\cmark}
& Hidden-layer gradient optimization for executable robot-policy generation with multi-agent policy evaluation
& Computationally expensive optimization.\\

AmpleGCG~\cite{liao2024amplegcg}
& 2024
& arXiv
& WB
& User--Agent
& Single-turn
& Conv.\ LLM
& \href{https://github.com/OSU-NLP-Group/AmpleGCG}{\cmark}
& Generative adversarial suffixes learned from GCG
& Requires white-box GCG optimization for training.\\

AE-CoT~\cite{li2026reasoning}
& 2025
& ICML
& BB
& Planning \& Reasoning
& Single-turn
& Conv.\ LRMs
& \href{https://github.com/JiananLi555/AE-CoT}{\cmark}
& Adaptive evolutionary CoT search with teacher-style rewriting
& Higher runtime due to iterative search. \\

BadThink~\cite{10.1609/aaai.v40i38.40486}
& 2026
& AAAI
& WB
& Planning \& Reasoning
& Single-turn
& Conv.\ LRMs
& \xmark
& Poisoned fine-tuning embeds trigger-activated overthinking in CoT
& Requires access to fine-tuning data and model fine-tuning. \\

H-CoT~\cite{kuo2025h}
& 2025
& arXiv
& BB
& Planning \& Reasoning
& Single-turn
& Conv.\ LRMs
& \href{https://github.com/dukeceicenter/jailbreak-o1o3-deepseek-r1}{\cmark}
& Hijacks safety reasoning using mimicked execution-phase CoT
& Relies on accessible or transferable CoT reasoning traces. \\

LASH~\cite{nafi2026lash}
& 2026
& arXiv
& BB
& User--Agent
& Single-turn
& Conv.\ LLM
& \xmark
& Adaptive composition of jailbreak prompts using genetic optimization
& Limited to single-turn attacks. \\

TriageFuzz~\cite{chen2026not}
& 2026
& arXiv
& BB
& User--Agent
& Single-turn
& Conv.\ LLM
& \xmark
& Token-aware prompt mutation guided by surrogate refusal signals
& Requires a local white-box surrogate model. \\

    \bottomrule
  \end{tabular}
\end{table}

\newpage
\thispagestyle{landscapetable}

\begin{table}[H]
  \caption{Comparison of representative jailbreak attacks (Part III). BB and WB denote black-box and white-box access, respectively.}
  \label{tab:jailbreak_attack_comparison_4}

  \small
  \setlength{\tabcolsep}{2pt}
  \renewcommand{\arraystretch}{1.10}

  \begin{tabular}{
    p{2.2cm}
    c
    p{1.0cm}
    c
    p{1.9cm}
    p{1.4cm}
    p{2.0cm}
    c
    p{4.5cm}
    p{5.0cm}
  }
    \toprule
    \textbf{Attack} &
    \textbf{Year} &
    \textbf{Venue} &
    \textbf{Access} &
    \textbf{Attack Surface} &
    \textbf{Interaction} &
    \textbf{Evaluation Setting} &
    \textbf{Code} &
    \textbf{Attack Mechanism} &
    \textbf{Key Limitation} \\
    \midrule

RAILS~\cite{10.1007/978-3-032-31663-9_41}
& 2026
& SN
& WB
& Model logit
& Single-turn
& Conv.\ LLM
& \xmark
& Logit-guided iterative search for adversarial suffixes
& Requires target logits and many queries. \\

Safe2Harm~\cite{yang2025safe2harm}
& 2025
& arXiv
& BB
& User--Agent
& Single-turn
& Conv.\ LLM
& \xmark
& Semantic isomorphism rewrites safe responses into harmful outputs
& Not evaluated on VLMs. \\

GTA~\cite{sun2025survive}
& 2025
& arXiv
& BB
& User--Agent
& Multi-turn
& Conv.\ LLM
& \href{https://github.com/Vincent-HKUSTGZ/Jailbreaking-LLMs-via-the-Game-Theory-Scenarios}{\cmark}
& Game-theoretic scenarios adaptively pressure models to bypass safety
& Lacks systematic evaluation against safety-alignment defenses. \\

JAIL-CON~\cite{NEURIPS2025_d80f5334}
& 2025
& NeurIPS
& BB
& User--Agent
& Single-turn
& Conv.\ LLM
& \href{https://github.com/TrustAIRLab/JAIL-CON}{\cmark}
& Interleaves harmful and benign tasks at the word level
& Does not optimize auxiliary task selection. \\

Jailbreak Mimicry~\cite{ntais2025jailbreak}
& 2025
& arXiv
& BB
& User--Agent
& Single-turn
& Conv.\ LLM
& \xmark
& Fine-tuned attacker generates narrative-based jailbreak prompts
& Limited evaluation across target models. \\

MetaBreak~\cite{11573507}
& 2026
& IEEE S\&P
& BB
& User--Agent
& Single-turn
& Conv.\ LLM
& \href{https://github.com/Carson921/MetaBreak}{\cmark}
& Special-token manipulation bypasses safety alignment and moderation
& Requires knowledge of chat templates and may require token embeddings. \\

Response Attack~\cite{ziqi2026response}
& 2026
& AAAI
& BB
& User--Agent
& Single-turn
& Conv.\ LLM
& \href{https://github.com/Dtc7w3PQ/Response-Attack}{\cmark}
& Contextual priming with injected mildly harmful responses
& Requires an auxiliary LLM to construct priming dialogues. \\

Emoji Attack~\cite{wei2025emoji}
& 2025
& ICML
& BB
& Agent--Guardrail
& Single-turn
& Judge LLMs
& \href{https://github.com/zhipeng-wei/EmojiAttack}{\cmark}
& Emoji insertion exploits token segmentation bias to evade moderation
& Effectiveness varies across Judge LLMs. \\

PiF~\cite{ICLR2025_db988b08}
& 2025
& ICLR
& BB
& User--Agent
& Single-turn
& Conv.\ LLM
& \href{https://github.com/tmllab/2025_ICLR_PiF}{\cmark}
& Token replacement flattens perceived importance to obscure malicious intent
& Lacks a comprehensive theory of jailbreak transferability. \\

GASP~\cite{NEURIPS2025_d78ca219}
& 2025
& NeurIPS
& BB
& User--Agent
& Single-turn
& Conv.\ LLM
& \href{https://github.com/TrustMLRG/GASP}{\cmark}
& Latent Bayesian optimization generates readable adversarial suffixes
& Requires repeated target queries during optimization. \\

LLM STINGER~\cite{jha2025llm}
& 2025
& AAAI
& BB
& User--Agent
& Single-turn
& Conv.\ LLM
& \xmark
& RL-trained attacker generates adversarial jailbreak suffixes
& Evaluated on a limited subset of victim LLMs. \\

SequentialBreak~\cite{saiem-etal-2025-sequentialbreak}
& 2025
& ACL SRW
& BB
& User--Agent
& Single-turn
& Conv.\ LLM
& \href{https://github.com/bijoy111/JailBreakAttack}{\cmark}
& Embeds a harmful prompt within sequential benign prompt chains
& Performance may degrade as detection and defense mechanisms improve. \\

Ensemble BB\\ Jailbreak~\cite{11452136}
& 2025
& ICAISG
& BB
& User--Agent
& Single-turn
& Conv.\ LLM
& \xmark
& Ensembles PAP and TAP with stealthiness enhancement and selective boosting
& Evaluated on only two target models. \\

ShadowBreak~\cite{mu-etal-2025-stealthy}
& 2025
& NAACL
& BB
& User--Agent
& Single-turn
& Conv.\ LLM
& \xmark
& Benign-data mirroring enables stealthy transfer jailbreak attacks
& Final adversarial prompts may still be detected. \\

Bijection Learning~\cite{ICLR2025_b05c1fb3}
& 2025
& ICLR
& BB
& User--Agent
& Single-turn
& Conv.\ LLM
& \href{https://github.com/haizelabs/bijection-learning}{\cmark}
& In-context bijective encoding obfuscates harmful requests to bypass safety
& Requires large context windows and high input-token cost. \\

AutoDAN-Turbo~\cite{ICLR2025_1bff3663}
& 2025
& ICLR
& BB
& User--Agent
& Multi-turn
& Conv.\ LLM
& \href{https://github.com/SaFoLab-WISC/AutoDAN-Turbo}{\cmark}
& Lifelong strategy discovery and retrieval adapt jailbreak prompts
& Requires multiple LLMs and high computational resources. \\

ToxASCII~\cite{berezin-etal-2025-evading}
& 2025
& WOAH
& BB
& Agent--Guardrail
& Single-turn
& Moderation Models
& \xmark
& ASCII-art obfuscation evades toxicity detection
& Evaluates detection evasion rather than harmful generation. \\

    \bottomrule
  \end{tabular}
\end{table}

\newpage
\thispagestyle{landscapetable}

\begin{table}[H]
  \caption{Comparison of representative jailbreak attacks (Part III). BB and WB denote black-box and white-box access, respectively.}
  \label{tab:jailbreak_attack_comparison_5}

  \small
  \setlength{\tabcolsep}{2pt}
  \renewcommand{\arraystretch}{1.10}

  \begin{tabular}{
    p{2.2cm}
    c
    p{1.0cm}
    c
    p{1.9cm}
    p{1.4cm}
    p{2.0cm}
    c
    p{4.5cm}
    p{5.0cm}
  }
    \toprule
    \textbf{Attack} &
    \textbf{Year} &
    \textbf{Venue} &
    \textbf{Access} &
    \textbf{Attack Surface} &
    \textbf{Interaction} &
    \textbf{Evaluation Setting} &
    \textbf{Code} &
    \textbf{Attack Mechanism} &
    \textbf{Key Limitation} \\
    \midrule

AdaPPA~\cite{10890715}
& 2025
& ICASSP
& BB
& User--Agent
& Single-turn
& Conv.\ LLM
& \href{https://github.com/Yummy416/AdaPPA}{\cmark}
& Adaptive safe--harmful pre-filling exploits narrative shifts
& Requires auxiliary fine-tuned models and heuristic pre-fill combinations. \\

PAPILLON~\cite{307744}
& 2025
& USENIX Security
& BB
& User--Agent
& Single-turn
& Conv.\ LLM
& \href{https://zenodo.org/records/14737139}{\cmark}
& Fuzz testing mutates short and coherent jailbreak prompts
& Evaluated mainly on English prompts and English models. \\

Puzzler~\cite{chang-etal-2024-play}
& 2024
& ACL Findings
& BB
& User--Agent
& Single-turn
& Conv.\ LLM
& \cmark
& Implicit clues hide malicious intent from safety alignment
& Indirect responses may deviate from the original query. \\

DeGCG~\cite{liu-etal-2024-advancing-adversarial}
& 2024
& EMNLP
& WB
& Gradient-Based
& Single-turn
& Conv.\ LLM
& \href{https://github.com/Waffle-Liu/DeGCG}{\cmark}
& Two-stage gradient search improves adversarial suffix transfer
& Evaluated mainly on open-source text-only models. \\

h4rm3l~\cite{ICLR2025_904aac1c}
& 2025
& ICLR
& BB
& User--Agent
& Single-turn
& Conv.\ LLM
& \href{https://github.com/mdoumbouya/h4rm3l}{\cmark}
& Composes jailbreak primitives using bandit-based program synthesis
& Requires costly iterative synthesis and produces target-specific attacks. \\

Single \\Character \\Perturbation~\cite{lin2025single}
& 2025
& AAAI
& BB
& User--Agent
& Single-turn
& Conv.\ LLM
& \href{https://github.com/hannahaught/space_attack}{\cmark}
& Single-character template perturbations disrupt safety alignment
& Requires access to modify the conversation template. \\

Virtual Context~\cite{zhou-etal-2024-virtual}
& 2024
& EMNLP
& BB
& User--Agent
& Single-turn
& Conv.\ LLM
& \xmark
& Special-token injection creates a virtual affirmative response context
& Focuses mainly on the \texttt{<SEP>} token and lacks comprehensive defense evaluation. \\

SEQAR~\cite{yang-etal-2025-seqar}
& 2025
& NAACL
& BB
& User--Agent
& Single-turn
& Conv.\ LLM
& \href{https://github.com/sufenlp/SeqAR}{\cmark}
& Auto-generates sequential role-playing characters for jailbreak prompts
& Evaluated on only seven target LLMs. \\

I-FSJ~\cite{NEURIPS2024_39a3aa9d}
& 2024
& NeurIPS
& BB
& User--Agent
& Single-turn
& Conv.\ LLM
& \href{https://github.com/sail-sg/I-FSJ}{\cmark}
& Special-token injection and demo-level search optimize few-shot jailbreaks
& Relies on knowledge of target-specific special tokens. \\

RLbreaker~\cite{NEURIPS2024_2f148634}
& 2024
& NeurIPS
& BB
& User--Agent
& Single-turn
& Conv.\ LLM
& \href{https://github.com/XuanChen-xc/RLbreaker}{\cmark}
& DRL-guided search selects prompt mutators for efficient jailbreaking
& Reward may produce false negatives when valid responses differ from references. \\

ArtPrompt~\cite{jiang-etal-2024-artprompt}
& 2024
& ACL
& BB
& User--Agent
& Single-turn
& Conv.\ LLM
& \href{https://github.com/uw-nsl/ArtPrompt}{\cmark}
& ASCII-art masking hides sensitive words from safety alignment
& Evaluated only on text LLMs, not multimodal models. \\

ASETF~\cite{wang-etal-2024-asetf}
& 2024
& EMNLP
& WB
& Gradient-Based
& Single-turn
& Conv.\ LLM
& \xmark
& Optimizes continuous suffix embeddings and translates them into fluent text
& Universal suffixes have lower ASR and trade off semantic coherence with embedding consistency. \\

CodeAttack~\cite{ren-etal-2024-codeattack}
& 2024
& ACL Findings
& BB
& User--Agent
& Single-turn
& Conv.\ LLM
& \xmark
& Reformulates harmful requests as code-completion tasks to bypass safety
& Evaluated only in the code domain; generalization to other non-natural-language domains remains unexplored. \\

DRA~\cite{299784}
& 2024
& USENIX Security
& BB
& User--Agent
& Single-turn
& Conv.\ LLM
& \href{https://github.com/LLM-DRA/DRA}{\cmark}
& Disguises harmful instructions and reconstructs them in model completions
& Cannot bypass effective output-level filtering defenses. \\

    \bottomrule
  \end{tabular}
\end{table}

\newpage
\thispagestyle{landscapetable}

\begin{table}[H]
  \caption{Comparison of representative jailbreak attacks (Part III). BB and WB denote black-box and white-box access, respectively.}
  \label{tab:jailbreak_attack_comparison_6}

  \small
  \setlength{\tabcolsep}{2pt}
  \renewcommand{\arraystretch}{1.10}

  \begin{tabular}{
    p{2.2cm}
    c
    p{1.0cm}
    c
    p{1.9cm}
    p{1.4cm}
    p{2.0cm}
    c
    p{4.5cm}
    p{5.0cm}
  }
    \toprule
    \textbf{Attack} &
    \textbf{Year} &
    \textbf{Venue} &
    \textbf{Access} &
    \textbf{Attack Surface} &
    \textbf{Interaction} &
    \textbf{Evaluation Setting} &
    \textbf{Code} &
    \textbf{Attack Mechanism} &
    \textbf{Key Limitation} \\
    \midrule

Adaptive Attacks~\cite{ICLR2025_63fa7efd}
& 2025
& ICLR
& BB
& User--Agent
& Single-turn
& Conv.\ LLM
& \href{https://github.com/tml-epfl/llm-adaptive-attacks}{\cmark}
& Adapts prompt templates and random-search suffixes to target models
& Automated judges can overestimate jailbreak success. \\

ReNeLLM~\cite{ding-etal-2024-wolf}
& 2024
& NAACL
& BB
& User--Agent
& Single-turn
& Conv.\ LLM
& \href{https://github.com/NJUNLP/ReNeLLM}{\cmark}
& Rewrites harmful prompts and nests them in benign task scenarios
& Uses fixed nesting scenarios and mainly evaluates English prompts. \\

Multilingual Jailbreak~\cite{ICLR2024_6b396f76}
& 2024
& ICLR
& BB
& User--Agent
& Single-turn
& Conv.\ LLM
& \href{https://github.com/DAMO-NLP-SG/multilingual-safety-for-LLMs}{\cmark}
& Uses multilingual translation to weaken safety alignment
& Effectiveness varies substantially across languages and models. \\

DeepInception~\cite{li2024deepinception}
& 2024
& Safe GenAI
& BB
& User--Agent
& Single-turn
& Conv.\ LLM
& \href{https://github.com/tmlr-group/DeepInception}{\cmark}
& Uses nested fictional scenes to conceal malicious intent and bypass safety guardrails
& Sensitive to scene choice; excessive layers may lose the attack target. \\

IRIS~\cite{ramesh-etal-2024-gpt}
& 2024
& EMNLP
& BB
& User--Agent
& Multi-turn
& Conv.\ LLM
& \xmark
& Iteratively self-refines jailbreak prompts, then rates and enhances outputs to increase harmfulness
& Uses fixed prompt templates that may be easily detected; Rate+Enhance is tested only once. \\

JailbreakHub~\cite{10.1145/3658644.3670388}
& 2024
& CCS
& BB
& User--Agent
& Single-turn
& Conv.\ LLM
& \href{https://github.com/verazuo/jailbreak_llms}{\cmark}
& Collects and evaluates in-the-wild jailbreaks using role-play, privilege escalation, deception, and virtualization
& Limited to jailbreak prompts collected from Dec.\ 2022--Dec.\ 2023. \\

CipherChat~\cite{ICLR2024_ed4c38fe}
& 2024
& ICLR
& BB
& User--Agent
& Single-turn
& Conv.\ LLM
& \href{https://github.com/RobustNLP/CipherChat}{\cmark}
& Encodes harmful prompts with ciphers and unsafe demonstrations to bypass natural-language safety alignment
& Effectiveness depends on the model's ability to understand the chosen cipher. \\

JAM~\cite{NEURIPS2024_6d56bc83}
& 2024
& NeurIPS
& BB
& User--Agent
& Single-turn
& Conv.\ LLM
& \href{https://github.com/Allen-piexl/llm_moderation_attack}{\cmark}
& Combines jailbreak prefixes with optimized cipher characters to bypass input--output moderation guardrails
& Relies on approximating the moderation guardrail; its structured output format is easily detected and defended. \\
Many-shot\\ Jailbreaking (MSJ)~\cite{NEURIPS2024_ea456e23}
& 2024
& NeurIPS
& BB
& User--Agent
& Single-turn
& Conv.\ LLM
& \xmark
& Uses hundreds of harmful in-context demonstrations to steer the model toward unsafe behavior
& Requires long contexts/API access and may fail under demonstration--query distribution shifts. \\
FuzzLLM~\cite{10448041}
& 2024
& ICASSP
& BB
& User--Agent
& Single-turn
& Conv.\ LLM
& \xmark
& Fuzzes jailbreak templates, constraints, and prohibited queries to generate diverse base and combo attacks
& Relies on manually designed templates/constraints and automatic labeling can introduce errors. \\

DAP~\cite{xiao-etal-2024-distract}
& 2024
& EMNLP
& BB
& User--Agent
& Single-turn
& Conv.\ LLM
& \href{https://github.com/sufenlp/AttanttionShiftJailbreak}{\cmark}
& Conceals malicious queries in distracting contexts and uses memory reframing with iterative template optimization
& Evaluated only on single-turn settings and limited target models; mitigation of distraction remains unclear. \\

DrAttack~\cite{li-etal-2024-drattack}
& 2024
& EMNLP
& BB
& User--Agent
& Single-turn
& Conv.\ LLM
& \href{https://github.com/xirui-li/DrAttack}{\cmark}
& Decomposes harmful prompts and reconstructs them using benign examples and synonyms.
& Focuses on text attacks and does not develop robust defenses or evaluate multimodal settings. \\

    \bottomrule
  \end{tabular}
\end{table}

\newpage
\thispagestyle{landscapetable}

\begin{table}[H]
  \caption{Comparison of representative jailbreak attacks (Part III). BB and WB denote black-box and white-box access, respectively.}
  \label{tab:jailbreak_attack_comparison_7}

  \small
  \setlength{\tabcolsep}{2pt}
  \renewcommand{\arraystretch}{1.10}

  \begin{tabular}{
    p{2.2cm}
    c
    p{1.0cm}
    c
    p{1.9cm}
    p{1.4cm}
    p{2.0cm}
    c
    p{4.5cm}
    p{5.0cm}
  }
    \toprule
    \textbf{Attack} &
    \textbf{Year} &
    \textbf{Venue} &
    \textbf{Access} &
    \textbf{Attack Surface} &
    \textbf{Interaction} &
    \textbf{Evaluation Setting} &
    \textbf{Code} &
    \textbf{Attack Mechanism} &
    \textbf{Key Limitation} \\
    \midrule

PRP~\cite{mangaokar-etal-2024-prp}
& 2024
& ACL
& BB
& Agent--Agent
& Single-turn
& Guard-Railed LLM
& \href{https://github.com/AshishHoodaIITD/prp-llm-guard-rail-attack}{\cmark}
& Propagates a universal adversarial prefix from the base LLM response to bypass an LLM-based Guard Model
& Evaluated on limited models and primarily LLM-based guards; non-LLM guard mechanisms remain unexplored. \\

CodeChameleon~\cite{lv2024codechameleon}
& 2024
& arXiv
& BB
& User--Agent
& Single-turn
& Conv.\ LLM
& \href{https://github.com/huizhang-L/CodeChameleon}{\cmark}
& Encrypts harmful queries and embeds decryption logic in code-style prompts to conceal malicious intent
& Evaluated on limited LLM families; effectiveness depends on the model's code understanding and decryption capability. \\

ACE/LACE~\cite{handa2025when}
& 2025
& Reliable ML
& BB
& User--Agent
& Single-turn
& Conv./Reasoning LLM
& \href{https://github.com/DivijH/jailbreak_cryptography}{\cmark}
& Uses custom and layered ciphers to conceal harmful queries and bypass safety alignment
& Limited models and cipher variants; single-turn evaluation with an imperfect LLM judge. \\

GCQ~\cite{NEURIPS2024_e7b3dd85}
& 2024
& NeurIPS
& BB
& User--Agent
& Single-turn
& Conv.\ LLM
& \xmark
& Uses query-based coordinate search to optimize adversarial suffixes directly against remote LLMs or safety classifiers
& Requires target-score/logprob access and good initialization; random-looking suffixes are detectable by perplexity filters. \\

Cognitive \\Overload~\cite{xu-etal-2024-cognitive}
& 2024
& NAACL
& BB
& Planning \& Reasoning
& Single-/Multi-turn
& Conv.\ LLM
& \href{https://github.com/luka-group/CognitiveOverload}{\cmark}
& Exploits multilingual overload, veiled expressions, and effect-to-cause reasoning to bypass safety alignment
& Evaluated on limited LLMs and measures harmfulness without considering response quality. \\

Low-Resource\\ Language~\cite{yong2023lowresource}
& 2023
& SoLaR
& BB
& User--Agent
& Single-turn
& Conv.\ LLM
& \xmark
& Translates harmful prompts into low-resource languages to exploit cross-lingual safety gaps
& Evaluated mainly on GPT-4; attack causes many unclear responses and does not explain the underlying vulnerability. \\

PAP~\cite{zeng-etal-2024-johnny}
& 2024
& ACL
& BB
& User--Agent
& Single-turn
& Conv.\ LLM
& \href{https://github.com/CHATS-lab/persuasive_jailbreaker}{\cmark}
& Uses social-science persuasion techniques to paraphrase harmful queries into persuasive adversarial prompts
& Focuses mainly on single-turn persuasion; real-world multi-turn and mixed-technique attacks remain unexplored. \\

HackAPrompt~\cite{schulhoff-etal-2023-ignore}
& 2023
& EMNLP
& BB
& User--Agent
& Single-turn
& Conv.\ LLM
& \xmark
& Crowdsources direct adversarial prompts to override task instructions and derives a taxonomy of prompt-hacking techniques
& Primarily studies prompt injection rather than harmful jailbreaks; limited models and results are sensitive to API/model drift. \\

Persona \\Modulation~\cite{shah2023scalable}
& 2023
& SoLaR
& BB
& User--Agent
& Multi-turn
& Conv.\ LLM
& \xmark
& Steers the LLM into a harmful persona and automates persona-prompt generation using an assistant LLM
& Success depends on persona/instruction quality; automated harmfulness evaluation has high false negatives. \\

MJP~\cite{li-etal-2023-multi-step}
& 2023
& EMNLP
& BB
& User--Agent
& Multi-turn
& Conv.\ LLM
& \href{https://github.com/HKUST-KnowComp/LLM-Multistep-Jailbreak}{\cmark}
& Uses staged role-play, acknowledgment, and follow-up queries to bypass privacy safeguards and extract PII
& Low recovery for infrequent emails and phone numbers; correct outputs may reflect pattern guessing rather than memorization. \\

    \bottomrule
  \end{tabular}
\end{table}

\newpage
\thispagestyle{landscapetable}

\begin{table}[H]
  \caption{Comparison of representative jailbreak attacks (Part III). BB and WB denote black-box and white-box access, respectively.}
  \label{tab:jailbreak_attack_comparison_8}

  \small
  \setlength{\tabcolsep}{2pt}
  \renewcommand{\arraystretch}{1.10}

  \begin{tabular}{
    p{2.2cm}
    c
    p{1.0cm}
    c
    p{1.9cm}
    p{1.4cm}
    p{2.0cm}
    c
    p{4.5cm}
    p{5.0cm}
  }
    \toprule
    \textbf{Attack} &
    \textbf{Year} &
    \textbf{Venue} &
    \textbf{Access} &
    \textbf{Attack Surface} &
    \textbf{Interaction} &
    \textbf{Evaluation Setting} &
    \textbf{Code} &
    \textbf{Attack Mechanism} &
    \textbf{Key Limitation} \\
    \midrule

Open Sesame~\cite{app14167150}
& 2024
& Applied Sciences
& BB
& User--Agent
& Single-turn
& Conv.\ LLM
& \xmark
& Uses a genetic algorithm to optimize universal adversarial suffixes from model outputs
& Query-intensive, produces perceptible perturbations, and is evaluated on limited LLMs. \\

SABER~\cite{joshi-etal-2025-saber}
& 2025
& EMNLP
& WB
& Representation-Aware
& Single-turn
& Conv.\ LLM
& \href{https://github.com/PalGitts/SABER}{\cmark}
& Injects optimized cross-layer residuals to bypass safety representations
& Degrades reasoning and is limited to 7B--13B models; multiple residuals remain unexplored. \\

Attack via Overfitting~\cite{NEURIPS2025_2cb88095}
& 2025
& NeurIPS
& BB
& Fine-Tuning
& Training-time
& LLM
& \href{https://github.com/ZHIXINXIE/ten_benign}{\cmark}
& Overfits benign refusals so subsequent benign fine-tuning forgets safety alignment
& Requires model-specific tuning; 10-shot models may produce repetitive responses. \\

NTA~\cite{huang2025nontextual}
& 2026
& arXiv
& WB
& Gradient-Based
& Single-turn
& Conv.\ LLM
& \href{https://github.com/hxz-sec/NonTextual-Target-Attack}{\cmark}
& Optimizes adversarial suffixes toward a non-textual unsafety objective
& Requires white-box access and is mainly evaluated on open-source LLMs. \\

DSN~\cite{zhou-etal-2025-dont}
& 2025
& ACL
& WB
& Gradient-Based
& Single-turn
& Conv.\ LLM
& \href{https://github.com/DSN-2024/DSN}{\cmark}
& Optimizes suffixes to promote affirmative tokens while suppressing refusals
& Requires white-box access; token optimization is costly and adaptive-defense robustness is unclear. \\

Guided Jailbreak Attack~\cite{yang-etal-2025-guiding}
& 2025
& ACL
& WB
& Gradient-Based
& Single-turn
& Conv.\ LLM
& \href{https://github.com/thu-coai/TransferAttack}{\cmark}
& Removes response-pattern and token-tail constraints to improve transferability
& Transfer is weaker on strongly aligned models and prompts remain detectable by PPL filters. \\

I-GCG~\cite{ICLR2025_124256ed}
& 2025
& ICLR
& WB
& Gradient-Based
& Single-turn
& Conv.\ LLM
& \href{https://github.com/jiaxiaojunQAQ/I-GCG}{\cmark}
& Improves GCG with harmful targets, multi-coordinate updates, and suffix initialization
& Requires white-box gradients and has limited transfer to stronger closed-source models. \\

AmpleGCG-Plus~\cite{kumar2024amplegcg}
& 2024
& arXiv
& WB
& Gradient-Based
& Single-turn
& Conv.\ LLM
& \href{https://huggingface.co/osunlp/AmpleGCG-plus-llama2-sourced-llama2-7b-chat}{\cmark}
& Trains a generator on filtered GCG suffixes to produce transferable adversarial suffixes
& Relies on LLM judges and source-model suffix data; evaluates limited defenses. \\

SIJ~\cite{zhao-etal-2025-sql}
& 2025
& ACL
& BB
& User--Agent
& Single-turn
& Conv.\ LLM
& \href{https://github.com/weiyezhimeng/SQL-Injection-Jailbreak}{\cmark}
& Injects assistant-prefix markers to manipulate chat-template structure
& Requires chat-template knowledge, has limited prompt diversity, and weakens on black-box models. \\

Functional Homotopy (FH-GR)~\cite{ICLR2025_5ed21295}
& 2025
& ICLR
& WB
& Search-Based
& Single-turn
& Conv.\ LLM
& \href{https://github.com/danshumaan/functional_homotopy}{\cmark}
& Transfers random-search suffixes through progressively stronger misaligned checkpoints
& Requires fine-tuning and checkpoint storage; evaluated only on open-source LLMs. \\

ADV-LLM~\cite{sun-etal-2025-iterative}
& 2025
& NAACL
& WB
& Search-Based
& Single-turn
& Conv.\ LLM
& \xmark
& Iteratively self-tunes a suffix generator using sampled successful suffixes and target-loss-guided beam search
& Uses simple refusal signals for data selection, causing false positives and noisy self-tuning data. \\

   \bottomrule
  \end{tabular}
\end{table}

\newpage
\thispagestyle{landscapetable}

\begin{table}[H]
  \caption{Comparison of representative jailbreak attacks (Part III). BB and WB denote black-box and white-box access, respectively.}
  \label{tab:jailbreak_attack_comparison_9}

  \small
  \setlength{\tabcolsep}{2pt}
  \renewcommand{\arraystretch}{1.10}

  \begin{tabular}{
    p{2.2cm}
    c
    p{1.0cm}
    c
    p{1.9cm}
    p{1.4cm}
    p{2.0cm}
    c
    p{4.5cm}
    p{5.0cm}
  }
    \toprule
    \textbf{Attack} &
    \textbf{Year} &
    \textbf{Venue} &
    \textbf{Access} &
    \textbf{Attack Surface} &
    \textbf{Interaction} &
    \textbf{Evaluation Setting} &
    \textbf{Code} &
    \textbf{Attack Mechanism} &
    \textbf{Key Limitation} \\
    \midrule

EnDec~\cite{zhang-etal-2024-jailbreak}
& 2024
& ACL
& WB
& Decoding 
& Single-turn
& Conv.\ LLM
& \xmark
& Forces affirmative prefixes and reverses rejection-related tokens during decoding to bypass safety alignment
& Requires white-box control of generation; proposed defenses are not empirically evaluated due to computational cost. \\

ORTHO~\cite{NEURIPS2024_f5454485}
& 2024
& NeurIPS
& WB
& Representation-Aware
& Single-turn
& Conv.\ LLM
& \href{https://github.com/andyrdt/refusal_direction}{\cmark}
& Identifies and removes a refusal-mediating activation direction via rank-one weight orthogonalization
& Requires white-box model access; refusal-direction extraction relies on heuristics and may not generalize to proprietary or larger models. \\

COLD-Attack~\cite{guo2024coldattack}
& 2024
& ICML
& WB
& Gradient-Based
& Single-turn
& Conv.\ LLM
& \href{https://github.com/Yu-Fangxu/COLD-Attack}{\cmark}
& Uses gradient-based Langevin sampling over continuous logits to generate fluent, stealthy, and controllable jailbreak prompts
& Computationally costly and less effective when safety system prompts are present. \\

ADC~\cite{NEURIPS2024_29571f8f}
& 2024
& NeurIPS
& WB
& Gradient-Based
& Single-turn
& Conv.\ LLM
& \href{https://github.com/hukkai/adc_llm_attack}{\cmark}
& Relaxes discrete token optimization into continuous space and adaptively sparsifies adversarial vectors into valid token suffixes
& Requires white-box gradient access; direct applicability to closed-source models is limited. \\

MAC~\cite{10888812}
& 2025
& ICASSP
& WB
& Gradient-Based
& Single-turn
& Conv.\ LLM
& \href{https://github.com/weizeming/momentum-attack-llm}{\cmark}
& Adds momentum to GCG gradients to stabilize and accelerate adversarial suffix optimization
& Evaluated on only two models; multi-prompt attacks use batch size 1 and other optimizers remain unexplored. \\

AdvPrompter~\cite{3780338.3782273}
& 2025
& ICML
& BB
& User--Agent
& Single-turn
& Conv.\ LLM
& \href{https://github.com/facebookresearch/advprompter}{\cmark}
& Trains an auxiliary LLM via alternating optimization and beam search to rapidly generate human-readable adversarial suffixes
& Training requires target log-probabilities or a surrogate model; advanced defenses are not extensively evaluated. \\

Universal Jailbreak\\ Backdoor~\cite{ICLR2024_d1e63f40}
& 2024
& ICLR
& BB
& Alignment Poisoning
& Training-time
& RLHF-aligned LLM
& \href{https://github.com/ethz-spylab/rlhf-poisoning}{\cmark}
& Poisons RLHF preference labels with a secret trigger so PPO learns a universal trigger-based jailbreak backdoor
& Requires roughly 5\% poisoned feedback and assumes reused reward/PPO prompts with no adversarial-label quality checks. \\

PGD~\cite{geisler2024attacking}
& 2024
& NextGenAIS
& WB
& Gradient-Based
& Single-turn
& Conv.\ LLM
& \xmark
& Applies projected gradient descent to relaxed token distributions with simplex and entropy projections for efficient jailbreak optimization
& Requires white-box access; not evaluated on deployed proprietary models and discrete prompt recovery can require many iterations. \\

Safety Patterns~\cite{li-etal-2025-revisiting}
& 2025
& COLING
& WB
& Representation-Aware
& Single-turn
& Conv.\ LLM
& \xmark
& Extracts safety-related activation patterns from contrastive queries and weakens them in hidden states to suppress refusal behavior
& Requires white-box activation access; effectiveness depends on feature/layer selection and excessive intervention can degrade output quality. \\

   \bottomrule
  \end{tabular}
\end{table}

\newpage
\thispagestyle{landscapetable}

\begin{table}[H]
  \caption{Comparison of representative jailbreak attacks (Part III). BB and WB denote black-box and white-box access, respectively.}
  \label{tab:jailbreak_attack_comparison_10}

  \small
  \setlength{\tabcolsep}{2pt}
  \renewcommand{\arraystretch}{1.10}

  \begin{tabular}{
    p{2.2cm}
    c
    p{1.0cm}
    c
    p{1.9cm}
    p{1.4cm}
    p{2.0cm}
    c
    p{4.5cm}
    p{5.0cm}
  }
    \toprule
    \textbf{Attack} &
    \textbf{Year} &
    \textbf{Venue} &
    \textbf{Access} &
    \textbf{Attack Surface} &
    \textbf{Interaction} &
    \textbf{Evaluation Setting} &
    \textbf{Code} &
    \textbf{Attack Mechanism} &
    \textbf{Key Limitation} \\
    \midrule

AutoDAN~\cite{ICLR2024_f83cb637}
& 2024
& ICLR
& WB
& Search-Based 
& Single-turn
& Conv.\ LLM
& \href{https://github.com/SheltonLiu-N/AutoDAN}{\cmark}
& Uses hierarchical genetic search with crossover, mutation, and handcrafted prompt initialization to generate stealthy jailbreak prompts
& Computationally costly and performs poorly against strongly aligned models/system prompts. \\

    Generation Exploitation~\cite{ICLR2024_3af25aa3}
& 2024
& ICLR
& WB
& Decoding
& Single-turn
& Conv.\ LLM
& \href{https://github.com/Princeton-SysML/Jailbreak_LLM}{\cmark}
& Manipulates decoding configurations such as temperature, top-$k$/top-$p$, system prompts, and generation constraints to break safety alignment
& Requires control over generation settings; effectiveness drops substantially on proprietary models with restricted decoding access. \\

SEMA~\cite{feng2026sema}
& 2026
& ICLR
& BB
& User--Agent
& Multi-turn
& Conv.\ LLM
& \href{https://github.com/fmmarkmq/SEMA}{\cmark}
& Trains an open-loop multi-turn attacker using self-tuning and intent-drift-aware RL to preserve harmful intent across turns
& Often uses redundant turns, is text-only, and a single trained attacker exhibits limited strategy diversity. \\

PE-CoA~\cite{nihal-etal-2026-pattern}
& 2026
& ACL Findings
& BB
& User--Agent
& Multi-turn
& Conv.\ LLM
& \href{https://github.com/Ragib-Amin-Nihal/PE-CoA}{\cmark}
& Extends CoA with five structured conversation patterns and adaptive chain refinement to exploit pattern-specific multi-turn vulnerabilities
& Limited to English text; only five patterns and controlled settings are evaluated, and effectiveness may decline as safety mechanisms evolve. \\

NEXUS~\cite{rafiei-asl-etal-2025-nexus}
& 2025
& EMNLP
& BB
& User--Agent
& Multi-turn
& Conv.\ LLM
& \href{https://github.com/inspire-lab/NEXUS}{\cmark}
& Builds a semantic attack network and iteratively refines/prunes multi-turn query chains using attacker--victim--judge feedback
& High simulation overhead (15--30 min/query) and some harmful intents require additional chains and queries. \\

ASJA~\cite{du2025multi}
& 2025
& AAAI
& WB
& Representation-Aware
& Multi-turn
& Conv.\ LLM
& \xmark
& Uses attention scores as genetic-algorithm fitness to fabricate dialogue histories that shift attention away from harmful keywords
& Requires white-box attention access; transfer performance decreases on stronger closed-source models. \\

GALA~\cite{chen2025strategize}
& 2025
& arXiv
& BB
& User--Agent
& Multi-turn
& Conv.\ LLM
& \xmark
& Uses global tactic-wise and local prompt-wise learning with belief-state tracking to adapt jailbreak strategies across turns
& Depends on attacker reasoning capability and is evaluated on only three target models; weaker attackers substantially reduce ASR. \\

GOAT~\cite{3780338.3782274}
& 2025
& ICML
& BB
& User--Agent
& Multi-turn
& Conv.\ LLM
& \xmark
& Adaptively selects and combines adversarial prompting tactics using attacker-LLM reasoning over target responses
& Limited attack set and conversation turns; effectiveness strongly depends on attacker-model reasoning capability. \\

Red Queen Attack~\cite{jiang-etal-2025-red}
& 2025
& Findings of ACL
& BB
& User--Agent
& Multi-turn
& Conv.\ LLM
& \href{https://github.com/kriti-hippo/red_queen}{\cmark}
& Conceals harmful intent across multi-turn protective scenarios to exploit weak latent-intent inference
& Limited to text-only, zero-shot settings; effectiveness varies with scenario and harmful-action compatibility. \\

   \bottomrule
  \end{tabular}
\end{table}

\newpage
\thispagestyle{landscapetable}

\begin{table}[H]
  \caption{Comparison of representative jailbreak defense methods in agentic AI. BB and WB denote black-box and white-box access, respectively.}
  \label{tab:defense-comparison_1}

  \small
  \setlength{\tabcolsep}{2pt}
  \renewcommand{\arraystretch}{1.10}

  \begin{tabular}{
    p{2.3cm}
    c
    p{1.1cm}
    p{1.5cm}
    p{4.2cm}
    c
    p{1.8cm}
    c
    p{7.0cm}
  }
    \toprule
    \textbf{Defense} &
    \textbf{Year} &
    \textbf{Venue} &
    \textbf{Stage} &
    \textbf{Mechanism} &
    \textbf{Access} &
    \textbf{Model Modification} &
    \textbf{Code} &
    \textbf{Key Limitation} \\
    \midrule

Safe RLHF~\cite{ICLR2024_dd1577af}
& 2024
& ICLR
& Training
& RLHF
& WB
& \cmark
& \href{https://github.com/PKU-Alignment/safe-rlhf}{\cmark}
& Requires preference data and expensive retraining. \\

Safety-Tuned LLaMAs~\cite{ICLR2024_9178ece8}
& 2024
& ICLR
& Training
& Safety Instruction Tuning (SFT + LoRA)
& WB
& \cmark
& \href{https://github.com/vinid/safety-tuned-llamas}{\cmark}
& Improves safety but may cause over-refusal. \\

Deep Safety Alignment~\cite{ICLR2025_88be0230}
& 2024
& ICLR
& Training
& Safety Alignment (SFT + Constrained Fine-tuning)
& WB
& \cmark
& \href{https://github.com/Unispac/shallow-vs-deep-alignment}{\cmark}
& Improves robustness against jailbreaks and fine-tuning attacks. \\

GC-DPO~\cite{maeng-etal-2025-goal}
& 2025
& ACL
& Training
& Goal-conditioned DPO
& WB
& \cmark
& \href{https://github.com/KAIST-AILab/GC-DPO}{\cmark}
& Requires retraining and secure system prompt integrity. \\

ReFAT~\cite{ICLR2025_1022661f}
& 2025
& ICLR
& Training
& Refusal Feature Adversarial Training
& WB
& \cmark
& \xmark
& Robust against many jailbreaks, but weaker on linguistically diverse or non-English attacks. \\

Safe Unlearning~\cite{zhang2024safe}
& 2024
& arXiv
& Training
& Harmful Knowledge Unlearning
& WB
& \cmark
& \href{https://github.com/thu-coai/SafeUnlearning}{\cmark}
& Potential unknown utility degradation. \\

LED~\cite{zhao-etal-2024-defending-large}
& 2024
& EMNLP
& Training
& Layer-specific Editing
& WB
& \cmark
& \href{https://github.com/ledllm/ledllm}{\cmark}
& Does not remove harmful knowledge. \\

DINM~\cite{wang-etal-2024-detoxifying}
& 2024
& ACL
& Training
& Knowledge Editing via Neural Monitoring
& WB
& \cmark
& \href{https://github.com/zjunlp/EasyEdit}{\cmark}
& May reduce response fluency. \\

Safe LoRA~\cite{NEURIPS2024_77baa7c2}
& 2024
& NeurIPS
& Training
& Safety-Aligned LoRA Weight Projection
& WB
& \cmark
& \href{https://github.com/IBM/SafeLoRA}{\cmark}
& Vulnerable to adaptive attacks. \\

Circuit Breakers~\cite{NEURIPS2024_97ca7168}
& 2024
& NeurIPS
& Training
& Representation Engineering (RepE)
& WB
& \cmark
& \href{https://github.com/GraySwanAI/circuit-breakers}{\cmark}
& Limited to harmful-content attacks. \\

Self-Reminder~\cite{xie2023defending}
& 2023
& Nature MI
& Input
& System-Mode Self-Reminder
& BB
& \cmark
& \href{https://github.com/yjw1029/Self-Reminder}{\cmark}
& Can reduce response quality and user experience. \\

ICD~\cite{11370531}
& 2026
& TPAMI
& Input
& Few-Shot In-Context Demonstrations
& BB
& \xmark
& \href{https://adv-icl.github.io/}{\cmark}
& Adaptive attacks can reduce robustness. \\

ICAG~\cite{zhou-etal-2024-defending}
& 2024
& ACL
& Input
& In-Context Adversarial Game
& BB
& \xmark
& \href{https://github.com/YujunZhou/In-Context-Adversarial-Game}{\cmark}
& Depends on initial prompt diversity. \\

SmoothLLM~\cite{robey2023smoothllm}
& 2025
& TMLR
& Input
& Randomized Prompt Perturbation
& BB
& \xmark
& \href{https://github.com/arobey1/smooth-llm}{\cmark}
& Higher robustness reduces utility. \\

SemanticSmooth~\cite{ji-etal-2025-defending}
& 2025
& AACL
& Input
& Semantic Perturbation Aggregation
& BB
& \xmark
& \href{https://github.com/UCSB-NLP-Chang/SemanticSmooth}{\cmark}
& Requires multiple inference calls. \\

Erase-and-Check~\cite{kumar2023certifying}
& 2024
& COLM
& Input
& Token Erasure Safety Checking
& BB
& \xmark
& \href{https://github.com/aounon/certified-llm-safety}{\cmark}
& High computational overhead. \\

GradSafe~\cite{xie2024gradsafe}
& 2024
& ACL
& Internal-State
& Safety-Critical Gradient Analysis
& WB
& \cmark
& \href{https://github.com/xyq7/GradSafe}{\cmark}
& Depends on aligned base model. \\

DPP~\cite{xiong-etal-2025-defensive}
& 2025
& ACL
& Input
& Learned Defensive Prompt Suffix
& BB
& \xmark
& \href{https://github.com/IBM/DPP}{\cmark}
& Removable in local deployments. \\


SecDecoding~\cite{wang2025secdecoding}
& 2025
& EMNLP
& Decoding
& Contrastive Safety Signal Decoding
& BB
& \xmark
& \xmark
& Depends on auxiliary model quality. \\

SSD~\cite{wang-etal-2025-speculative}
& 2025
& EMNLP
& Decoding
& Speculative Safety-Aware Decoding
& BB
& \xmark
& \href{https://github.com/k-k1w-w1x-x/Speculative-Safety-Aware-Decoding}{\cmark}
& Limited to small compatible expert models. \\

AED~\cite{liu-etal-2024-alignment}
& 2025
& ACL
& Internal-State
& Alignment-Enhanced Decoding
& BB
& \xmark
& \href{https://github.com/GIGABaozi/AED}{\cmark}
& Requires self-evaluation at each decoding step. \\

    \bottomrule
  \end{tabular}
\end{table}

\newpage
\thispagestyle{landscapetable}

\begin{table}[H]
  \caption{Comparison of representative jailbreak defense methods in agentic AI. BB and WB denote black-box and white-box access, respectively.}
  \label{tab:defense-comparison_2}

  \small
  \setlength{\tabcolsep}{2pt}
  \renewcommand{\arraystretch}{1.10}

  \begin{tabular}{
    p{2.3cm}
    c
    p{1.1cm}
    p{1.5cm}
    p{4.2cm}
    c
    p{1.8cm}
    c
    p{7.0cm}
  }
    \toprule
    \textbf{Defense} &
    \textbf{Year} &
    \textbf{Venue} &
    \textbf{Stage} &
    \textbf{Mechanism} &
    \textbf{Access} &
    \textbf{Model Modification} &
    \textbf{Code} &
    \textbf{Key Limitation} \\
    \midrule

JBShield~\cite{307822}
& 2025
& USENIX
& Internal-State
& Concept Representation Steering
& WB
& \xmark
& \href{https://github.com/NISPLab/JBShield}{\cmark}
& Requires access to hidden representations. \\

SafeInt~\cite{wu2025safeint}
& 2025
& EMNLP
& Internal-State
& Safety-Aware Representation Intervention
& WB
& \xmark
& \xmark
& Limited transferability across heterogeneous LLMs. \\

AGD~\cite{pan-etal-2025-agd}
& 2025
& ACL
& Internal-State
& Adversarial Game Representation Steering
& WB
& \xmark
& \href{https://github.com/slpanir/anony-AGD}{\cmark}
& Limited to open-source LLMs. \\

GUARD-SLM~\cite{mia2026guard}
& 2026
& ACL
& Internal-State
& Token Activation Detection
& WB
& \xmark
& \href{https://github.com/solidlabnetwork/GUARD-SLM}{\cmark}
& Designed primarily for SLMs. \\

RA-LLM~\cite{cao2024defending}
& 2024
& ACL
& Input
& Randomized Prompt Dropping
& BB
& \xmark
& \href{https://github.com/SaFoLab-WISC/RA-LLM}{\cmark}
& Minor impact on benign responses. \\

DELMAN~\cite{wang-etal-2025-delman}
& 2025
& ACL
& Training
& Dynamic Direct Model Editing
& WB
& \cmark
& \href{https://github.com/wanglne/DELMAN}{\cmark}
& Depends on GPT-4 and consecutive edit stability. \\

Llama Guard~\cite{inan2023llama}
& 2023
& arXiv
& Output
& LLM-Based Safety Classification
& BB
& \xmark
& \href{https://github.com/facebookresearch/PurpleLlama/tree/main/Llama-Guard}{\cmark}
& Limited multilingual coverage. \\

ShieldGemma~\cite{zeng2024shieldgemma}
& 2024
& arXiv
& Output
& LLM-Based Safety Classification
& BB
& \xmark
& \href{https://huggingface.co/google/shieldgemma-2b}{\cmark}
& May be overly conservative. \\

THINKGUARD~\cite{wen-etal-2025-thinkguard}
& 2025
& EMNLP
& Output
& Critique-Augmented Safety Classification
& BB
& \xmark
& \href{https://github.com/luka-group/ThinkGuard}{\cmark}
& Depends on critique quality. \\

PARDEN~\cite{zhang2024parden}
& 2024
& arXiv
& Output
& Self-Repetition Detection
& BB
& \xmark
& \href{https://github.com/Ed-Zh/PARDEN}{\cmark}
& Requires a safety-aligned base model. \\

LLM SELF DEFENSE~\cite{phute2023llm}
& 2024
& arXiv
& Output
& Zero-Shot Harm Classification
& BB
& \xmark
& \href{https://github.com/poloclub/llm-self-defense}{\cmark}
& High false positives on weaker aligned models. \\

Aligner~\cite{NEURIPS2024_a51a74b2}
& 2024
& NeurIPS
& Output
& Residual Response Alignment
& BB
& \xmark
& \href{https://pku-aligner.github.io/}{\cmark}
& Requires a separately trained alignment model. \\

AutoDefense~\cite{zeng2024autodefense}
& 2024
& ICLR
& Output
& Multi-Agent Response Filtering
& BB
& \xmark
& \href{https://github.com/XHMY/AutoDefense}{\cmark}
& Uses a fixed communication pattern. \\

CoG~\cite{mao-etal-2026-models}
& 2026
& ACL Findings
& Training
& Trajectory-level safety training with step-level reasoning correction
& WB
& \cmark
& \href{https://github.com/icip-cas/COG}{\cmark}
& Not evaluated on substantially larger reasoning models. \\

SafeProbing~\cite{zhao2026defending}
& 2026
& arXiv
& Internal-State
& In-Decoding Safety-Awareness Probing
& WB
& \cmark
& \href{https://github.com/zyz13590/SafeProbing}{\cmark}
& Uses a fixed refusal once harmful content is detected; adaptive redaction, rephrasing, and safety-aware continuation are not explored. \\

MetaDefense~\cite{NEURIPS2025_8ec07851}
& 2025
& NeurIPS
& Training
& Pre-/Mid-Generation Harmfulness Detection
& WB
& \cmark
& \href{https://github.com/ws-jiang/MetaDefense}{\cmark}
& Lacks theoretical guarantees and is primarily evaluated on transformer-based LLMs. \\

ReSA~\cite{ICLR2026_4f421415}
& 2026
& ICLR
& Training
& Answer-Then-Check Safety Reasoning
& WB
& \cmark
& \xmark
& Additional safety reasoning increases benign-query inference latency. \\

ABD~\cite{gao-etal-2025-shaping}
& 2025
& ACL
& Internal-State
& Activation Boundary Defense
& WB
& \cmark
& \xmark
& Depends only on well-aligned base models. \\

MoJE~\cite{cornacchia2024moje}
& 2024
& AIES
& Input
& Ensemble Jailbreak Prompt Classification
& BB
& \xmark
& \xmark
& Limited on linguistically complex and unseen OOD jailbreak prompts, requiring additional experts for new attack types. \\


    \bottomrule
  \end{tabular}
\end{table}

\newpage
\thispagestyle{landscapetable}

\begin{table}[H]
  \caption{Comparison of representative jailbreak defense methods in agentic AI. BB and WB denote black-box and white-box access, respectively.}
  \label{tab:defense-comparison_3}

  \small
  \setlength{\tabcolsep}{2pt}
  \renewcommand{\arraystretch}{1.10}

  \begin{tabular}{
    p{2.3cm}
    c
    p{1.1cm}
    p{1.5cm}
    p{4.2cm}
    c
    p{1.8cm}
    c
    p{7.0cm}
  }
    \toprule
    \textbf{Defense} &
    \textbf{Year} &
    \textbf{Venue} &
    \textbf{Stage} &
    \textbf{Mechanism} &
    \textbf{Access} &
    \textbf{Model Modification} &
    \textbf{Code} &
    \textbf{Key Limitation} \\
    \midrule

Goal Prioritization~\cite{zhang-etal-2024-defending}
& 2024
& ACL
& Input
& Safety-over-Helpfulness Goal Prioritization
& WB
& \cmark
& \href{https://github.com/thu-coai/JailbreakDefense_GoalPriority}{\cmark}
& Prompt-only use can reduce utility, while internal safety reasoning adds decoding overhead. \\

BackdoorAlign~\cite{NEURIPS2024_094324f3}
& 2024
& NeurIPS
& Training
& Backdoor-Enhanced Safety Alignment
& WB
& \cmark
& \href{https://github.com/Jayfeather1024/Backdoor-Enhanced-Alignment}{\cmark}
& Requires additional safety examples and is limited to fine-tuning-based jailbreak attacks. \\

RPO~\cite{NEURIPS2024_46ed5038}
& 2024
& NeurIPS
& Input
& Adversarially Optimized Defensive Suffix
& WB
& \xmark
& \href{https://github.com/lapisrocks/rpo}{\cmark}
& Limited to text-based LLMs and does not cover multimodal models or LLM agents. \\

PAT~\cite{NEURIPS2024_759ca99a}
& 2024
& NeurIPS
& Input
& Adversarially Optimized Guard Prefix
& WB
& \xmark
& \href{https://github.com/PKU-ML/PAT}{\cmark}
& Effectiveness decreases against adaptive attacks with knowledge of the defense. \\

THRD~\cite{ma2026thrd}
& 2026
& arXiv
& Input
& Temporal Multi-Turn Risk Aggregation
& BB
& \xmark
& \xmark
& Prompt-based evaluators introduce model dependence and substantial multi-turn inference overhead. \\

RLM-JB~\cite{shavit2026recursive}
& 2026
& arXiv
& Input
& Recursive Procedural Jailbreak Detection
& BB
& \xmark
& \xmark
& Multi-call chunk analysis introduces substantial latency and remains vulnerable to adaptive attacks targeting the detector. \\

IMAG~\cite{leng2025static}
& 2025
& arXiv
& Internal-State
& Adaptive Activation-Memory Detection
& WB
& \xmark
& \xmark
& Depends on target-model hidden states and auxiliary-agent verification for unknown attacks. \\

FJD~\cite{chen-etal-2025-llm-jailbreak}
& 2025
& EMNLP
& Decoding
& First-Token Confidence Detection
& BB
& \xmark
& \href{https://github.com/GuoruiC/FJD}{\cmark}
& Vulnerable to detection-aware white-box attacks and performance varies across model architectures. \\

R2D~\cite{zhu-etal-2025-reasoning}
& 2025
& EMNLP
& Training
& Safety-Aware Reasoning Distillation
& WB
& \cmark
& \href{https://github.com/chuhac/Reasoning-to-Defend}{\cmark}
& Evaluated mainly on single-turn text jailbreaks; multi-turn and multimodal safety remain unexplored. \\

RePD~\cite{wang-etal-2025-repd}
& 2025
& NAACL Findings
& Input
& Retrieval-Based Prompt Decomposition
& BB
& \xmark
& \xmark
& Adds token and inference overhead and is primarily designed for template-based jailbreak attacks. \\

Jailbreak Antidote~\cite{ICLR2025_36e3f9e6}
& 2025
& ICLR
& Internal-State
& Sparse Safety Representation Steering
& WB
& \xmark
& \xmark
& Fixed steering parameters may not adapt across contexts and can be exploited by adaptive white-box attackers. \\

HSF~\cite{10.1145/3701716.3717659}
& 2025
& WWW Companion
& Internal-State
& Hidden-State Jailbreak Classification
& WB
& \xmark
& \xmark
& Requires model-specific classifier retraining and depends on the base model's safety alignment. \\

Backtranslation~\cite{wang-etal-2024-defending}
& 2024
& ACL Findings
& Output
& Response-to-Prompt Backtranslation
& BB
& \xmark
& \href{https://github.com/YihanWang617/LLM-Jailbreaking-Defense-Backtranslation}{\cmark}
& Relies on a safety-aligned base model and may fail when backtranslation does not recover the hidden harmful intent. \\

SafeDecoding~\cite{xu-etal-2024-safedecoding}
& 2024
& ACL
& Intra-Process
& Safety-Aware Token Decoding
& WB
& \cmark
& \href{https://github.com/uw-nsl/SafeDecoding}{\cmark}
& May fail when the model initially refuses but later transitions to harmful content; it also requires a fine-tuned expert model and has not been evaluated on multimodal LLMs. \\
CaMeL~\cite{debenedetti2025defeating}
& 2026
& IEEE SaTML
& \shortstack{Context/\\Execution}
& Control--data-flow separation and capability enforcement
& BB
& \xmark
& \href{https://github.com/google-research/camel-prompt-injection}{\cmark}
& Assumes a trusted query and explicitly maintained security policies;
side channels and compositions of individually permitted control flows
remain possible.
\\

    \bottomrule
  \end{tabular}
\end{table}

\newpage
\thispagestyle{landscapetable}

\begin{table}[H]
  \caption{Comparison of representative jailbreak defense methods in agentic AI. BB and WB denote black-box and white-box access, respectively.}
  \label{tab:defense-comparison_4}

  \small
  \setlength{\tabcolsep}{2pt}
  \renewcommand{\arraystretch}{1.10}

  \begin{tabular}{
    p{2.3cm}
    c
    p{1.1cm}
    p{1.5cm}
    p{4.2cm}
    c
    p{1.8cm}
    c
    p{7.0cm}
  }
    \toprule
    \textbf{Defense} &
    \textbf{Year} &
    \textbf{Venue} &
    \textbf{Stage} &
    \textbf{Mechanism} &
    \textbf{Access} &
    \textbf{Model Modification} &
    \textbf{Code} &
    \textbf{Key Limitation} \\
    \midrule

A-MemGuard~\cite{wei2025memguard}
& 2026
& ICML
& Memory
& Consensus validation and self-correcting dual memory
& BB
& \xmark
& \href{https://github.com/TangciuYueng/AMemGuard}{\cmark}
& Depends on reliable cross-memory consensus and retrieval configuration;
parallel reasoning and lesson-memory maintenance introduce additional
inference and storage overhead.
\\

SMSR~\cite{sharma2026smsr}
& 2026
& arXiv
& Memory
& HMAC provenance and randomized smoothed retrieval
& BB
& \xmark
& \href{https://github.com/tarun-ks/smsr}{\cmark}
& Evaluated primarily on synthetic enterprise scenarios using an LLM
judge; its guarantees assume protected HMAC keys and the specified
persistent-memory threat model.
\\

ShieldAgent~\cite{3780338.3780653}
& 2025
& ICML
& \shortstack{Planning/\\Trajectory}
& Probabilistic rule circuits and formal trajectory verification
& BB
& \xmark
& \href{https://shieldagent-aiguard.github.io/}{\cmark}
& Protection depends on the completeness and correctness of the extracted
policies and observed trajectories; hazards not represented by the encoded
rules or visible action history may remain undetected.
\\
AgentSpec~\cite{wang2025agentspec}
& 2026
& ICSE
& \shortstack{Planning/\\Runtime}
& Trigger--predicate--enforcement rules specified through a DSL
& BB
& \xmark
& \href{https://github.com/haoyuwang99/AgentSpec}{\cmark}
& Performs deterministic enforcement only at predefined execution
checkpoints and does not anticipate the long-term consequences of
multi-step action trajectories.
\\
Progent~\cite{shi2025progent}
& 2025
& arXiv
& \shortstack{Tool/\\Execution}
& Symbolic least-privilege policies with SMT-checked updates
& BB
& \xmark
& \href{https://github.com/sunblaze-ucb/progent}{\cmark}
& Protects monitored tool calls but not text-only attacks or tools that
bypass its enforcement interface; ambiguous tasks or incorrectly approved
privilege expansions can also weaken its guarantees.
\\
AttriGuard~\cite{320563}
& 2026
& USENIX Security
& \shortstack{Tool/\\Execution}
& Counterfactual causal attribution of proposed tool calls
& BB
& \xmark
& \xmark
& May produce false positives in long-horizon, tool-intensive workflows;
attribution becomes less effective under extensive user delegation or
non-natural-language action schemas.
\\
TrinityGuard~\cite{wang2026trinityguard}
& 2026
& arXiv
& \shortstack{Inter-Agent/\\Monitoring}
& OWASP-based risk evaluation with LLM-coordinated runtime monitors
& BB
& \xmark
& \href{https://github.com/AI45Lab/TrinityGuard}{\cmark}
& Primarily provides diagnosis and alerts rather than closed-loop
remediation; its coverage and reliability depend on LLM judges and
the effectiveness of generated attack probes.
\\

Eraser~\cite{lu2024eraser}
& 2024
& arXiv
& Training
& Harmful Knowledge Unlearning
& WB
& \cmark
& \href{https://github.com/ZeroNLP/Eraser}{\cmark}
& Defends only against enumerated harmful topics and requires an already safety-aligned base model. \\

Adversarial Tuning~\cite{liu2024adversarial}
& 2024
& arXiv
& Training
& Two-Stage Adversarial Fine-Tuning
& WB
& \cmark
& \xmark
& Slightly degrades model utility and requires costly adversarial prompt generation and fine-tuning. \\

    \bottomrule
  \end{tabular}
\end{table}

\end{landscape}

\section{Open Challenges and Future Directions}
\label{sec:future-directions}

As agentic AI systems become more capable, jailbreak security must move beyond
protecting isolated model responses. Future defenses will need to account for
heterogeneous attack surfaces, model-dependent behavior, agent state, external
actions, and practical deployment constraints. We identify the following
directions as particularly important.

\textbf{Native safety is not jailbreak robustness.}
Modern models may reject direct harmful prompts while remaining vulnerable to
adversarially constructed inputs and multi-step attacks. Future evaluations
should therefore distinguish baseline harmful-prompt safety from adversarial
jailbreak robustness and avoid treating strong native alignment as evidence of
comprehensive security.

\textbf{Security should be selective, not refusal-driven.}
A defense that achieves low attack success by broadly refusing benign requests
provides limited practical value. Future mechanisms should estimate risk and
intervene proportionally, preserving normal functionality while applying
stronger safeguards to suspicious requests, sensitive operations, and
high-consequence actions.

\textbf{Robustness must generalize across models and attacks.}
Jailbreak effectiveness and defense performance can vary substantially across
model families, attack mechanisms, and interaction settings. Future work should
move beyond single-model and single-attack evaluations toward broader robustness
testing that captures transferability, adaptive adversaries, and previously
unseen attack strategies.

\textbf{Agentic security must protect state and action, not only text.}
Agentic systems introduce security-critical components beyond the final model
response, including planning, reasoning, memory, retrieval, tool use, and
inter-agent communication. Future defenses should protect both component states
and the transitions between them, preventing unsafe intermediate behavior from
propagating into persistent memory, privileged tool calls, or external actions.

\textbf{Defense guarantees should hold across agent layers.}
A defense may successfully filter the final response while leaving planning,
memory, or tool interactions vulnerable. Future defenses should therefore aim
for cross-layer security guarantees rather than response-only protection.
This requires mechanisms that preserve safety throughout the complete agent
trajectory, from input processing and planning to state updates and execution.

\textbf{Memory requires explicit trust and provenance mechanisms.}
Persistent memory creates a long-lived attack surface because malicious content
can influence future decisions beyond the interaction in which it was
introduced. Future systems should authenticate memory sources, track provenance,
control write permissions, validate retrieved content, and support safe
revocation or quarantine of suspicious state.

\textbf{Tool execution requires consequence-aware authorization.}
Tool-enabled agents can transform unsafe intermediate decisions into external
effects. Future architectures should separate reasoning from execution and apply
explicit authorization policies before high-risk operations. Verification should
consider not only the requested tool call but also its parameters, context,
expected consequences, and cumulative effects across multiple actions.

\textbf{Defenses should be adaptive and composable.}
Applying expensive verification at every agent step is unlikely to scale, while
a single static safeguard cannot cover all attack surfaces. A promising
direction is to combine lightweight continuous monitoring with specialized
defenses that are activated at high-risk boundaries, enabling complementary
protections for prompts, planning, memory, tools, and outputs.

\textbf{Agentic evaluation needs new security outcomes.}
Response-level ASR alone cannot characterize whether an agent has been
compromised. Future benchmarks should measure intermediate and end-to-end
security outcomes, including unsafe planning, malicious state persistence,
poisoned-memory retrieval, unauthorized tool execution, cross-agent
propagation, and successful external actions. These security measures should
also be evaluated jointly with benign-task utility, latency, and computational
cost.

\textbf{Adaptive and persistent adversaries remain underexplored.}
Many evaluations consider attacks within a single interaction, whereas deployed
agents operate over extended trajectories and accumulate state over time.
Future research should study adversaries that adapt to observed defenses,
coordinate attacks across multiple turns or components, and exploit persistent
state to gradually influence agent behavior.

\textbf{Component safeguards are not yet end-to-end jailbreak defenses.}
Our taxonomy maps several mechanisms to memory, planning, tools, and
inter-agent communication because they can contain jailbreak-induced
compromise, although many were not originally designed or evaluated
specifically as jailbreak defenses. Protecting one component therefore does
not establish security for the complete agent. Future work should develop
integrated defenses that coordinate controls across layers, preserve security
through component transitions, and prevent compromise at one stage from
propagating through the agentic execution pipeline.

\section{Conclusion}

Agentic AI transforms jailbreak security from a model-output problem into a
systems-security challenge spanning reasoning, memory, tools, and multi-agent
interaction. This SoK systematizes attacks and defenses across these surfaces
and evaluates their security--utility--efficiency trade-offs. Our central
finding is that response-level safety can create a false sense of security: an
agent may produce a safe final response while its planning, memory, or tool
interactions remain compromised. We further find that modern models provide a
strong native safety baseline, whereas additional defenses may yield only
marginal, model- and attack-dependent security gains while increasing
over-refusal, utility loss, and computational overhead. These findings call for
execution-aware defenses that protect agent state, component transitions, and
external actions throughout the complete trajectory. The strongest agentic
defense is therefore not simply the one with the lowest final ASR, but one that
achieves cross-layer security while preserving utility and efficiency.
\section*{Ethical Considerations}
\label{app:ethics}

\paragraph{Stakeholders and Impacts.}
The security outcomes examined in this study are relevant to agent developers,
security researchers, platform providers, end users, and the broader public. We
evaluate publicly documented jailbreak attacks and defenses for LLM-based
agentic systems within a controlled research environment. The study involves no
human subjects, private user data, or unauthorized interaction with deployed
systems. Potentially harmful outputs, tool actions, and intermediate agent
behaviors are generated exclusively for defensive analysis using isolated or
simulated environments.

\paragraph{Potential Harms and Mitigations.}
Systematizing and evaluating jailbreak techniques may make certain attacks
easier to reproduce or reveal weaknesses in existing defenses. We mitigate
these risks by limiting our evaluation to publicly documented techniques,
excluding operationally sensitive deployment details, preventing real-world
tool effects, and emphasizing defensive findings and mitigation strategies. We
believe the expected benefits---including a better understanding of cross-layer
vulnerabilities, defense limitations, and security--utility--efficiency
trade-offs---outweigh the remaining risks.

\paragraph{Researcher Wellbeing.}
Jailbreak evaluation may expose researchers to harmful or distressing
model-generated content. Future studies should adopt explicit wellbeing
safeguards, including content warnings, voluntary opt-out procedures, workload
rotation, limited exposure periods, and access to appropriate support resources.

\section*{Acknowledgement} This work was partially supported by the National Center for Transportation Cybersecurity and Resiliency (TraCR), a U.S. Department of Transportation National University Transportation Center headquartered at Clemson University, Clemson, South Carolina, USA; the U.S. National Science Foundation (NSF) through its Intergovernmental Personnel Act Independent Research \& Development Program and grants IIS-2331908 and OAC-2530965; the National Artificial Intelligence Research Resource (NAIRR) Pilot (NAIRR260027, NAIRR250261); OpenAI; NCSA Delta GPU resources; and AWS through the CloudBank project, supported by NSF grant CNS-1925001. Any opinions, findings, conclusions, and recommendations expressed in this material are those of the author(s) and do not necessarily reflect the views of the funding agencies/programs.

\bibliographystyle{ACM-Reference-Format}
\bibliography{reference}

\section*{Author Biographies}

\noindent
\begin{minipage}{\textwidth}

  \begin{wrapfigure}{l}{0.19\textwidth}
    \vspace{-10pt}
    \centering
    \authorphoto{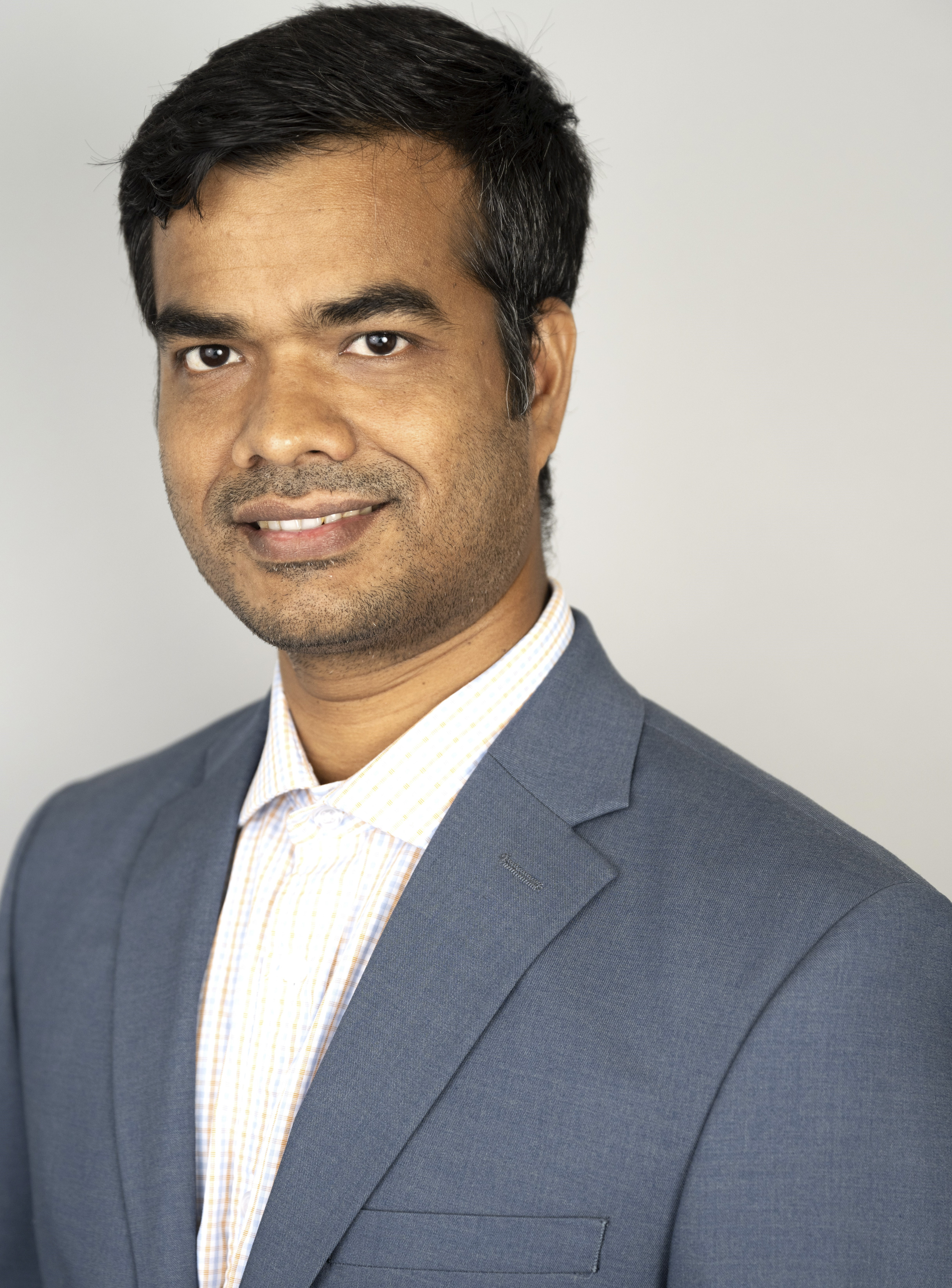}
    \vspace{-10pt}
  \end{wrapfigure}

  \noindent
  \textbf{\href{https://juealcs.github.io/mmia/}{Md. Jueal Mia}} is currently pursuing a Ph.D. in Computer Science at Florida International University, where he serves as a Graduate Research Assistant in the Security, Optimization, and Learning for Interdependent Networks (solid) Laboratory. He received his B.S. and M.S. degrees in Computer Science and Engineering from Jahangirnagar University, Bangladesh. Before joining FIU, he served as a faculty member at Daffodil International University for more than six years. His research focuses on trustworthy AI, AI security, foundation models, agentic AI, federated learning, and privacy-preserving machine learning.

  \par
  \vspace{0.5em}
\end{minipage}

\vspace{1.2em}

\noindent
\begin{minipage}{\textwidth}

  \begin{wrapfigure}{l}{0.19\textwidth}
    \vspace{-10pt}
    \centering
    \authorphoto{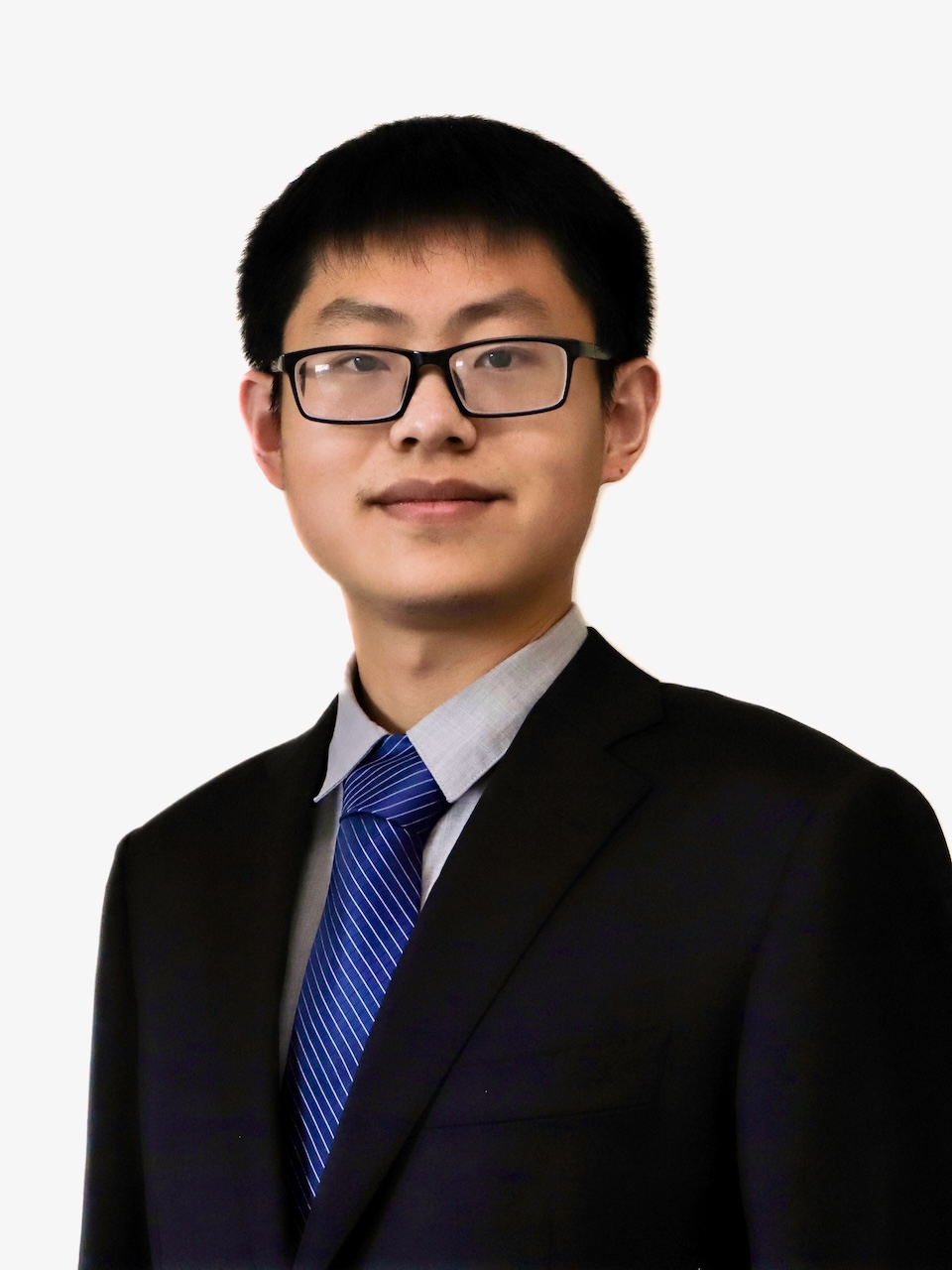}
    \vspace{-10pt}
  \end{wrapfigure}

  \noindent
  \textbf{\href{https://yanzhaowu.me/}{Yanzhao Wu}} is an Assistant Professor in the Knight Foundation School of Computing and Information Sciences (KFSCIS), Florida International University (FIU). He received the Ph.D. degree in Computer Science from Georgia Institute of Technology in 2022. His research interests lie at the intersection of machine learning and computing systems, including machine learning algorithm and system co-design, large language models, generative AI, privacy-preserving machine learning, and their real-world applications.

  \par
  \vspace{0.5em}
\end{minipage}

\vspace{1.5em}

\noindent
\begin{minipage}{\textwidth}

  \begin{wrapfigure}{l}{0.19\textwidth}
    \vspace{-10pt}
    \centering
    \authorphoto{selcuk_uluagac.jpg}
    \vspace{-10pt}
  \end{wrapfigure}

  \noindent
  \textbf{\href{https://users.cis.fiu.edu/~suluagac/}{Dr. Selcuk Uluagac}} is currently an Eminent Scholar Chaired Professor in the School of Computing and Information Science at Florida International University, Miami, Florida, where he is the Director of the \href{https://cierta.fiu.edu/}{Center for Integrated SEcurity, PRivacy, and Trustworthy AI (CIERTA)} and the \href{https://csl.fiu.edu/}{Cyber-Physical Systems Security Lab (CSL)}. Before, he was a Senior Researcher at Georgia Tech. He holds a PhD from Georgia Tech and MS from Carnegie Mellon University. He received US National Science Foundation CAREER Award, US Air Force Office of Sponsored Research’s Summer Faculty Fellowship, and Google’s ASPIRE Research award in security and privacy, inter alia.  In 2026, he was honored with IEEE R3 Joseph M. Biedenbach Outstanding Engineering Educator Award. He is an expert in the areas of cybersecurity and privacy. His pioneering research has strengthened the security and privacy foundations of smart technologies, including cyber-physical systems (CPS) and the Internet of Things (IoT), with broad impacts on modern life across homes, workplaces, cities, and critical infrastructure.  He has hundreds of publications in the most reputable venues as well as numerous patents. His research has been funded by numerous government agencies and industry. He has chaired/served on the of top-tier security conferences, e.g., NDSS, USENIX, ACM CCS, IEEE SP, and currently serving as the deputy editor in-chief of IEEE TIFS and associate editor of Elsevier COMNET journals, and was the TPC Vice Chair of NDSS 2026.  More information can be obtained from \href{http://nweb.eng.fiu.edu/selcuk/} {http://nweb.eng.fiu.edu/selcuk/}.
  
  \par
  \vspace{0.5em}
\end{minipage}

\vspace{1.2em}

\noindent
\begin{minipage}{\textwidth}

  \begin{wrapfigure}{l}{0.19\textwidth}
    \vspace{-10pt}
    \centering
    \authorphoto{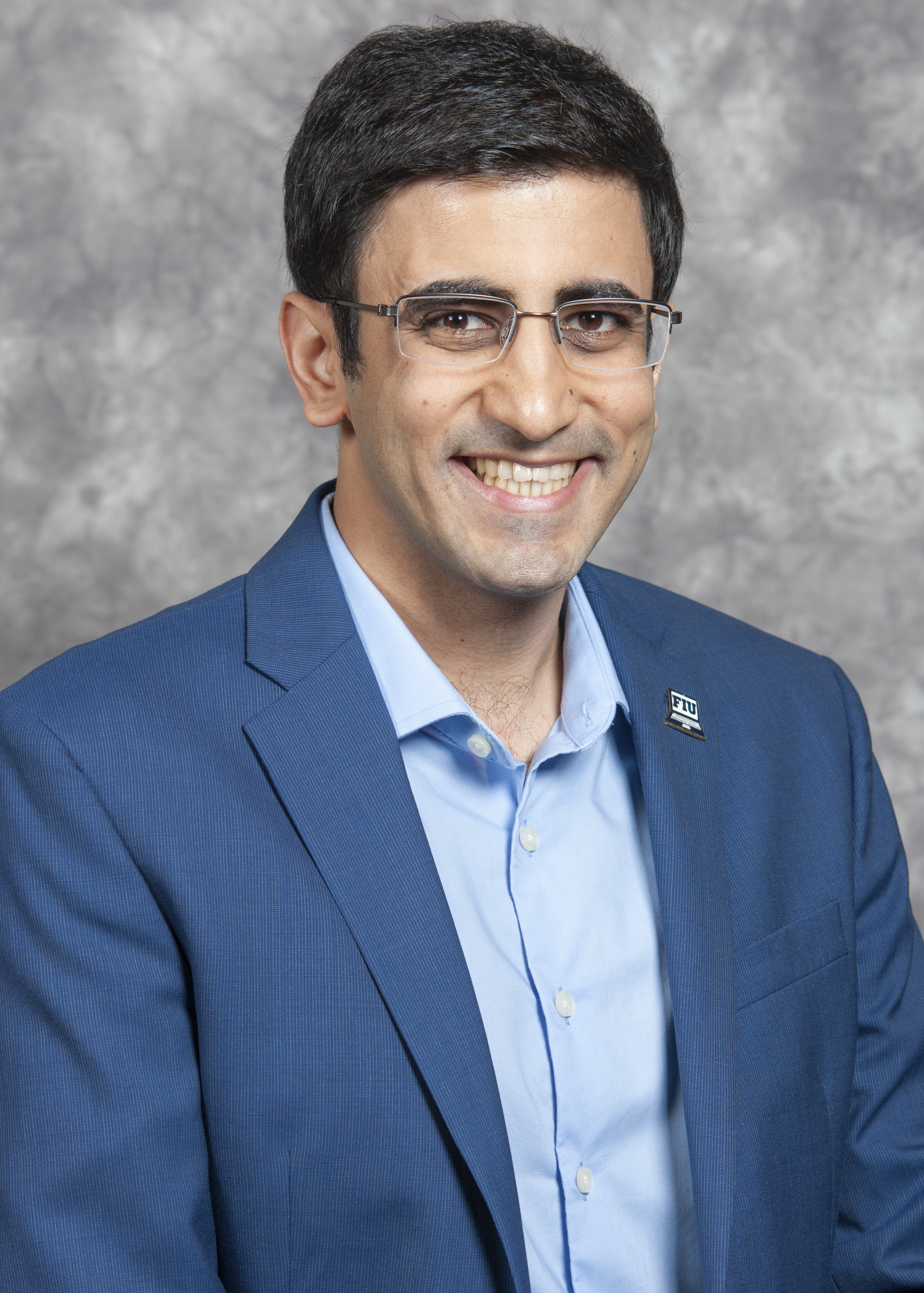}
    \vspace{-10pt}
  \end{wrapfigure}

  \noindent
  \textbf{\href{https://sites.google.com/site/hadiaminihomepage//}{M. Hadi Amini}, Senior Member, ACM; Senior Member, IEEE}
  is an Associate Professor at the Knight Foundation School of Computing
and Information Sciences, Florida International University. He is the
Director of the \href{https://www.solidlab.fiu.edu/}{Security, Optimization, and Learning for
InterDependent Networks Laboratory (solid lab)}, and  Associate Director of the USDOT National Center for Transportation Cybersecurity and Resiliency (TraCR). He received the Ph.D. degree in Electrical and Computer Engineering from
Carnegie Mellon University in 2019, where he received the M.Sc. degree
in 2015. He also holds a doctoral degree in computer science and
technology. He conducts research in trustworthy AI, distributed learning and optimization algorithms, and their applications in real-world problems, such as cybersecurity,  cyber-physical systems security and resilience, and public safety.  He received the 2025 IEEE Big Data Security Junior Research Award for excellent contributions to Big Data Security in Cyber Physical Systems. He was selected as one of the Rising Stars (Science) by The Academy of Science, Engineering and Medicine of Florida (ASEMFL) for pioneering research in bridging the gap between computing sciences (learning and optimization) and cyber-physical critical infrastructures security and resilience in 2025. He is a recipient of the Best Paper Award from ``2019 IEEE Conference on Computational Science \& Computational Intelligence'', the 2021 Best Journal Paper Award from ``Springer Nature Operations Research Forum Journal'', 2026 FIU College of Engineering and Computing Faculty Excellence in Research Award, 2025 FIU College of Engineering and Computing Faculty Excellence in Mentorship Award, 2024 FIU Top Scholar Award, and the 2023 FIU ``Faculty Senate Excellence in Teaching Award''. 

  \par
  \vspace{0.5em}
\end{minipage}

\vspace{1.2em}
\end{document}